\documentclass[10pt,twocolumn,letterpaper]{article}

\usepackage[pagenumbers]{cvpr} 

\usepackage{graphicx}
\usepackage{amsmath}
\usepackage{amssymb}
\usepackage{booktabs}
\usepackage{colortbl}
\usepackage{xcolor}
\usepackage{graphicx}
\usepackage{multirow}

\usepackage[pagebackref,breaklinks,colorlinks]{hyperref}

\usepackage[capitalize]{cleveref}
\crefname{section}{Sec.}{Secs.}
\Crefname{section}{Section}{Sections}
\Crefname{table}{Table}{Tables}
\crefname{table}{Tab.}{Tabs.}

\def\cvprPaperID{*****} 
\def\confName{CVPR}
\def\confYear{2022}

\begin{document}

\title{FeatureGS: Eigenvalue-Feature Optimization in 3D Gaussian Splatting for Geometrically Accurate and Artifact-Reduced Reconstruction}

\author{Miriam Jäger, Markus Hillemann, Boris Jutzi\\
Institute of Photogrammetry and Remote Sensing\\
Karlsruhe Institute of Technology, Karlsruhe, Germany\\
{\tt\small miriam.jaeger@kit.edu, markus.hillemann@kit.edu, boris.jutzi@kit.edu}
}

\maketitle


\begin{abstract}
3D Gaussian Splatting (3DGS) has emerged as a powerful approach for 3D scene reconstruction using 3D Gaussians. However, neither the centers nor surfaces of the Gaussians are accurately aligned to the object surface, complicating their direct use in point cloud and mesh reconstruction. Additionally, 3DGS typically produces floater artifacts, increasing the number of Gaussians and storage requirements.
To address these issues, we present FeatureGS, which incorporates an additional geometric loss term based on an eigenvalue-derived 3D shape feature into the optimization process of 3DGS. The goal is to improve geometric accuracy and enhance properties of planar surfaces with reduced structural entropy in local 3D neighborhoods.
We present four alternative formulations for the geometric loss term based on 'planarity' of Gaussians, as well as 'planarity', 'omnivariance', and 'eigenentropy' of Gaussian neighborhoods.
We provide quantitative and qualitative evaluations on 15 scenes of the DTU benchmark dataset focusing on following key aspects: Geometric accuracy and artifact-reduction, measured by the Chamfer distance, and memory efficiency, evaluated by the total number of Gaussians. Additionally, rendering quality is monitored by Peak Signal-to-Noise Ratio.
FeatureGS achieves a 30\% improvement in geometric accuracy, reduces the number of Gaussians by 90\%, and suppresses floater artifacts, while maintaining comparable photometric rendering quality. The geometric loss with 'planarity' from Gaussians provides the highest geometric accuracy, while 'omnivariance' in Gaussian neighborhoods reduces floater artifacts and number of Gaussians the most.
This makes FeatureGS a strong method for geometrically accurate, artifact-reduced and memory-efficient 3D scene reconstruction, enabling the direct use of Gaussian centers for geometric representation.
\end{abstract}


\section{Introduction}
\label{sec:intro}
The creation of geometric 3D scene reconstructions has developed rapidly since the introduction of Neural Radiance Fields (NeRFs) \cite{mildenhall_et_al_2020}. In NeRFs, a network implicitly describes the scene by estimating color and volume density for each position and direction. In contrast, 3D Gaussian Splatting (3DGS) offers new possibilities for 3D scene and point cloud reconstruction as it represents the scene through 3D Gaussians. These are ellipsoid-like structures, characterized by scaling, rotation, and color. During the optimization process, the 3D Gaussians are projected onto the image. To minimize the photometric error between the rendered images and the training images, the Gaussians are refined and adapted. 
Unlike NeRFs, Gaussians in 3DGS explicitly represent the scene where geometric information is allegedly present. 
Nevertheless, the centers and surfaces of Gaussians do not directly represent the object surface, which makes their direct use for 3D point cloud and mesh reconstruction impractical. In addition, the 3DGS often leads to floater artifacts, which further increase the already high number of Gaussians and thus storage requirements.

In this work, we present FeatureGS, which incorporates four different formulations of an additional geometric loss term based on eigenvalue-derived 3D shape features into the optimization process of 3DGS.
3D shape features are widely used for tasks for semantic interpretation and point cloud classification \cite{Weinmann_2015, Weinmann_2017_Feature_Relevance}. Thereby the 3D covariance matrix (3D structure tensor), derived from a point and its local neighborhood, is well-known to characterize such shape properties \cite{Weinmann_2015}. The three eigenvalues, \( \lambda_1 \geq \lambda_2 \geq \lambda_3 \geq 0 \), correspond to an orthogonal system of eigenvectors (\( \epsilon_1, \epsilon_2, \epsilon_3 \)), which indicate the direction (\textit{rotation}) of the three ellipsoid principal axes and correspond to the extent (\textit{scales}) of the 3D ellipsoid along the principal axes. Based on the behavior of the eigenvalues \( \lambda_1, \lambda_2 \), and \( \lambda_3 \) structures can be described.
\begin{figure*}[h!]
  \centering
   \includegraphics[width=0.55\textwidth]{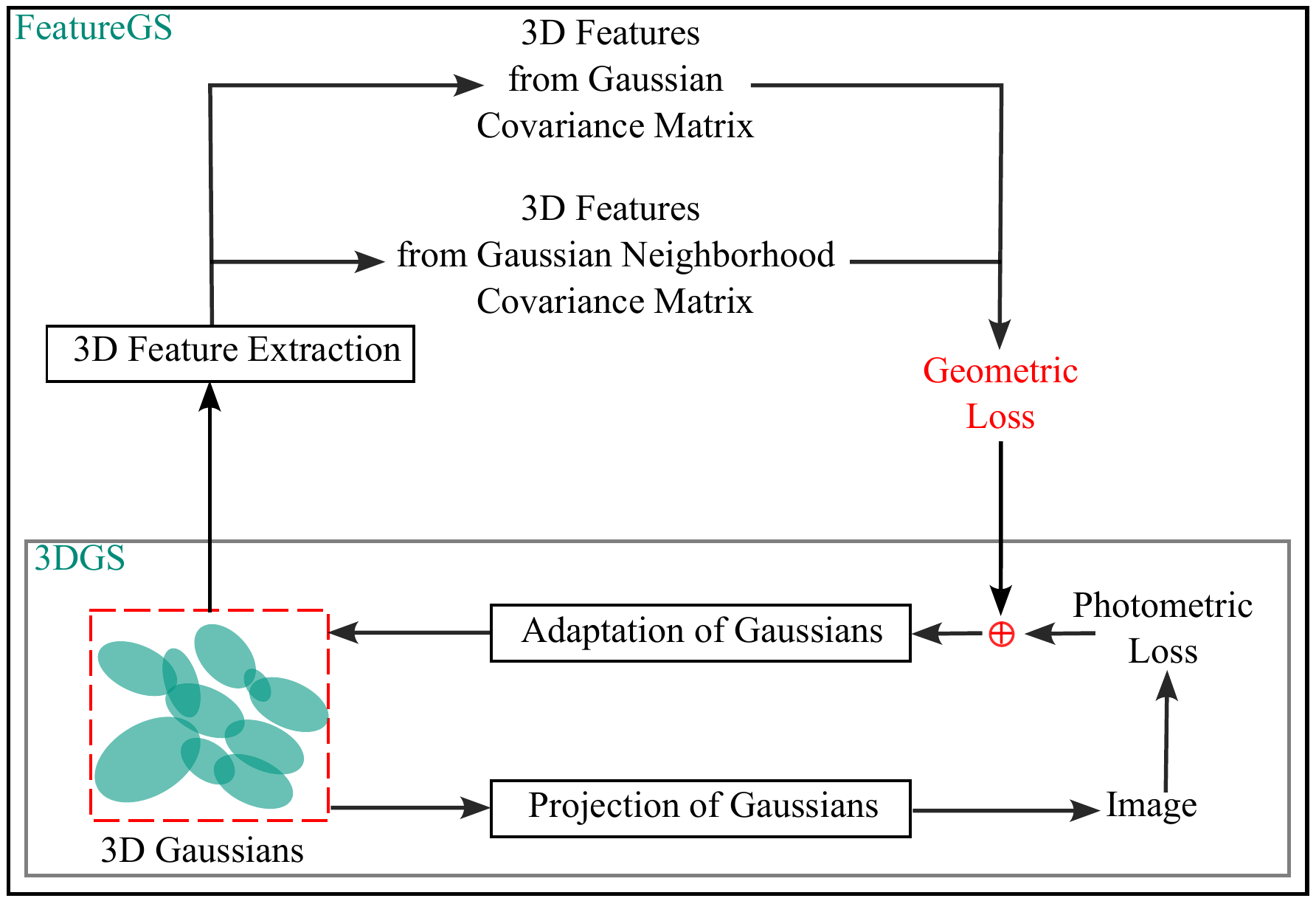}
   \caption{Methodology of FeatureGS: Geometric loss based on 3D shape features added to 3DGS \cite{3DGS}. The features are derived from eigenvalues of the covariance matrix from individual Gaussians or the covariance matric from Gaussians in a local neighborhood surrounding each Gaussian center. The geometric loss is combined with the photometric loss in 3DGS.}
   \label{fig:FeatureGS}
\end{figure*}
FeatureGS aims to improve geometric accuracy of Gaussians and enhance properties of planar surfaces with a reduced structural entropy in local 3D neighborhoods of Gaussians.
Firstly, like previous flattening approaches \cite{SuGaR, Surfels, PGSR, 2DGS}, FeatureGS aims to flatten 3D Gaussians by enhancing the 'planarity' of Gaussians as 3D feature, in order to achieve higher geometric accuracy of Gaussian centers. 
Secondly, real physical circumstances of point clouds can be described by interpretable geometric features with a single value \cite{hillemann2019automatic}. To enhance the structural representation of the 3D Gaussian centers in a neighborhood, particularly for manmade objects aligning with Manhattan-Word-Assumption \cite{coughlan1999manhattan, coughlan2000manhattan}, we leverage neighborhood-based 3D features derived from the k-nearest neighbors (kNN) of each Gaussian. By incorporating the 3D features either 'planarity', 'omnivariance', or 'eigenentropy' in the geometric loss, the characterization of local 3D structures with a predominance of planar surfaces with a structural entropy is reinforced.

We investigate whether integrating of different geometric loss terms of FeatureGS can enhance the 3D geometric accuracy of Gaussian centers and suppress floater artifacts by reinforcing specific 3D shape properties of Gaussians and Gaussian neighborhoods. The evaluation focuses on the Chamfer cloud-to-cloud distance for geometrically 3D accuracy and artifact-reduction, and the total number of Gaussians required to represent the scene for memory efficiency. While our primary goal is to achieve precise geometric representation and efficient memory usage, we additionally report the rendering quality, measured by Peak Signal-to-Noise Ratio (PSNR), to ensure consistency in scene reconstruction. Experiments are conducted on 15 scenes from the DTU benchmark dataset.

We demonstrate that FeatureGS strikes a remarkable balance between geometric accuracy, floater artifact suppression, and memory efficiency by integrating 3D shape feature properties into the optimization process of 3D Gaussian Splatting. FeatureGS improves geometric accuracy, enabling the Gaussian centers to serve as a more precise geometric representation. Furthermore, FeatureGS reduces the total number of Gaussians required to represent a scene for the same rendering quality as 3DGS. The resulting 3D scene reconstructions with high-accurate Gaussian centers for the geometric representation are both artifact-reduced and memory-efficient.

\section{Related Work}
\label{sec:RelatedWork}
In this section, we provide an overview of the different types of 3D features in Section \ref{sec:3D_Features}, which are essential for FeatureGS. Subsequently, in Section \ref{sec:RelatedWork_a}, we present a brief overview of novel view synthesis and 3D reconstructions, followed by an introduction to 3D reconstructions with Gaussian splats in Section \ref{sec:RelatedWork_c}.

\subsection{3D Features}
\label{sec:3D_Features}

Several types of 3D features exist for point cloud-based applications such as classification, registration, or calibration.
Complex features, which can not be interpreted directly are descriptors such as Shape Context 3D (SC3D) \cite{SC3D}, Signature of Histogram of OrienTations (SHOT) \cite{SHOT} or Fast Point Feature Histograms (FPFH) \cite{FPFH}. In contrast, interpretable features are those that are directly interpretable, such as local 2D and 3D shape features. 
To describe the local structure around a 3D point, the spatial arrangement of other 3D points in the local neighborhood is often considered. Thereby the 3D covariance matrix, also known as the 3D structure tensor, is well-known and suitable for characterizing the shape properties of 3D data \cite{Weinmann_2015}. It is derived explicitly for each point from the point itself and its local neighbors. The three eigenvalues, \( \lambda_1 \geq \lambda_2 \geq \lambda_3 \geq 0 \), correspond to an orthogonal system of eigenvectors (\( \epsilon_1, \epsilon_2, \epsilon_3 \)), which indicate the direction (\textit{rotation}) of the three ellipsoid principal axes and correspond to the extent (\textit{scales}) of the 3D ellipsoid along the principal axes. Based on the behavior of the eigenvalues \( \lambda_1, \lambda_2 \), and \( \lambda_3 \), linear (\( \lambda_1 \gg \lambda_2, \lambda_3 \)), planar (\( \lambda_1 \approx \lambda_2 \gg \lambda_3 \)), and spherical (\( \lambda_1 \approx \lambda_2 \approx \lambda_3 \)) structures can be described.
The use of geometric 3D shape features has led to thousands of publications in various fields over the past few decades.
They are especially used for the automatic semantic interpretation and classification \cite{Weinmann_2015, Weinmann_2017_Feature_Relevance, weinmann20203d} of point clouds. But also for calibration \cite{hillemann2019automatic} or registration \cite{pointcloudregistration} of 3D point clouds.

\subsection{Novel View Synthesis and 3D Reconstruction}
\label{sec:RelatedWork_a}

The pioneering research on Neural Radiance Fields (NeRFs) \cite{mildenhall_et_al_2020} builds upon Scene Representation Networks \cite{SceneRepresenationNetworks}, which represent the scene as a function of 3D coordinates within the scene. NeRFs extend this concept by estimating color values and densities for each 3D coordinate through 6D camera positions and associated 2D images by training a multilayer perceptron (MLP).
NeRF was followed by thousands of publications driving research and development of neural surface reconstructions, point cloud and mesh reconstruction \cite{unisurf, neus, volsdf, neuralangelo, densitygradient} in various domains.
However, NeRF describes the scene implicitly by estimating a color and volume density for each position and direction, which are also subject to a certain degree of uncertainty \cite{uncertainty}.

\subsection{3D Reconstruction with Gaussian Splats}
\label{sec:RelatedWork_c}
In contrast 3D Gaussian Splatting (3DGS) \cite{3DGS} offers new possibilities for 3D scene reconstruction. 
With 3DGS a novel concept of 3D scene representation was elaborated, in which a scene is explicitly represented by a large set of 3D Gaussians.
Each Gaussian is defined by its mean, covariance, opacity, and spherical harmonics for color definition. The covariance is parameterized using scaling and rotation. These 3D Gaussians are projected into 2D Gaussians to the 2D image space, allowing fast rendering.
To optimize the scene, the Gaussians are initialized from a point cloud produced by Structure from Motion (SfM). The Gaussians' parameters (means for the Gaussian centers, scaling, rotations, opacities, and color) are then refined during optimization to match the training images. More Gaussians are added as needed to improve the scene representation. This optimization process leads to scenes with millions of small Gaussians that represent the 3D object geometry. 
Nevertheless, Gaussians do not take an ordered structure in general \cite{SuGaR}, and the center or surface of a Gaussian does not directly align with the actual object surface. In addition, 3DGS often leads to floater artifacts, which further increase the high number of Gaussians and thus the storage requirements.

The concept of transforming 3D Gaussians into 2D ellipses or planar ellipse-like structures in order to achieve higher geometric accuracy is widely used in many approaches.
SuGaR \cite{SuGaR} extracts meshes from 3DGS by introducing a regularization term that aligns Gaussians with the scene surface.
Surfels \cite{Surfels} combines 3D Gaussian points' optimization flexibility with the surface alignment of surfels by flattening 3D Gaussians into 2D ellipses, setting the z-scale to zero. 
PGSNR \cite{PGSR} flattens Gaussians into planes, using unbiased depth rendering to obtain precise depth information.
2DGS \cite{2DGS} follows a similar approach and collapses 3D volumes directly into 2D planar Gaussian disks for view-consistent geometry, using perspective-accurate splatting with ray-splat intersection and depth and normal consistency terms. 
MVG-Splatting \cite{MVG_Splatting} improves 2DGS by optimizing normal calculation and using an adaptive densification method guided by depth maps.
MIP-Splatting \cite{Mip-Splatting} introduces a 3D smoothing filter to constrain Gaussian sizes based on the input views' sampling frequency, eliminating high-frequency artifacts.

In contrast to other works, FeatureGS utilizes geometric 3D shape features to enhance specific Gaussian and Gaussian neighborhood properties for geometrically accurate and artifact-reduced 3D reconstruction. The 3D features are embedded within the optimization process of 3DGS through an additional geometric loss term in four alternative formulations into a photometric-geometric loss term.
On the one hand, like previous approaches \cite{SuGaR, Surfels, PGSR, 2DGS}, FeatureGS flattens 3D Gaussians. However, FeatureGS incorporates the 3D feature 'planarity' for that.
On the other hand, FeatureGS can enhance properties of planar surfaces with reduced structural entropy by utizling 3D features 'planarity', 'omnivariance' and 'eigenentropy' in Gaussian neighborhoods.

\section{Methodology}
\label{sec:Methodology}

In this section, we describe FeatureGS (Figure \ref{fig:FeatureGS}) with an additional geometric loss term based on 3D shape features. These features are derived from the eigenvalues of the covariance matrix and provide insights into the spatial structure within both individual Gaussians and the Gaussians in a local neighborhood surrounding each Gaussian center. Our proposed geometric loss is combined with the photometric loss used in 3DGS to create a comprehensive photometric-geometric loss function.

\subsection{Photometric Loss}
The photometric loss term of 3DGS measures the similarity between rendered images and ground truth images using pixel-level comparison metrics. This loss includes both L1 loss and a Structural Similarity Index (SSIM) term to capture differences in luminance, contrast, and structure between the images.
The photometric loss is given by the following loss function \ref{equ:loss}.
\begin{equation}\label{equ:loss}
L_{\text{photometric}} = (1 - \theta) L_{\text{1}} + \theta L_{\text{D-SSIM}} \\
\end{equation}
with $\theta$, $L_{\text{1}}$-Norm of the per pixel color difference and $L_{\text{D-SSIM}}$-Term \cite{3DGS}.

\subsection{Geometric Loss}

\begin{figure}[h!]
  \centering
   \includegraphics[width=1.0\linewidth]{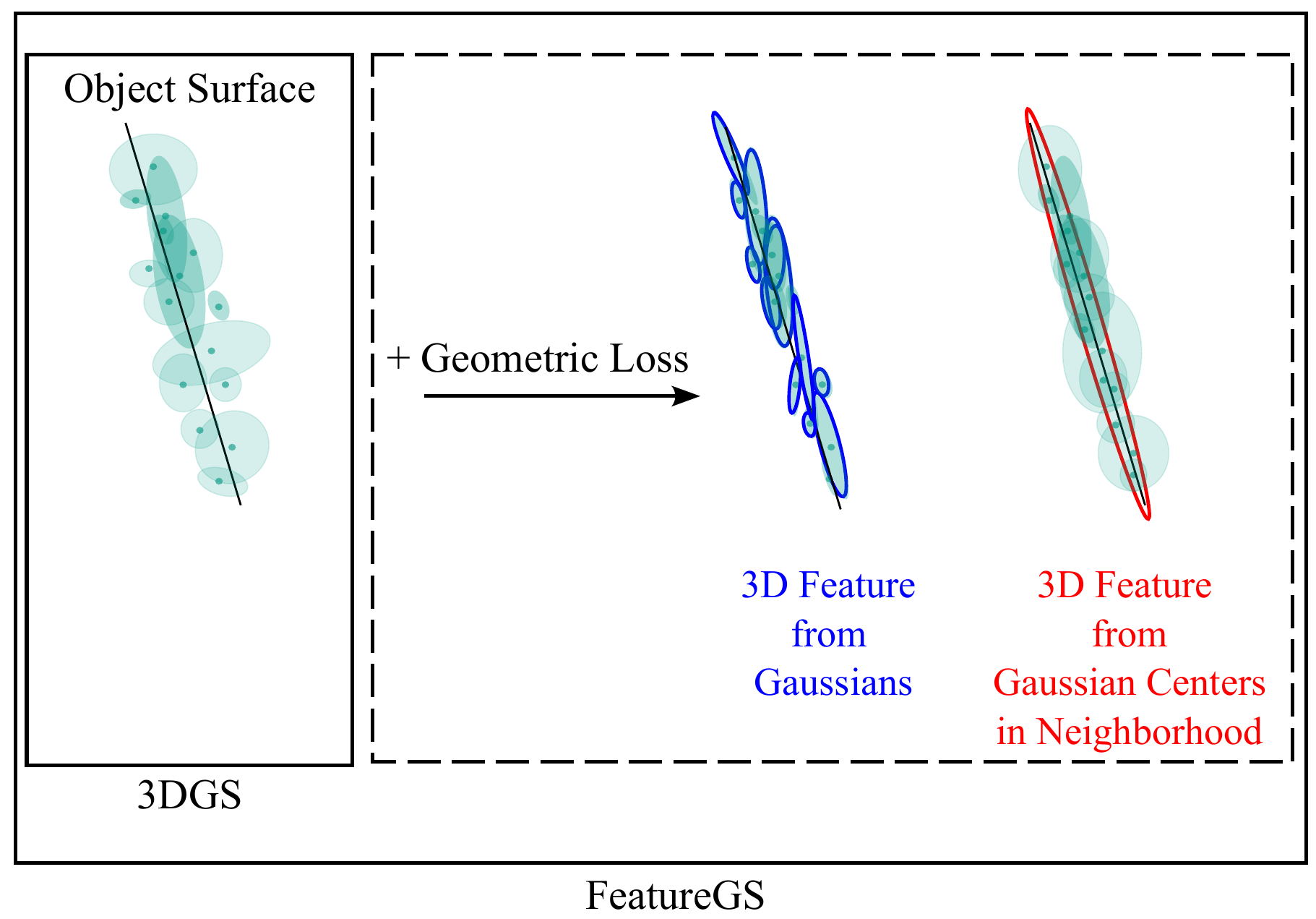}
   \caption{Additional geometric loss of FeatureGS, illustrated by a 3D feature from Gaussians and a 3D feature from Gaussian centers in a local neighborhood. For example, through the loss with planarity of Gaussians, the Gaussians become more planar, and through the loss with planarity, omnivariance, or eigenentropy in the Gaussian neighborhood, the alignment of the Gaussian centers in the neighborhood becomes more planar. All configurations have the effect that the Gaussians move closer to the object surface and are less randomly oriented. This enables the Gaussian centers to serve as a geometric representation of the surface.}
   \label{fig:geometric_loss}
\end{figure}

We introduce four different novel additional geometric loss terms, based on the eigenvalue-derived 3D shape features to enhance specific properties (see Figure \ref{fig:geometric_loss}) of 3D Gaussian itself and Gaussian centers in a neighborhood.
For the first approach, we aim to flatten Gaussians to achieve a high geometric accuracy of the Gaussian centers. This is done by incorporating the 3D shape feature 'planarity' from eigenvalues (scales) (Figure \ref{fig:gaussian_ellipsoid}) of each Gaussian itself in the geometric loss term.
For the second approach, we incorporate a neighborhood-based geometric loss term using the 3D shape features from covariance matrix (Figure \ref{fig:neighborhood_ellipsoid}) by the \(k\)-nearest neighbors (kNN) of each Gaussian center. To enhance a specific characterization of local 3D structures of manmade objects aligning with Manhattan-Word-Assumption \cite{coughlan1999manhattan, coughlan2000manhattan}, we strengthen the predominance of planar surfaces, and a structural entropy. This is done through the Gaussian neighborhood 3D shape features 'planarity', 'omnivariance', and 'eigenentropy'.

\subsubsection{Covariance Matrix}

\begin{figure}[h!]
    \centering
    \begin{subfigure}[]{0.23\textwidth}
        \centering
        \includegraphics[width=\textwidth]{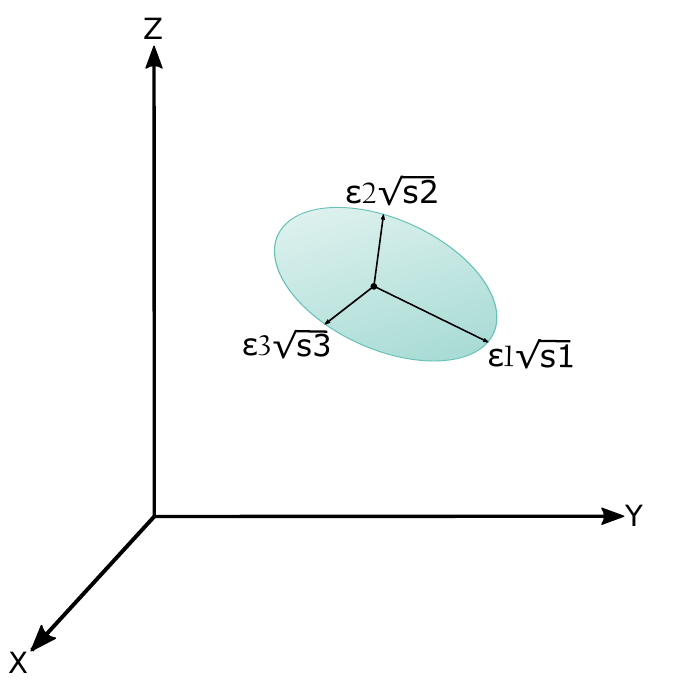}
        \caption{}
        \label{fig:gaussian_ellipsoid}
    \end{subfigure}
    \begin{subfigure}[]{0.23\textwidth}
        \centering
        \includegraphics[width=\textwidth]{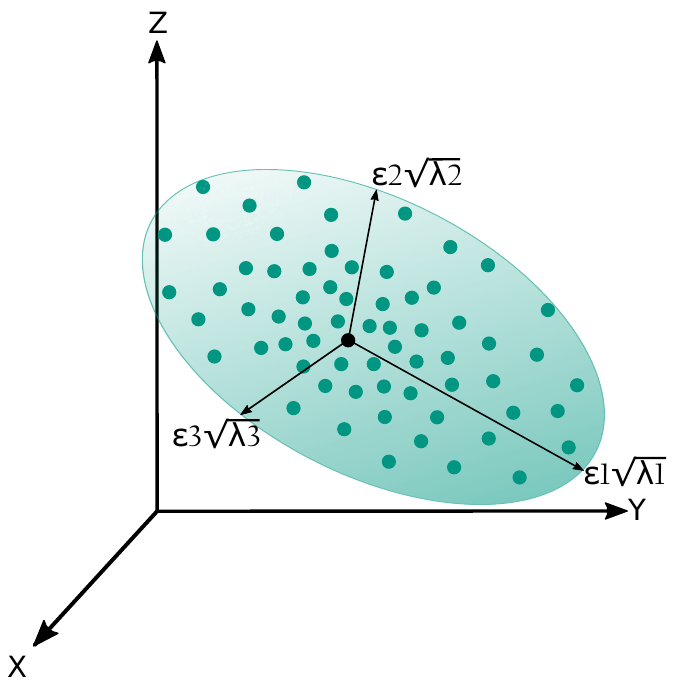}  \caption{}
        \label{fig:neighborhood_ellipsoid}
    \end{subfigure}
    \caption{\ref{fig:gaussian_ellipsoid} Representation of a single Gaussian ellipsoid with the three eigenvectors (\(\mathbf{\epsilon_1}, \mathbf{\epsilon_2}, \mathbf{\epsilon_3}\)) and the corresponding eigenvalues (\(s_1, s_2, s_3\)) in the three-dimensional coordinate system.
    \ref{fig:neighborhood_ellipsoid} Representation of an ellipsoid from the neighborhood points represented by the Gaussian centers with the three eigenvectors (\(\mathbf{\epsilon_1}, \mathbf{\epsilon_2}, \mathbf{\epsilon_3}\)) and the corresponding eigenvalues (\(\lambda_1, \lambda_2, \lambda_3\)) in the three-dimensional coordinate system.}
    \label{fig:ellisoids}
\end{figure}

\paragraph{Gaussian.}
3DGS uses an explicit representation of the scene through 3D Gaussians. These ellipsoid-like structures are characterized by scaling, rotation, and color, including opacity. The scaling components can be interpreted analogously to the three eigenvalues \( s_1 \geq s_2 \geq s_3 \geq 0 \) and the rotation components to the eigenvectors (\( \epsilon_1, \epsilon_2, \epsilon_3 \)) of the covariance matrix.
By using the normalized eigenvalues (scales) of the Gaussian covariance matrix (Figure \ref{fig:gaussian_ellipsoid}), we compute the 3D shape feature.

\paragraph{Gaussian Neighborhood.}

Given a point \( p_0\) in the 3D space, i.e., the center of a Gaussian, we define its \(k\)-nearest neighbors \(\{p_1, p_2, \ldots, p_k\}\). The centroid \(\bar{p}\) (Equation \ref{equ:p}) of this neighborhood is computed as:

\begin{equation}\label{equ:p}
\bar{p} = \frac{1}{k+1} \sum_{i=0}^{k} p_i
\end{equation}

The covariance matrix \(C\) (Equation \ref{equ:C}) \cite{weinmann_covariance} for the neighborhood (Figure \ref{fig:neighborhood_ellipsoid}) is then:

\begin{equation}\label{equ:C}
C = \frac{1}{k+1} \sum_{i=0}^{k} (p_i - \bar{p})(p_i - \bar{p})^T
\end{equation}

From \(C\), eigenvalues \(\lambda_1 \geq \lambda_2 \geq \lambda_3\) are derived, providing shape properties for the neighborhood.

\subsubsection{Eigenvalue Normalization}
To ensure consistency, eigenvalues \(s_1, s_2, s_3\) from the Gaussian covariance matrix and \(\lambda_1, \lambda_2, \lambda_3\) from the Gaussian neighborhood covariance matrix are normalized by dividing by the sum of the eigenvalues for each case.

For the Gaussian covariance matrix:
\begin{equation}
s'_i = \frac{s_i}{\text{sum}(\mathbf{s})} \quad \text{for} \enspace i \in \{1, 2, 3\} \enspace \text{with} \enspace \text{sum}(\mathbf{s}) = \sum_{i=1}^3 s_i.
\end{equation}

For the Gaussian neighborhood covariance matrix:
\begin{equation}
\lambda'_i = \frac{\lambda_i}{\text{sum}(\mathbf{\lambda})} \quad \text{for} \enspace i \in \{1, 2, 3\} \enspace \text{with} \enspace \text{sum}(\mathbf{\lambda}) = \sum_{i=1}^3 \lambda_i.
\end{equation}

The normalized eigenvalues \(s'_1, s'_2, s'_3\) and \(\lambda'_1, \lambda'_2, \lambda'_3\) are then ordered in descending order:
\[
s'_1 \geq s'_2 \geq s'_3 \geq 0 \quad \text{and} \quad \lambda'_1 \geq \lambda'_2 \geq \lambda'_3 \geq 0.
\]

The normalized eigenvalues are then used for the final geometric 3D feature computation.

\subsubsection{Geometric Loss with Gaussians}

\paragraph{Gaussian Planarity.} Planarity measures the extent to which a Gaussian resembles a planar structure. It is defined as:


\begin{equation}
\text{Planarity}_{\text{Gaussian}} = \frac{s'_2 - s'_3}{s'_1}
\end{equation}

The planarity Gaussian loss, preferring high planarity similar to other flattening approaches, is:

\begin{equation}
L_\text{Planarity, Gaussian} = \left( 1 - \frac{s'_2 - s'_3}{s'_1} \right)
\end{equation}

\subsubsection{Geometric Loss with Gaussian Neighborhoods}
To enhance the structural properties that 3D point clouds of manmade objects exhibit, we incorporate a neighborhood-based geometric loss using the \(k\)-nearest neighbors (kNN) of each point. This approach allows for the calculation of spatial features in the local neighborhood of each Gaussian. The strengthening of the characterization of planar surfaces with reduced structural entropy in local 3D neighborhoods is achieved by including a geometric neighborhood loss. For this purpose, we consider the 3D features $\text{Planarity}_{\text{kNN}}$, $\text{Omnivariance}_{\text{kNN}}$ and $\text{Eigenentropy}_{\text{kNN}}$ in Gaussian neighborhoods of kNN from the normalized eigenvalues \(\lambda'_1 \geq \lambda'_2 \geq \lambda'_3 \geq 0\), explained in more detail below.

\paragraph{Neighborhood Planarity.} Similar to the purpose of maintaining planarity of each Gaussian itself, we want to strengthen the properties of manmade objects according to the Manhattan-Word-Assumption \cite{coughlan1999manhattan, coughlan2000manhattan} and other (almost) planar surfaces, and suppress the spherical spread of the Gaussians in a neighborhood. Therefore, in addition to the planarity of the Gaussians, we use the planarity in the neighborhood. This is defined as:

\begin{equation}
\text{Planarity}_{\text{kNN}} = \frac{\lambda'_2 - \lambda'_3}{\lambda'_1}
\end{equation}
  
The neighborhood planarity loss is:

\begin{equation}
L_\text{Planarity, kNN} = \left( 1 - \frac{\lambda'_2 - \lambda'_3}{\lambda'_1} \right)
\end{equation}

\paragraph{Neighborhood Omnivariance.} The Omnivariance indicates the volume of the neighborhood and expresses whether the respective points scatter locally in all directions. In previous work \cite{Weinmann_2015_Omnivariance}, omnivariance was shown to be a highly relevant feature of point cloud classification. Omnivariance and the neighborhood omnivariance loss are defined as:

\begin{equation}
L_\text{Omnivariance, kNN} = \text{Omnivariance}_{\text{kNN}} = \sqrt[3]{\lambda'_1 \lambda'_2 \lambda'_3}
\end{equation}
  
Minimizing the neighborhood omnivariance loss reduces the local scattering of the points.

\paragraph{Neighborhood Eigenentropy.} The Eigenentropy quantifies the order/disorder of the local structure of the neighborhood points by measuring the entropy within the local 3D neighborhood based on the normalized eigenvalues. Additionally, it has shown to be a good 3D feature for characterizing plane point cloud structures \cite{Dittrich}.  Eigenentropy and the neighborhood eigenentropy loss is defined as:

\begin{equation}
L_\text{Eigenentropy, kNN} = \text{Eigenentropy}_{\text{kNN}} = - \sum_{i=1}^{3} \lambda'_i \log(\lambda'_i)
\end{equation}
  
Minimizing the neighborhood eigenentropy loss favors a minimum disorder \cite{Weinmann_2017_Feature_Relevance} and therefore low entropy of 3D points.

\subsection{Combined Photometric-Geometric Loss} 
Our four different final loss functions $L$ combine the conventional photometric loss $L_{\text{photometric}}$ of 3DGS with each one of four different geometric loss $L_{\text{geometric}}$ terms. This incorporates both the 3D shape properties of each Gaussian itself $L_{\text{geometric, Gaussian}}$ or the neighborhood features $L_{\text{geometric, kNN}}$ based on Gaussian centers. The photometric loss ensures the quality of pixel rendering by adjusting the Gaussians according to their projection onto the image plane, while the geometric loss term enhance specific properties of 3D structures.
The total photometric-geometric loss $L$ (Equation \ref{eq:loss_total}) is defined as:

\begin{equation} L = h_{\text{photo}} \cdot L_{\text{photometric}} + L_{\text{geometric}}, \label{eq:loss_total} \end{equation}

with

\begin{equation}
\begin{split}
L_{\text{geometric}}\in 
\{L_\text{Planarity, Gaussian}, L_\text{Planarity, kNN}, \\L_\text{Omnivariance, kNN}, L_\text{Eigenentropy, kNN}\}. 
\label{eq}
\end{split}
\end{equation}

\section{Experiments}
\label{sec:Experiments}

\subsection{Dataset}
For the evaluation of FeatureGS, we use the DTU benchmark dataset \cite{dtu}. The dataset consists of scenes featuring real objects, including either 49 or 64 RGB images, corresponding camera poses, and reference point clouds obtained from a structured-light scanner (SLS). We specifically focus on the same 12 scenes as as previous approaches \cite{MVG_Splatting, Surfels, PGSR, 2DGS}.

\subsection{Metrics}

To evaluate our method, we report the 3D geometric accuracy, the number of Gaussians needed to represent the scene for memory efficiency and the rendering quality. For 3D evaluation we report the Chamfer cloud-to-cloud distance. To evaluate surface accuracy, we use the DTU evaluation procedure \cite{dtu}, which masks out points above 10\,mm. In addition, we use the Chamfer cloud-to-cloud distance for all points to evaluate the presence of floater artifacts external to the object. Low Chamfer distance indicates high accuracy and less artifacts. We evaluate the 2D rendering quality of the images with the Peak Signal-to-Noise Ratio (PSNR) in dB, whereby a high PSNR is targeted.

\subsection{Implementation Details}

3D Gaussian Splatting is processed according to the original implementation, using default parameters with learning rates of 0.0025 for spherical harmonics features, 0.05 for opacity adjustments, 0.005 for scaling operations and 0.001 for rotation transformations, on a NVIDIA RTX3090 GPU.

Firstly we consider the same number of training iterations of 15\,000, which are recommended from 3DGS, and evaluate on the geometric accuracy by Chamfer Cloud-to-Cloud distance, Gaussian storage requirements by the total numbers of Gaussian as well as photometric quality by PSNR. This is done quantitatively and qualitatively.
For a fair comparison, we consider the evaluation procedure by training with early-stopping on each the same reached PSNR value.
This should demonstrate that FeatureGS enables the same photometric rendering quality by pushing down the total numbers of Gaussians representing the scene, while also achieving higher geometric accuracy and artifact-reduced rendering.

\subsection{Loss configurations}

The photometric loss for the optimization is given by the loss function in Equation \ref{equ:loss} with $\theta$ = 0.2 by default \cite{3DGS}.

For the final different photometric-geometric loss formulations (Equation \ref{eq:loss_total}) of FeatureGS the weighting with $h_{\text{photo}} = 0.05$ is chosen. This is based on hyperparameter tuning, see Figure \ref{fig:h_c2c_psnr}, to create a proper balance between rendering quality and geometric accuracy, approximately where the PSNR remains the same but the Chamfer cloud-to-cloud distance increases.

As the 3D distribution of the Gaussians and hence their centers changes through the optimization, we decide on a fixed number of $\text{kNN} = 50$ \cite{Weinmann_2017_Feature_Relevance} nearest neighbors. Through the variable distribution and density of the points during the training process, we aim to achieve an effect similar to multi-scale \cite{multiscale_knn} neighborhoods, which have proven to be robust in point cloud classification tasks.

\section{Results}
\label{sec:Experiments}
The following sections show qualitative (Section \ref{sec:quantitative}) and quantitative (Section \ref{sec:qualitative}) results of FeatureGS in comparison to 3DGS. We distinguish between the training with a fixed number of training iterations in Section \ref{sec:quantitative_iterations} and a fixed achievable rendering quality in Section \ref{sec:quantitative_rendering}, represented by an early-stopping of the PSNR. This should demonstrate the performance of FeatureGS in terms of geometric accuracy, floater artifact-reduction, memory efficiency, and yet strong rendering quality, based on the two criteria.

\subsection{Quantitative Results}\label{sec:quantitative}

\subsubsection{Training Process}\label{sec:training}
Over the training process of 15\,000 iterations, the original photometric loss of 3DGS and the photometric-geometric loss of FeatureGS demonstrate distinct behaviors in terms of geometric accuracy, presence of floater artifacts, number of Gaussians representing the scene, and rendering quality.

\begin{figure}[h!]
  \centering
   \includegraphics[width=1.0\linewidth]{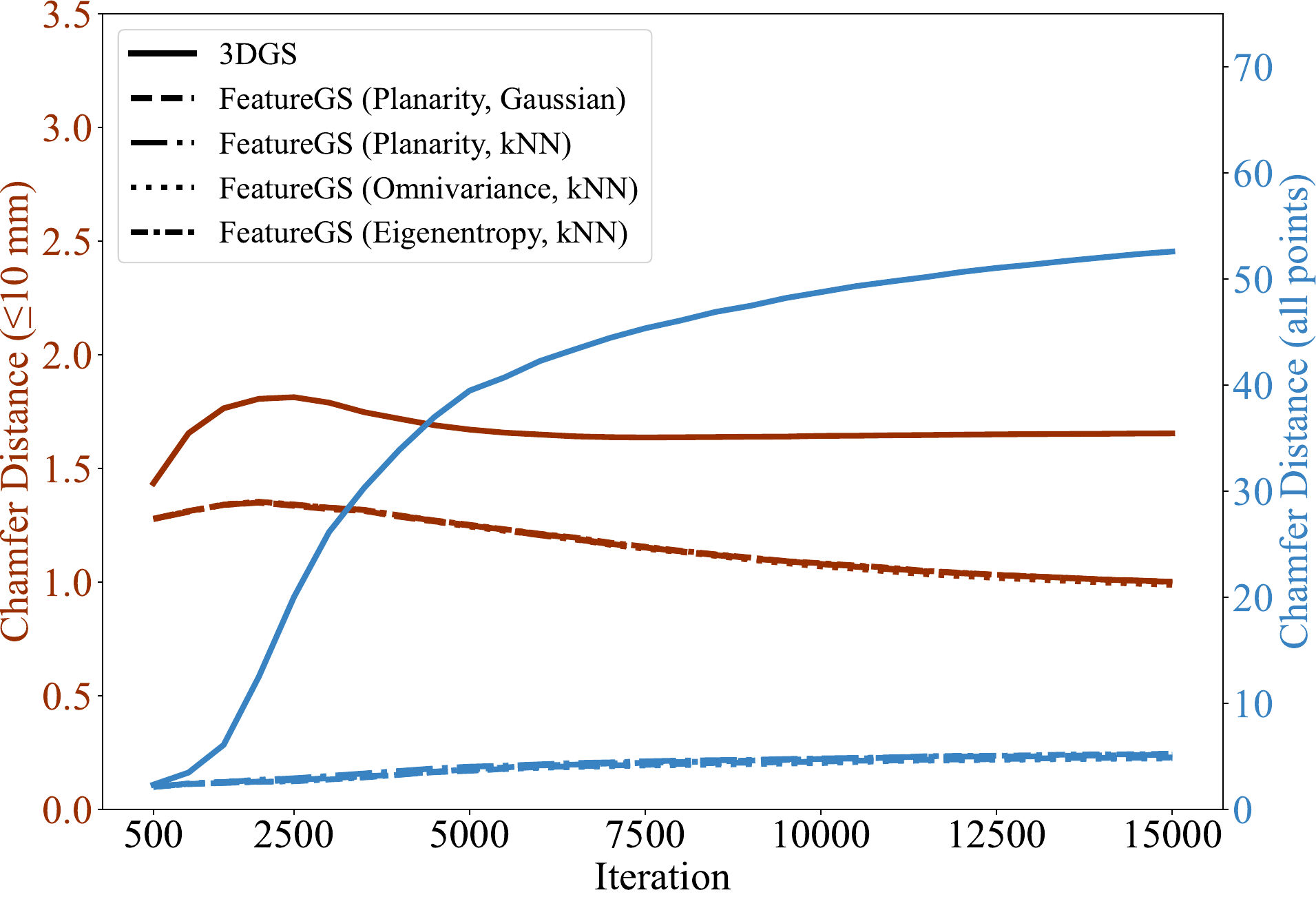}
   \caption{\textbf{Geometric accuracy} during training process on the DTU scan40. Chamfer cloud-to-cloud distances $\downarrow$ in mm for points $\leq$10\,mm and all points (floater artifacts). The curves show for the different loss types.}
   \label{fig:c2c_vs_iterations}
\end{figure}

It is observed that the Chamfer cloud-to-cloud distance (Figure \ref{fig:c2c_vs_iterations}) for 3DGS continuously increases for all points during training process. For instance, in the case of scene40, the distance rises to approximately 50\,mm. In contrast, for all geometric FeatureGS losses, the distance remains consistently low throughout the training process. Only a slight increase is present, which is due to the fact that the initial point cloud from SfM nonetheless has the highest accuracy and FeatureGS also reconstructs points that are not in the (incomplete) reference point cloud.
For scan40, this distance stabilizes at around 4–5\,mm. This indicates that, unlike FeatureGS, the 3DGS training process incorporates a significant number of points with higher geometric inaccuracies. Regarding the geometric surface accuracy, measured by masking out points with errors over 10\,mm, the distance for 3DGS initially increases to approximately 1.9\,mm, then decreases and stabilizes at a constant value. For scan40, the distance starts at approximately 1.2\,mm, peaks at 1.9 mm, and eventually stabilizes at 1.7\,mm. Conversely, the distance for FeatureGS rises less at the start of training and then decreases further as training progresses. For scan40, it decreases from approximately 1.3\,mm to 1.0\,mm.

With regard to rendering quality (Figure \ref{fig:psnr_vs_iterations}), as measured by PSNR (and SSIM), the original photometric loss of 3DGS significantly outperforms the combined photometric-geometric loss of FeatureGS. For 3DGS, the PSNR continuously increases and appears to converge after approximately 14\,000 training iterations. In contrast, for all FeatureGS loss functions, the PSNR initially increases rapidly but saturates at a noticeably lower value after about 10\,000 iterations. The behavior of SSIM follows a similar trend.

\begin{figure}[h!]
  \centering
   \includegraphics[width=1.0\linewidth]{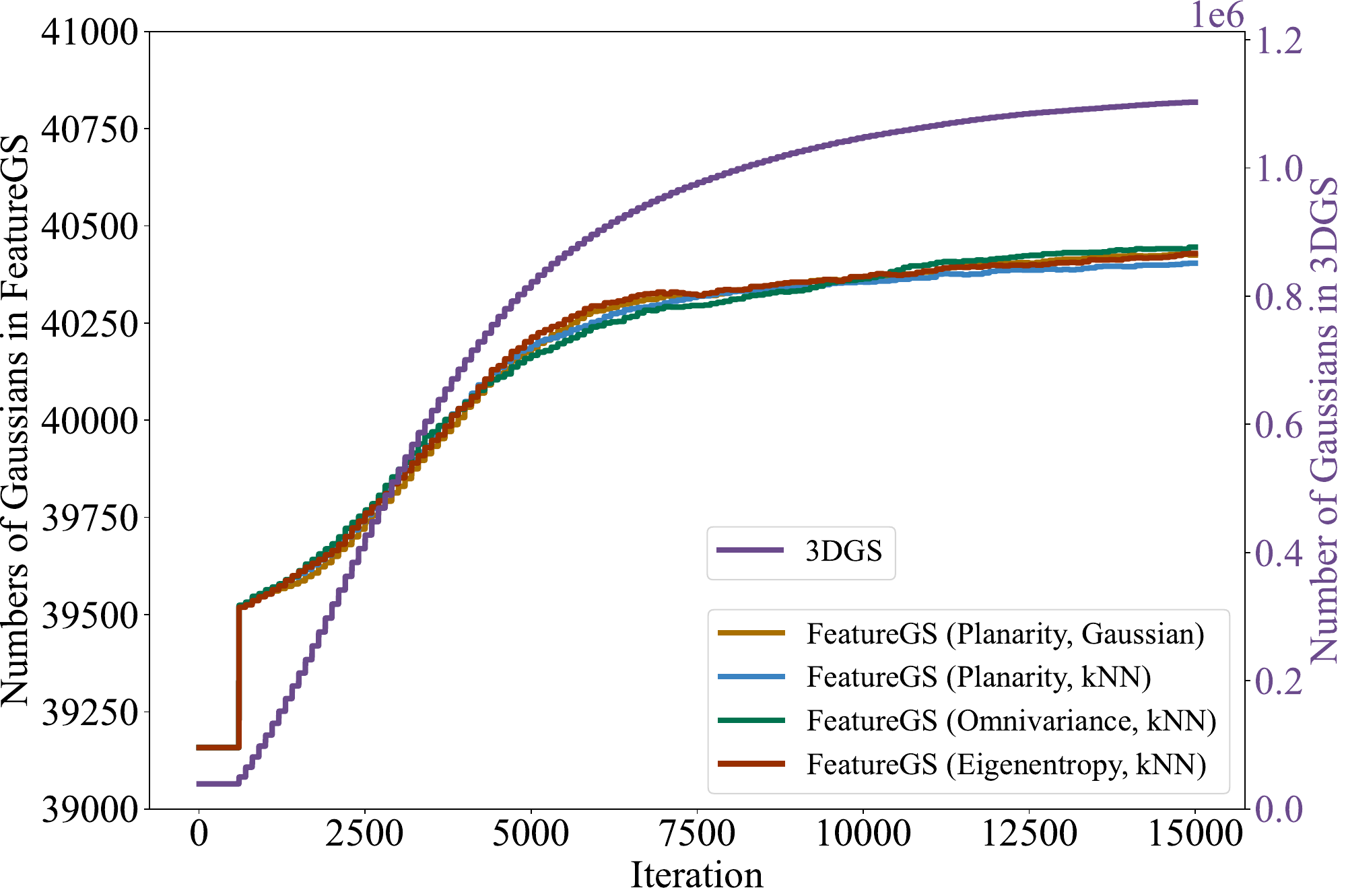}
   \caption{\textbf{Numbers of Gaussians} during training process on the DTU scan40. The curves show for the different loss types.}
   \label{fig:psnr_vs_iterations}
\end{figure}

\begin{figure}[h!]
  \centering
   \includegraphics[width=1.0\linewidth]{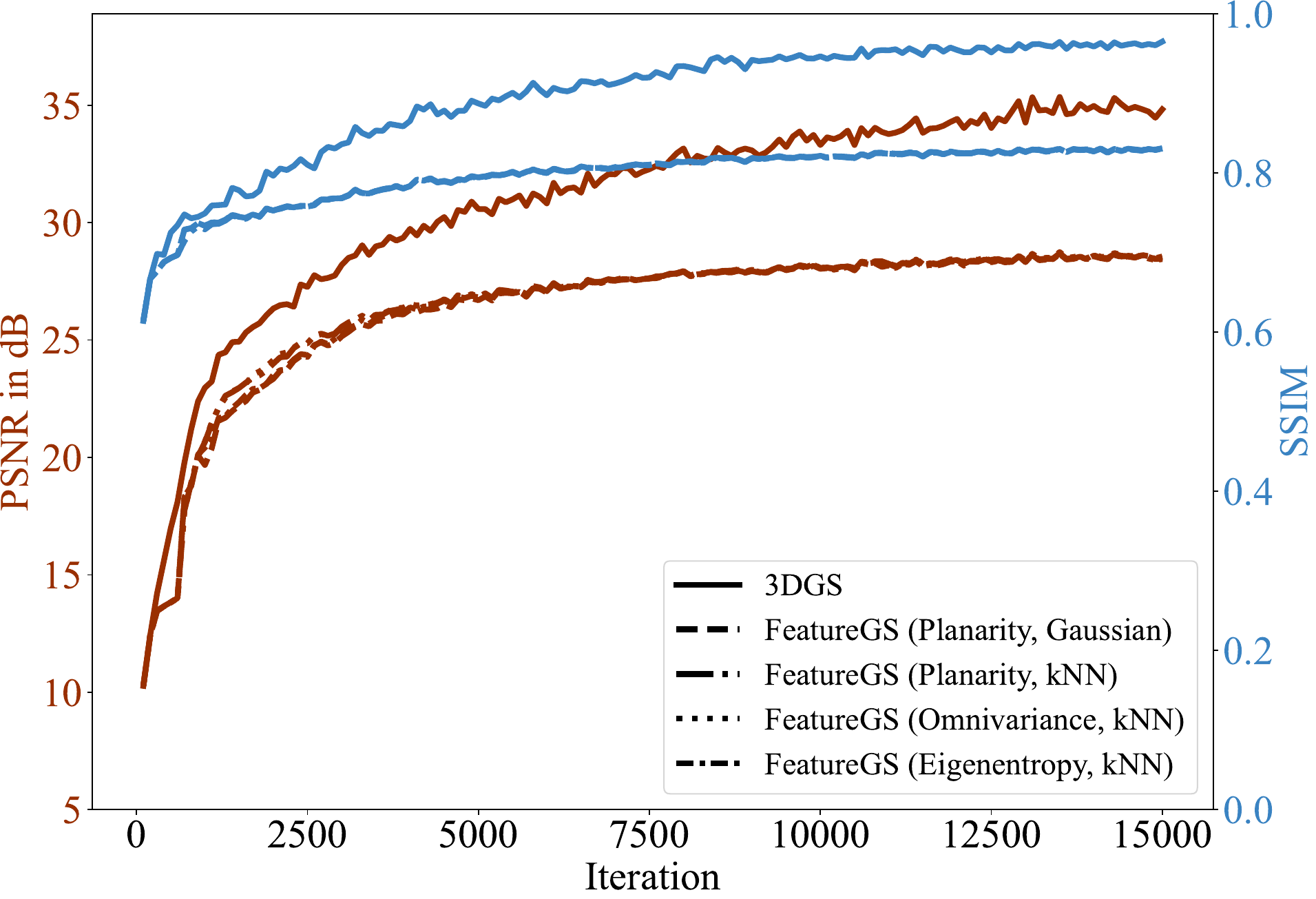}
   \caption{\textbf{Rendering quality} during training process on the DTU scan40. Peak Signal-to-Noise Ratio (PSNR) $\uparrow$ in dB and SSIM $\uparrow$. The curves show for the different loss types.}
   \label{fig:psnr_vs_iterations}
\end{figure}

\begin{figure*}[h!]
  \centering
   \includegraphics[width=0.7\linewidth]{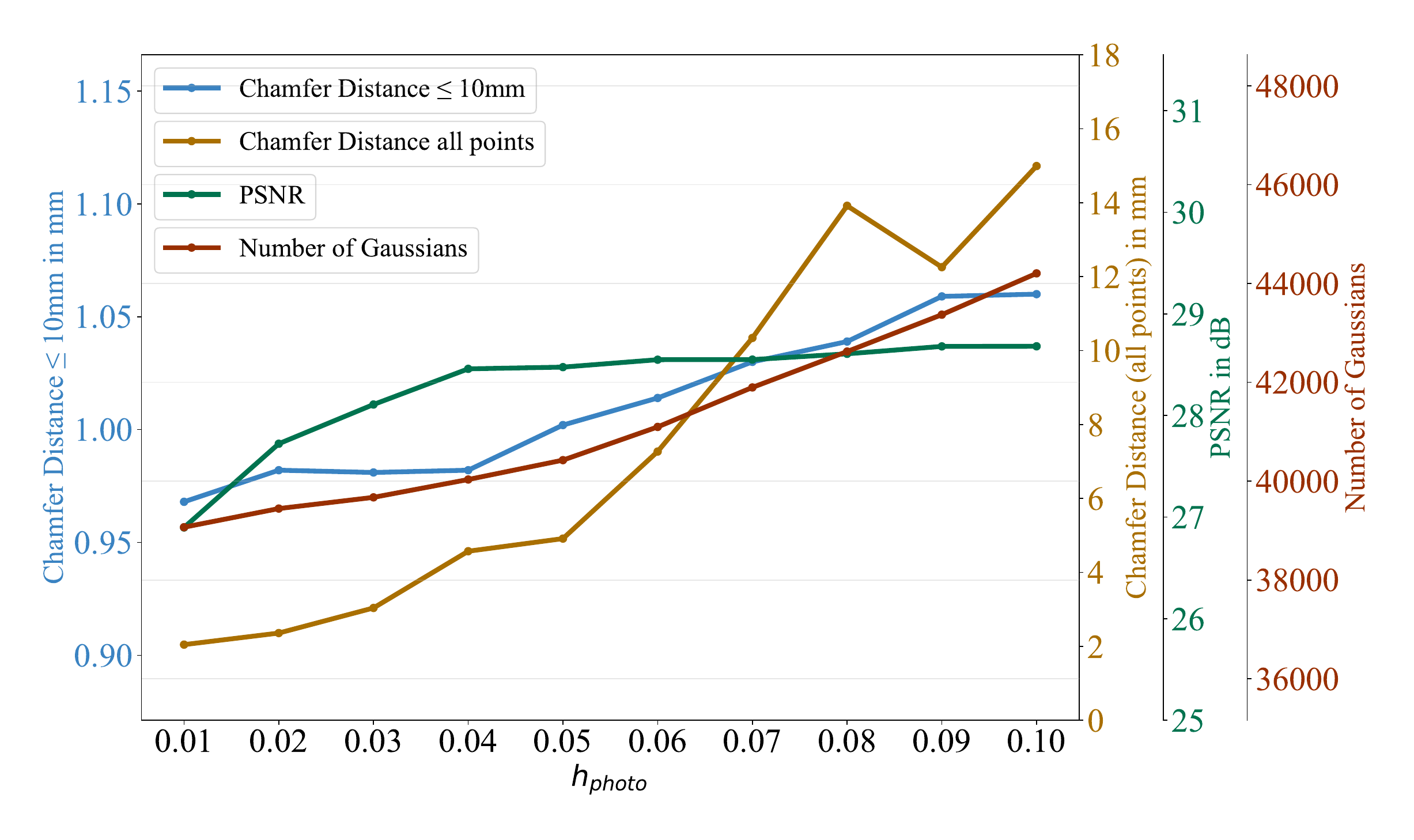}
   \vspace{-2mm}
   \caption{Chamfer cloud-to-cloud Distances, PSNR and Number of Gaussians at different weighting configurations of the photometric-geometric loss term with varying $h_{\text{photo}}$. The training incorporates 15\,000 iterations.}
   \label{fig:h_c2c_psnr}
\end{figure*}

The Chamfer cloud-to-cloud distance and PSNR for varying weights of the photometric loss \( h_{\text{photo}} \) is presented in Figure \ref{fig:h_c2c_psnr}. As \( h_{\text{photo}} \) increases, the Chamfer distance over all points increases (from 2.047\,mm at \( h_{\text{photo}} = 0.01 \) to 14.993\,mm at \( h_{\text{photo}} = 0.10 \)), indicating a decrease in geometric accuracy. The Chamfer distance for points within 10\,mm slightly increase from 0.968 to 1.060\,mm. The PSNR shows an improvement from 26.898 to 28.681\,dB. That suggest that lower values for \( h_{\text{photo}} \) improve geometric accuracy, while higher values enhance image quality at the cost of increased Chamfer distance and less accurate geometry. Therefore, the weight should be optimized according to the specific application of FeatureGS. Alternatively, achieving a higher PSNR with high geometric accuracy may require more training iterations.

\subsubsection{Fixed Training Time}\label{sec:quantitative_iterations}
The following quantitative results for the fixed number of training iterations of 15\,000 provide the geometric accuracy by Chamfer cloud-to-cloud distance, the number of resulting Gaussians, and the rendering quality reported by the PSNR.

\paragraph{Geometric Accuracy}

For the geometric accuracy of the surface points (Table \ref{tab:iterations_c2c_surface}), which are located at a distance of 10\,mm from the reference point cloud
$L_\text{Planarity, Gaussian}$ and $L_\text{Eigenentropy, kNN}$ yield often the best and second best highest geometric accuracies. 
$L_\text{Planarity, kNN}$ and $L_\text{Omnivariance, kNN}$ achieve a mixed result, but show good performance in some scenes such as scan24, scan37.
Nevertheless, the differences between all geometric-radiometric FeatureGS configurations are mostly marginal and stable across all scans (see e.g. scan55 with Chamfer distances from 0.967 to 0.971 mm). This is also reflected in the mean geometric accuracies.

Floater artifacts due to presumably incorrectly reconstructed Gaussians external to the actual object, where smaller values mean less disturbing artifacts, are illustrated by Table \ref{tab:iterations_c2c_floater}.
Regarding the reduction of floater artifacts, $L_\text{Planarity, Gaussian}$ and $L_\text{Eigenentropy, kNN}$ prove to be particularly effective. $L_\text{Planarity, Gaussian}$ often achieves the best results and shows a strong ability to minimize floater artifacts, especially for scans such as scan40 (4.816\,mm) and scan55 (4.782\,mm). On average, $L_\text{Planarity, Gaussian}$ performs best with 10.593\,mm, followed by $L_\text{Planarity, kNN}$ with 10.793 mm.
Overall, there is a significant improvement in all FeatureGS configurations compared to 3DGS, both in terms of surface accuracy and floater reduction.
FeatureGS reduces the mean Chamfer distance for surface accuracy by around 0.3 mm (approx. 20\% improvement). In particular, FeatureGS achieves a massive reduction of floater artifacts by approx. 90\%. 

\definecolor{yellow}{RGB}{255, 255, 204}
\definecolor{orange}{RGB}{255, 204, 153}
\definecolor{red}{RGB}{255, 153, 153}
\begin{table*}[h!]
\centering
\begin{tabular}{|l|c|c|c|c|c|}
\hline
\multirow{2}{*}{Methods} & \multirow{2}{*}{3DGS} & \multicolumn{4}{c|}{FeatureGS} \\ 
& & $L_\text{Planarity, Gaussian}$ & $L_\text{Planarity, kNN}$ & $L_\text{Omnivariance, kNN}$ & $L_\text{Eigenentropy, kNN}$ \\
\hline
scan24  & 1.702 & 1.421 & 1.438 & 1.434 & 1.432 \\ 
scan37  & 1.782 & 1.324 & 1.309 & 1.317 & 1.360 \\ 
scan40  & 1.625 & 1.002 & 1.002 & 0.989 & 1.001 \\ 
scan55  & 1.361 & 0.969 & 0.968 & 0.967 & 0.971 \\ 
scan63  & 2.061 & 1.483 & 1.449 & 1.462 & 1.481 \\ 
scan65  & 1.708 & 1.518 & 1.526 & 1.506 & 1.513 \\ 
scan69  & 1.671 & 1.299 & 1.316 & 1.312 & 1.314 \\ 
scan83  & 2.285 & 1.428 & 1.425 & 1.417 & 1.412 \\ 
scan97  & 1.855 & 1.689 & 1.684 & 1.695 & 1.689 \\ 
scan105 & 1.778 & 1.172 & 1.168 & 1.163 & 1.180 \\ 
scan106 & 1.514 & 0.936 & 0.939 & 0.948 & 0.950 \\ 
scan110 & 1.486 & 1.819 & 1.821 & 1.800 & 1.808 \\ 
scan114 & 1.549 & 0.966 & 0.952 & 0.960 & 0.945 \\ 
scan118 & 1.291 & 0.875 & 0.854 & 0.873 & 0.866 \\ 
scan122 & 1.289 & 0.992 & 1.000 & 0.991 & 0.990 \\ 
\hline
Mean    & 1.609 & \cellcolor{yellow}1.313 & \cellcolor{red}1.310 & \cellcolor{orange}1.311 & 1.315 \\  
\hline
\end{tabular}
\caption{Surface accuracy. \textbf{Geometric accuracy} comparison on the DTU dataset with Chamfer cloud-to-cloud distances $\downarrow$ in mm for points $\leq$10\,mm from the reference, according to the DTU evaluation script. Best results are highlighted as \textcolor{red}{1st}, \textcolor{orange}{2nd}, and \textcolor{yellow}{3rd}. Mean scores are listed at the bottom. The training incorporates 15\,000 iterations.}
\label{tab:iterations_c2c_surface}
\end{table*}

\definecolor{yellow}{RGB}{255, 255, 204}
\definecolor{orange}{RGB}{255, 204, 153}
\definecolor{red}{RGB}{255, 153, 153}
\begin{table*}[h!]
\centering
\begin{tabular}{|l|c|c|c|c|c|}
\hline
\multirow{2}{*}{Methods} & \multirow{2}{*}{3DGS} & \multicolumn{4}{c|}{FeatureGS} \\ 
& & $L_\text{Planarity, Gaussian}$ & $L_\text{Planarity, kNN}$ & $L_\text{Omnivariance, kNN}$ & $L_\text{Eigenentropy, kNN}$ \\
\hline
scan24  & 50.850 & 8.909 & 9.378 & 21.158 & 12.469 \\ 
scan37  & 53.919 & 9.575 & 8.312 & 9.435 & 9.045 \\ 
scan40  &43.597 & 4.915 & 5.174 & 4.816 & 5.267 \\
scan55  & 58.004 & 5.050 & 5.990 & 4.782 & 5.059 \\
scan63  & 279.172 & 19.130 & 24.350 & 20.405 & 22.034 \\
scan65  & 179.180 & 17.916 & 15.357 & 19.246 & 18.741 \\ 
scan69  & 121.251 & 10.110 & 9.653 & 9.708 & 9.770 \\ 
scan83  & 178.645 & 24.628 & 21.874 & 22.426 &  21.545\\ 
scan97  & 111.836 & 13.099 & 12.033 & 11.333 & 9.755 \\ 
scan105 & 132.986 & 8.221 & 8.159 & 8.260 & 8.480 \\ 
scan106 & 88.501 & 3.272 & 3.459 & 3.058 & 3.211 \\ 
scan110 & 164.030 & 17.16 & 14.134 & 17.584 & 18.517 \\ 
scan114 & 173.681 & 5.850 & 6.773 & 6.138 &6.002 \\ 
scan118 & 83.070& 6.977 & 6.374 & 7.005 & 7.087 \\ 
scan122 & 124.686 & 9.332 & 9.755 & 9.265 & 9.369 \\ 
\hline
Mean    & 116.587 & \cellcolor{red}10.593 & \cellcolor{orange}10.793 & 12.212 & \cellcolor{yellow}11.721 \\  
\hline
\end{tabular}
\caption{Floater Artifacts. \textbf{Geometric accuracy} comparison on the DTU dataset with Chamfer cloud-to-cloud distances$\downarrow$ in mm are reported for all points to focus on floaters external to the point cloud. Best results are highlighted as \textcolor{red}{1st}, \textcolor{orange}{2nd}, and \textcolor{yellow}{3rd}. Mean scores are listed at the bottom. The training incorporates 15\,000 iterations.}
\label{tab:iterations_c2c_floater}
\end{table*}

\paragraph{Number of Gaussians}
Table \ref{tab:iterations_gaussians} shows the number of Gaussians generated by 3DGS and the different loss configurations of FeatureGS. 
The mean values indicate that all the FeatureGS configurations reduce the number of Gaussians by around 440\,000 Gaussians on average, which corresponds to a reduction of around 95\%.
FeatureGS increases the number of initial points by only around 7\%. The relative differences between the FeatureGS configurations are only minor. All FeatureGS configurations deliver a consistently clear reduction compared to 3DGS.

\definecolor{yellow}{RGB}{255, 255, 204}
\definecolor{orange}{RGB}{255, 204, 153}
\definecolor{red}{RGB}{255, 153, 153}
\begin{table*}[h!]
\centering
\begin{tabular}{|l|c|c|c|c|c|c|c|c|}
\hline
\multirow{2}{*}{Methods} & \multirow{2}{*}{3DGS} & \multicolumn{4}{c|}{FeatureGS} & \multirow{2}{*}{Initial SfM Points} \\ 
& &$L_\text{Planarity, Gaussian}$&  $L_\text{Planarity, kNN}$ & $L_\text{Omnivariance, kNN}$ & $L_\text{Eigenentropy, kNN}$ & \\
\hline
scan24  & 673\,276 & 20\,105 & 20\,423 & 20\,485 & 20\,440 &    15\,479 \\ 
scan37  & 766\,722 & 29\,431 & 29\,111 & 29\,247 & 29\,291 &    24\,857 \\ 
scan40  & 831\,896 & 40\,425 & 40\,404 & 40\,445 & 40\,429 & 39\,158 \\ 
scan55  & 739\,171 & 34\,760 & 34\,774 & 34\,738 & 34\,780 &    33\,506 \\ 
scan63  & 249\,496 & 13\,343 & 13\,461 & 13\,610 & 13\,509 &    10\,869 \\ 
scan65  & 347\,906& 14\,231 & 14\,154 & 14\,213 & 14\,216 &    13\,203 \\ 
scan69  & 304\,854 & 15\,931 & 15\,911 & 15\,906 & 15\,911 &    15\,264 \\ 
scan83  & 216\,765 & 11\,982 & 11\,921& 12\,054 & 11\,913 & 10\,652\\ 
scan97  & 595\,899& 22\,699 & 22\,717 & 22\,579 & 22\,436 &    20\,467\\ 
scan105 & 250\,257 & 26\,102 & 26\,154 & 26\,111 & 26\,210 &    25\,291 \\ 
scan106 & 269\,773 & 33\,701 & 33\,707 & 33\,696 & 33\,705 &    33\,523 \\ 
scan110 & 227\,484 & 11\,822 & 11\,768 & 11\,835 & 11\,838 & 11\,382 \\ 
scan114 & 361\,373 & 26\,208 & 26\,248 & 26\,226 & 26\,199 &       25\,761 \\ 
scan118 & 357\,583 & 27\,964 & 27\,948 & 27\,973 & 27\,967 &    27\,650 \\ 
scan122 & 318\,226 & 21\,427 & 21\,423 & 21\,417 & 21\,417& 20\,975\\ 
\hline
Mean    & 462\,699 & \cellcolor{red}24\,275 & \cellcolor{orange}24\,277 & 24\,302 & \cellcolor{yellow}24\,291 & 22\,771 \\ 
\hline
\end{tabular}
\caption{\textbf{Number of Gaussians} on the DTU dataset. We report the Total Number of Gaussians $\downarrow$ compared with baselines. Mean scores are listed at the bottom. Best results (lowest total number) concerning memory are highlighted as \textcolor{red}{1st}, \textcolor{orange}{2nd}, and \textcolor{yellow}{3rd}. The training incorporates 15\,000 iterations.}
\label{tab:iterations_gaussians}
\end{table*}

\paragraph{Rendering Quality}
While FeatureGS is significantly more memory efficient (fewer Gaussians, less storage required), has less floater artifacts and delivers geometrically more accurate results, there are drawbacks in rendering quality (Table \ref{tab:iterations_psnr}). On average, the mean PSNR values appear lower with a decrease in rendering quality of approximately 3.3\,dB. The differences between the different FeatureGS loss formulations are minimal (less than 0.1\,dB).

\definecolor{yellow}{RGB}{255, 255, 204}
\definecolor{orange}{RGB}{255, 204, 153}
\definecolor{red}{RGB}{255, 153, 153}
\begin{table*}[h!]
\centering
\begin{tabular}{|l|c|c|c|c|c|c|}
\hline
\multirow{2}{*}{Methods} & \multirow{2}{*}{3DGS} & \multicolumn{4}{c|}{FeatureGS} \\ 
& & $L_\text{Planarity, Gaussian}$ &  $L_\text{Planarity, kNN}$ & $L_\text{Omnivariance, kNN}$& $L_\text{Eigenentropy, kNN}$ \\
\hline
scan24 & 35.16 & 29.86 & 29.90 & 29.98& 29.93 \\ 
scan37 & 29.98 & 26.32 & 26.35 & 26.39 & 26.36 \\ 
scan40  & 34.59 & 28.48 & 28.52 & 28.56 & 28.45\\
scan55  & 34.08 & 29.48 & 29.55 & 29.50 & 29.56\\
scan63  & 37.35 & 32.65 & 32.66 & 32.74 & 32.81\\
scan65  & 35.19 & 30.35 & 30.35 & 30.36 & 30.35 \\
scan69  & 33.50 & 28.33 & 28.57 & 28.54 & 28.53\\
scan83  & 34.08 & 32.69 & 32.82 & 32.84 & 31.85\\ 
scan97  & 32.57 & 30.03 & 30.06 & 30.12 & 30.01  \\ 
scan105 & 36.70 & 34.68 & 34.64 & 34.51 & 34.59 \\ 
scan106 & 37.48 & 36.01 & 36.09 & 36.06 & 36.03  \\ 
scan110 & 31.81 & 29.94 & 29.96 & 29.92 & 29.94  \\ 
scan114 & 34.78 & 32.73 & 32.55 & 32.70 & 32.64  \\ 
scan118 & 36.71 & 34.81 & 34.85 & 34.82 & 34.83 \\ 
scan122 & 36.06 & 34.15 & 34.17 & 34.15 & 34.17 \\ 
\hline
Mean &  \cellcolor{red}34.67 &  31.37 & \cellcolor{yellow}31.40 & \cellcolor{orange}31.41 &  31.34\\ 
\hline
\end{tabular}
\caption{\textbf{Rendering quality} comparison on the DTU dataset. We report the PSNR $\uparrow$ in dB. Mean scores are listed at the bottom. Best results are highlighted as \textcolor{red}{1st}, \textcolor{orange}{2nd}, and \textcolor{yellow}{3rd}. The training incorporates 15\,000 iterations.}
\label{tab:iterations_psnr}
\end{table*}

\subsubsection{Fixed Rendering Quality}\label{sec:quantitative_rendering}
The quantitative results for the fixed PSNR using early stopping demonstrate the geometric accuracy due to the Chamfer cloud-to-cloud distance and the number of Gaussians required for this. The comparison of 3DGS and the FeatureGS configurations with identical PSNR serves to evaluate different aspects of the methods under comparable rendering qualities. This ensures that differences in other metrics such as geometric accuracy, number of Gaussians or floater artifacts are not influenced by a varying of the rendering quality.

\paragraph{Geometric Accuracy}
FeatureGS consistently outperforms 3DGS in geometric accuracy of surface points (Table \ref{tab:earlystop_c2c_surface}) for the same rendering quality, with an average improvement of about 30\%, with a mean geometric accuracy of 1.826\,mm for 3DGS to 1.300 to 1.310\,mm for FeatureGS. The different loss formulations of FeatureGS show only minimally different results with differences of less than 1 percent.

\definecolor{yellow}{RGB}{255, 255, 204}
\definecolor{orange}{RGB}{255, 204, 153}
\definecolor{red}{RGB}{255, 153, 153}
\begin{table*}[h!]
\centering
\begin{tabular}{|l|c|c|c|c|c|}
\hline
\multirow{2}{*}{Methods} & \multirow{2}{*}{3DGS} & \multicolumn{4}{c|}{FeatureGS} \\ 
& & $L_\text{Planarity, Gaussian}$ & $L_\text{Planarity, kNN}$ & $L_\text{Omnivariance, kNN}$ & $L_\text{Eigenentropy, kNN}$ \\
\hline
scan24  & 2.026 & 1.424 & 1.463 & 1.475 & 1.446 \\      
scan37  & 1.847 & 1.297 & 1.313 &  1.278 & 1.280 \\                         
scan40  & 1.758 & 0.954 & 0.948 & 0.952 & 0.951 \\ 
scan55  & 1.672 & 0.935 & 0.918 & 0.944 & 0.914 \\ 
scan63  & 2.155 & 1.530 & 1.534 & 1.504 & 1.500\\                         
scan65  & 2.095 & 1.589 & 1.576 & 1.581 & 1.582 \\ 
scan69  & 1.916 & 1.288 & 1.271 & 1.290 & 1.271 \\                
scan83  & 2.211 & 1.438 & 1.489 & 1.507 & 1.509 \\                        
scan97  & 1.912 & 1.680 & 1.699 & 1.704 & 1.704 \\ 
scan105 & 1.769 & 1.264 & 1.332 & 1.293 & 1.280\\ 
scan106 & 1.574 & 1.095 & 1.104 & 1.104 & 1.100 \\ 
scan110 & 1.902 & 1.866 & 1.854 & 1.831 & 1.853 \\ 
scan114 & 1.453 & 1.010 & 1.015 & 1.008 & 1.022  \\ 
scan118 & 1.503 & 1.064 & 1.063  & 1.053 & 1.063 \\ 
scan122 & 1.604 & 1.060 & 1.070 & 1.060 & 1.051 \\ 
\hline
Mean    & 1.826 & \cellcolor{red}1.300 & 1.310 & \cellcolor{yellow}1.306 & \cellcolor{orange}1.302 \\ 
\hline
\end{tabular}
\caption{Surface accuracy. \textbf{Geometric accuracy} comparison on the DTU dataset with Chamfer cloud-to-cloud distances $\downarrow$ in mm for surface points $\leq$10\,mm from the reference, according to the DTU evaluation script. Best results are highlighted as \textcolor{red}{1st}, \textcolor{orange}{2nd}, and \textcolor{yellow}{3rd}. Mean scores are listed at the bottom. The training incorporates iterations until early-stopping at \textbf{same PSNR}.}
\label{tab:earlystop_c2c_surface}
\end{table*}

FeatureGS heavily reduces floater artifacts (Table \ref{tab:earlystop_c2c_floater}) at the same rendering quality by an average of 90\% compared to 3DGS. This is shown in the mean Chamfer Distance, which is reduced from 93.690\,mm for 3DGS to 10.620 - 10.771\,mm for FeatureGS. Overall, the photometric-geometric loss formulations with $L_\text{Planarity, kNN}$ and $L_\text{Omnivariance, kNN}$ result in the lowest amount of floater artifacts.

\definecolor{yellow}{RGB}{255, 255, 204}
\definecolor{orange}{RGB}{255, 204, 153}
\definecolor{red}{RGB}{255, 153, 153}
\begin{table*}[h!]
\centering
\begin{tabular}{|l|c|c|c|c|c|}
\hline
\multirow{2}{*}{Methods} & \multirow{2}{*}{3DGS} & \multicolumn{4}{c|}{FeatureGS} \\ 
& & $L_\text{Planarity, Gaussian}$ & $L_\text{Planarity, kNN}$ & $L_\text{Omnivariance, kNN}$ & $L_\text{Eigenentropy, kNN}$ \\
\hline
scan24  & 32.241 & 11.835 & 8.441 & 14.137 & 9.151 \\ 
scan37  & 72.622 & 9.451 & 11.153 & 8.361 & 8.852 \\ 
scan40  & 19.356 & 4.796 & 5.475 & 5.191 & 5.751 \\ 
scan55  & 36.010 & 5.233 & 4.872 & 4.727 & 5.199 \\ 
scan63  & 200.478 & 20.862 & 19.600 & 20.744 & 22.942 \\ 
scan65  & 163.601 & 17.775 & 18.103 & 15.096 & 17.191 \\ 
scan69  & 61.014 & 9.524 & 9.5613 & 9.736 & 9.561 \\ 
scan83  & 139.395 & 22.819 & 22.671 & 21.552 & 23.454 \\ 
scan97  & 70.390 & 11.543 & 12.408 & 12.085 & 11.960 \\ 
scan105 & 80.220 & 8.102 &  8.037 & 8.067 & 8.172 \\ 
scan106 & 32.873 & 3.228 & 3.021 & 3.031 & 3.206 \\ 
scan110 & 111.052 & 16.205 & 15.742 & 16.463 & 16.771 \\  
scan114 & 57.211 & 6.006 & 5.732 & 6.078 & 5.409 \\ 
scan118 & 58.760 & 5.199 & 5.172 &  4.870 & 4.814 \\ 
scan122 & 270.132 & 8.983 & 9.504 & 9.155 & 8.688 \\ 
\hline
Mean    &  93.690 & 10.771 & \cellcolor{orange}10.633 & \cellcolor{red}10.620 &  \cellcolor{yellow}10.741 \\ 
\hline
\end{tabular}
\caption{Floater Artifacts. \textbf{Geometric accuracy} comparison on the DTU dataset with Chamfer cloud-to-cloud distances $\downarrow$  in mm are reported for all points to focus on floaters external to the point cloud. Best results are highlighted as \textcolor{red}{1st}, \textcolor{orange}{2nd}, and \textcolor{yellow}{3rd}. Mean scores are listed at the bottom. The training incorporates iterations until early-stopping at \textbf{same PSNR}.}
\label{tab:earlystop_c2c_floater}
\end{table*}

\paragraph{Number of Gaussians}

In addition, the number of Gaussians (Table \ref{tab:earlystop_gaussians}) is reduced by FeatureGS while maintaining the same rendering quality compared to 3DGS. FeatureGS drastically reduces the number of Gaussians by around 90\% from an average of 249\,986 Gaussians to between 26\,380 and 26\,389 Gaussians. This leads to a lower memory requirement. Within FeatureGS, the variants show equivalent compression of the number of Gaussians.

\definecolor{yellow}{RGB}{255, 255, 204}
\definecolor{orange}{RGB}{255, 204, 153}
\definecolor{red}{RGB}{255, 153, 153}
\begin{table*}[h!]
\centering
\begin{tabular}{|l|c|c|c|c|c|c|c|c|}
\hline
\multirow{2}{*}{Methods} & \multirow{2}{*}{3DGS} & \multicolumn{4}{c|}{FeatureGS} & \multirow{2}{*}{Initial SfM Points} \\ 
& &$L_\text{Planarity, Gaussian}$&  $L_\text{Planarity, kNN}$ & $L_\text{Omnivariance, kNN}$ & $L_\text{Eigenentropy, kNN}$ & \\
\hline
scan24  & 333\,870& 20\,491 & 20\,100 & 20\,553 & 20\,155 &  15\,479 \\ 
scan37  & 527\,713& 29\,199 & 29\,596& 29\,371 & 28\,972 &  24\,857 \\   
scan40  & 537\,082& 40\,353 & 40\,406 & 40\,446 & 40\,466 &  39\,158 \\ 
scan55  & 470\,449 & 34\,788 & 34\,744 & 34\,742 & 34\,769 & 33\,506 \\ 
scan63  & 113\,493& 13\,346 & 13\,323& 13\,183 & 13\,155 & 10\,869 \\ 
scan65  & 151\,776 & 14\,213 & 14\,176 & 14\,187 & 14\,179 & 13\,203 \\ 
scan69  & 147\,690 & 15\,893 & 15\,908 & 15\,905 & 15\,908 & 15\,264 \\ 
scan83  & 132\,898 & 11\,926 & 12\,044& 11\,892 & 12\,046 & 10\,652\\  
scan97  & 303\,676  & 22\,454 & 22\,824 & 22\,732 & 22\,712 & 20\,467\\   
scan105 &  184\,679 & 26\,166 & 26\,158 & 26\,149 & 26\,178 & 25\,291 \\  
scan106 &  125\,419  & 33\,730 & 33\,719 & 33\,724 & 33\,731 &  33\,523 \\  
scan110 & 140\,378  & 11\,819& 11\,802& 11\,811 & 11\,824 &  11\,382 \\   
scan114 & 190\,674 & 26\,247 & 26\,232 & 26\,255 & 26\,228 & 25\,761 \\  
scan118 & 165\,017 & 27\,927 & 27\,926 & 27\,910 & 27\,902 &  27\,650 \\  
scan122 & 172\,589 & 21\,427 & 21\,432 & 21\,436 & 21\,407 & 20\,975\\     
\hline
Mean    & 249\,986 &  26\,389 & \cellcolor{orange}26\,385 & \cellcolor{red}26\,380 & \cellcolor{yellow}26\,387 &  22\,771\\ 
\hline
\end{tabular}
\caption{\textbf{Number of Gaussians} on the DTU dataset. We report the Total Number of Gaussians $\downarrow$. Mean scores are listed at the bottom. Best results (lowest total number) concerning memory are highlighted as \textcolor{red}{1st}, \textcolor{orange}{2nd}, and \textcolor{yellow}{3rd}. The training incorporates iterations until early-stopping at \textbf{same PSNR}.}
\label{tab:earlystop_gaussians}
\end{table*}

\subsection{Qualitative Results}\label{sec:qualitative}
Similar to the quantitative results, FeatureGS yields promising results in terms of geometric accuracy of the 3D point clouds as well as rendering quality by removing floater artifacts. Through all geometric loss terms of FeatureGS, consistently accurate and floater artifact-reduced results are generated across all 15 scenes, compared to 3DGS. All results are shown for the exact same PSNR values by early-stopping, thus the same rendering quality.

\paragraph{Geometric Accuracy}

\begin{figure*}[htbp]
    \centering
    \begin{tabular}{c c c c c}
        \textbf{} & \textbf{scan24} & \textbf{scan37} & \textbf{scan40} & \textbf{scan55} \\[1ex]
        
        \rotatebox{90}{\textbf{3DGS}} &
         \vspace{-4mm}
        \includegraphics[width=0.225\linewidth]{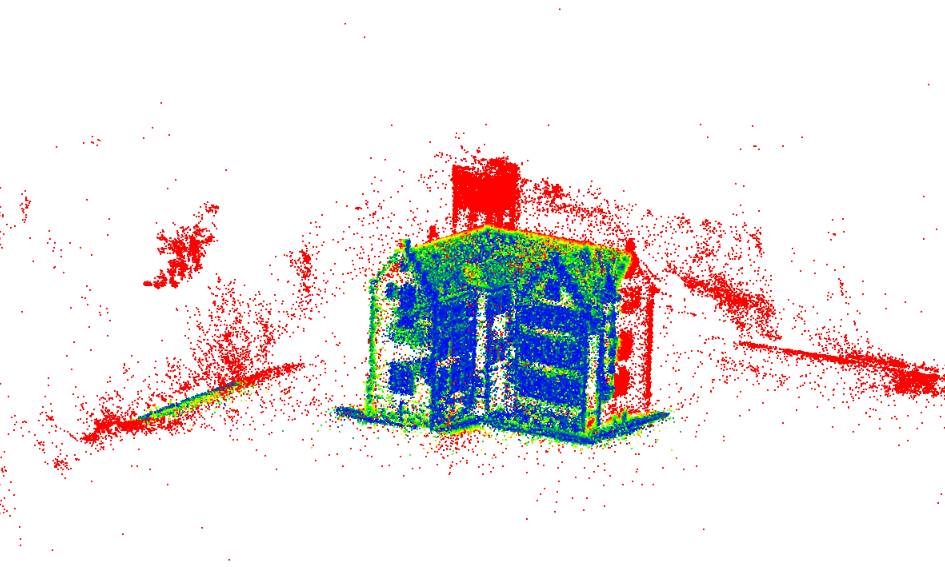} &
      \hspace{-4mm}
        \includegraphics[width=0.225\linewidth]{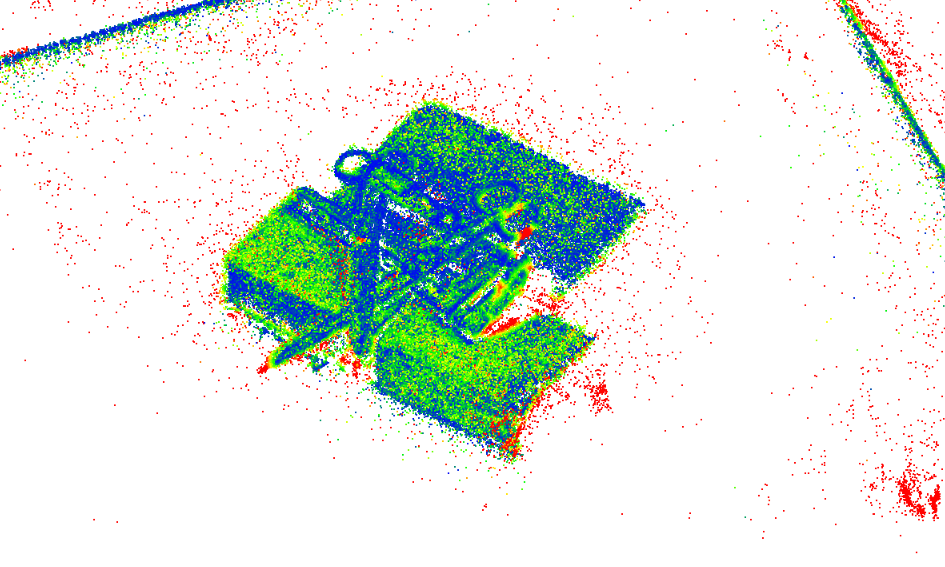} &
         \hspace{-4mm}
        \includegraphics[width=0.225\linewidth]{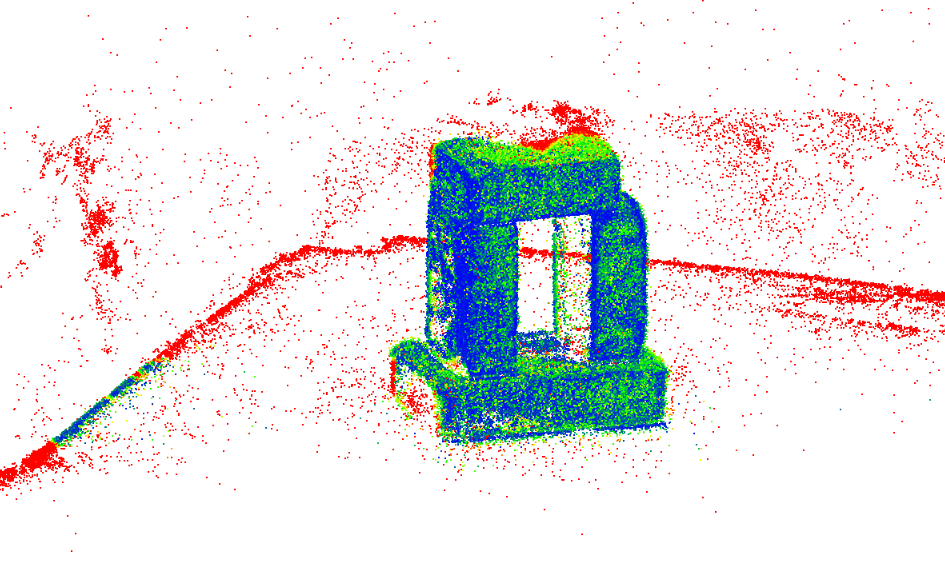} &
        \hspace{-4mm}
        \includegraphics[width=0.225\linewidth]{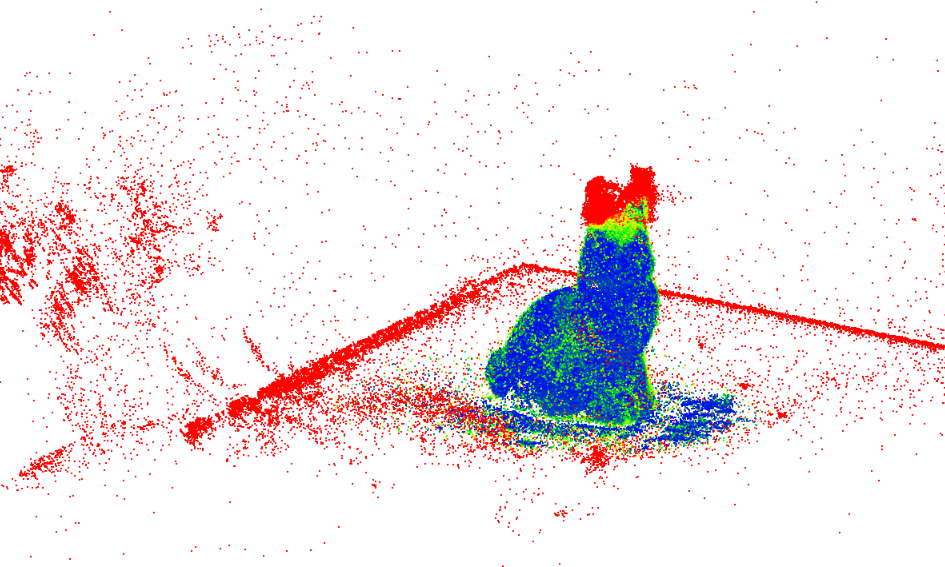} \vspace{-2mm}\\[2ex]
       
        \rotatebox{90}{\textbf{FeatureGS}} &
         \vspace{-4mm}
        \includegraphics[width=0.225\linewidth]{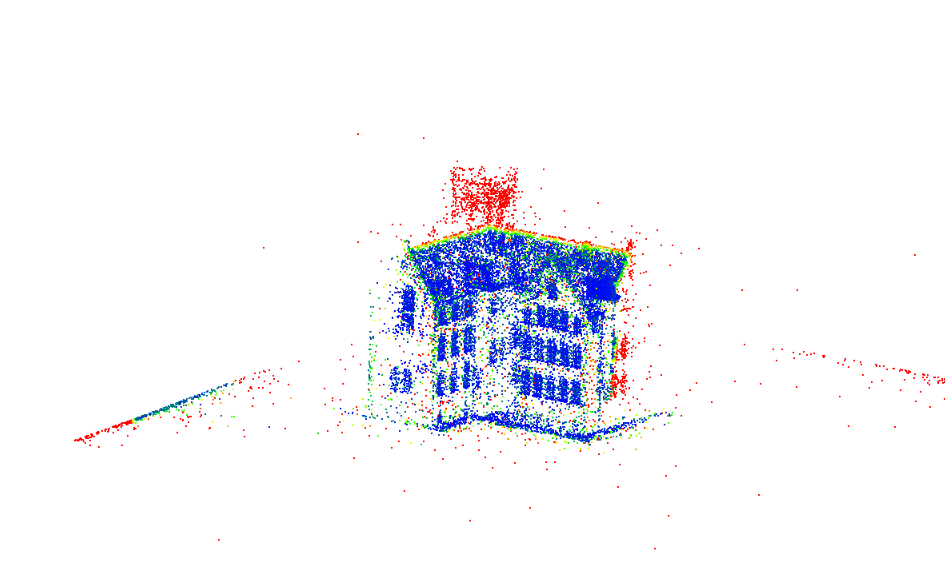} &
      \hspace{-4mm}
        \includegraphics[width=0.225\linewidth]{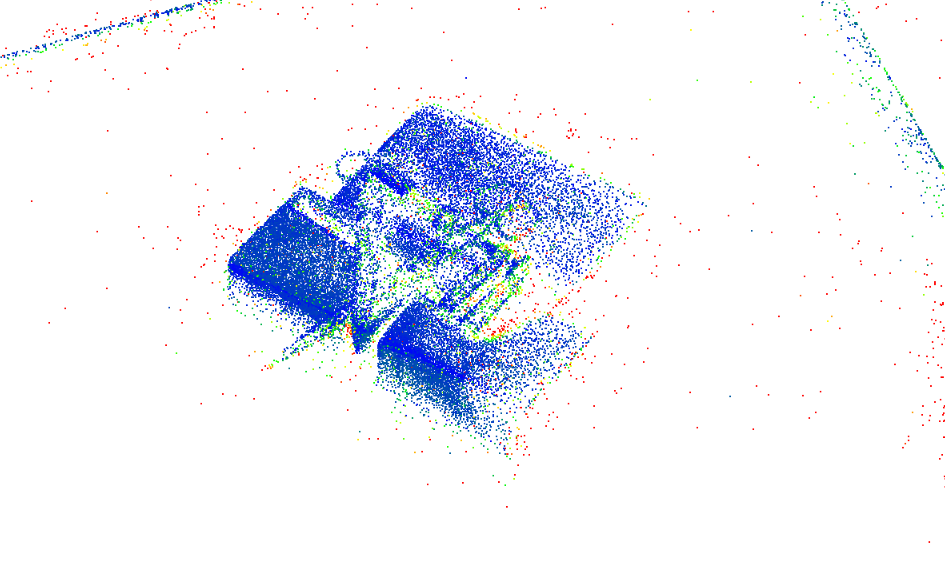} & 
         \hspace{-4mm}
        \includegraphics[width=0.225\linewidth]{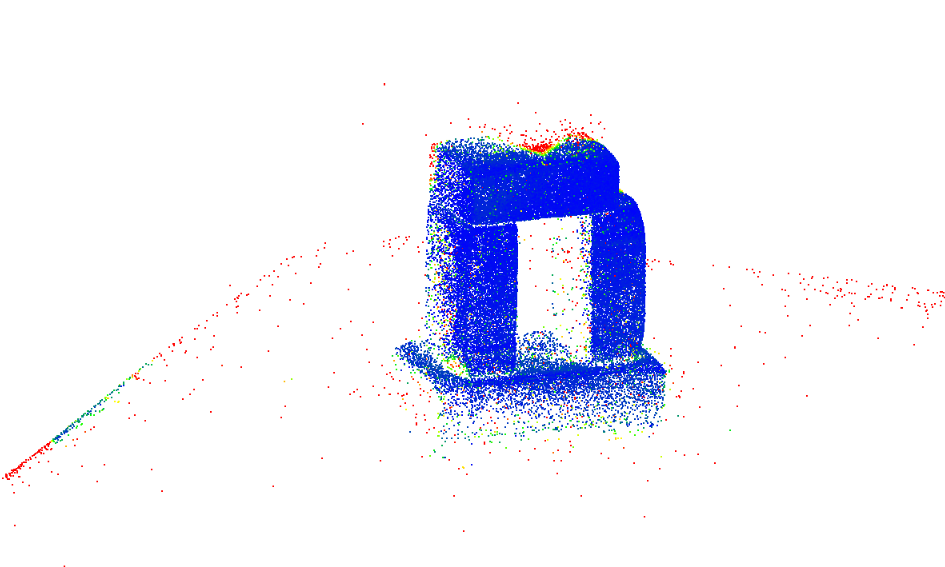} & 
      \hspace{-4mm}
        \includegraphics[width=0.225\linewidth]{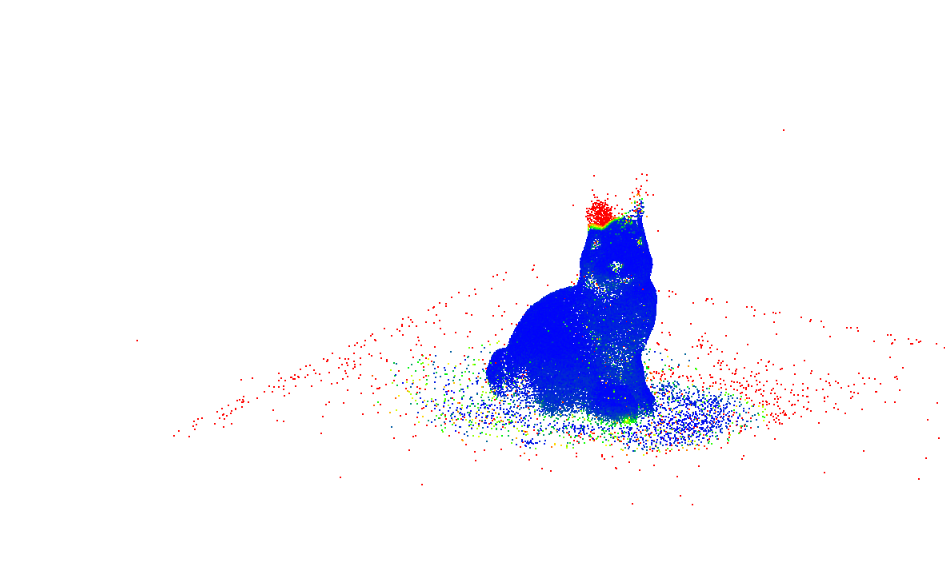} \\[2ex]

        \textbf{} & \textbf{scan63} & \textbf{scan65} & \textbf{scan69} & \textbf{scan83}\\[1ex]
        
        \rotatebox{90}{\textbf{3DGS}} &
         \vspace{-4mm}
        \includegraphics[width=0.225\linewidth]{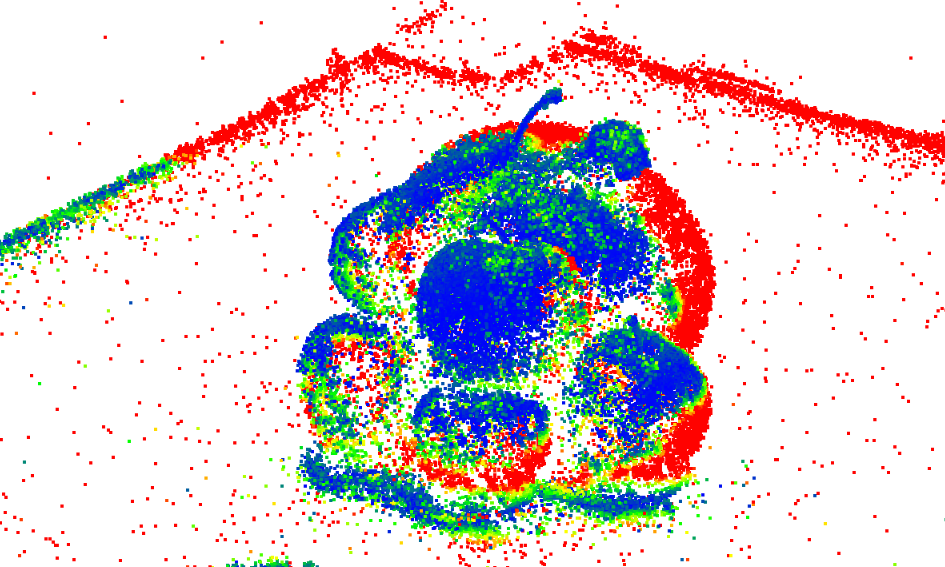} &
        \hspace{-4mm}
        \includegraphics[width=0.225\linewidth]{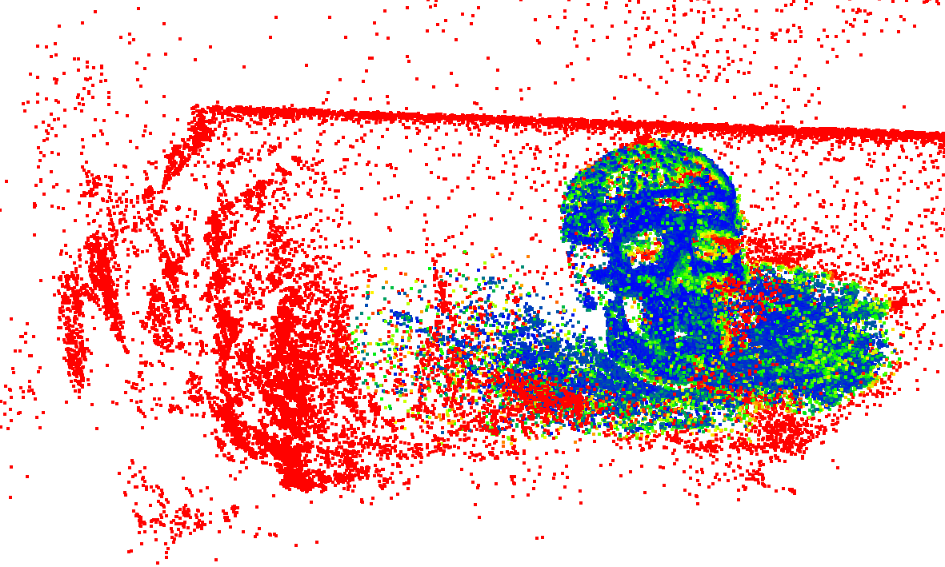} &
       \hspace{-4mm}
        \includegraphics[width=0.225\linewidth]{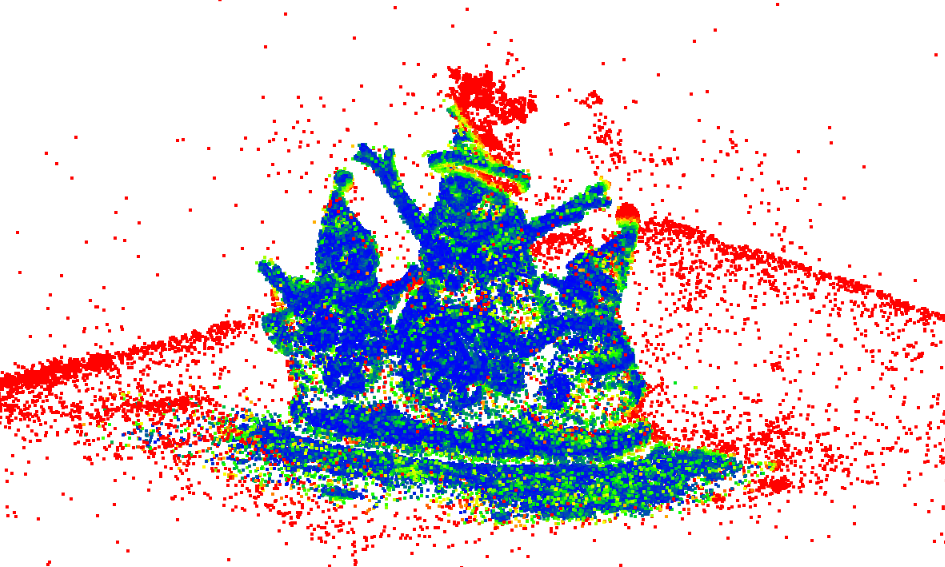} &
       \hspace{-4mm}
        \includegraphics[width=0.225\linewidth]{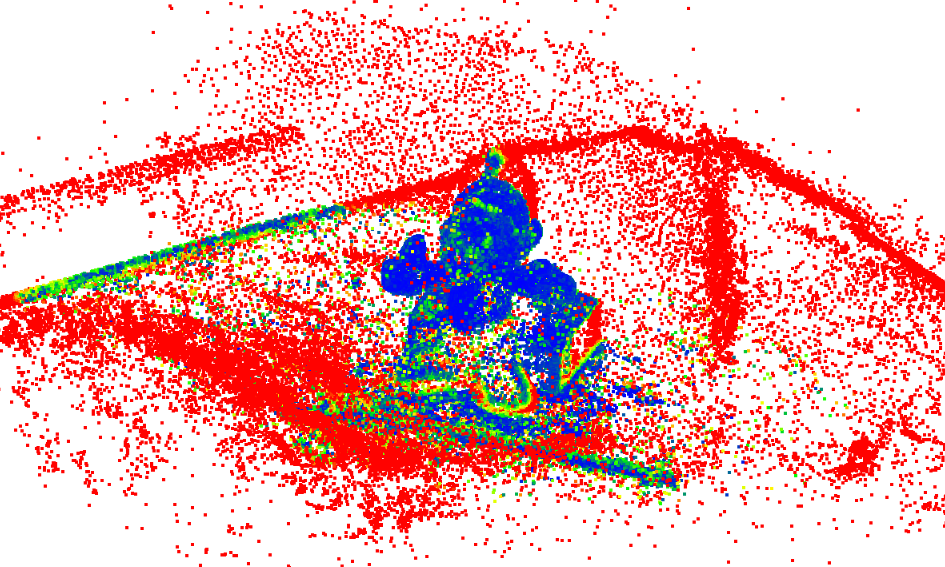}  \vspace{-2mm}\\[2ex]
    
        \rotatebox{90}{\textbf{FeatureGS}} &
         \vspace{-4mm}
        \includegraphics[width=0.225\linewidth]{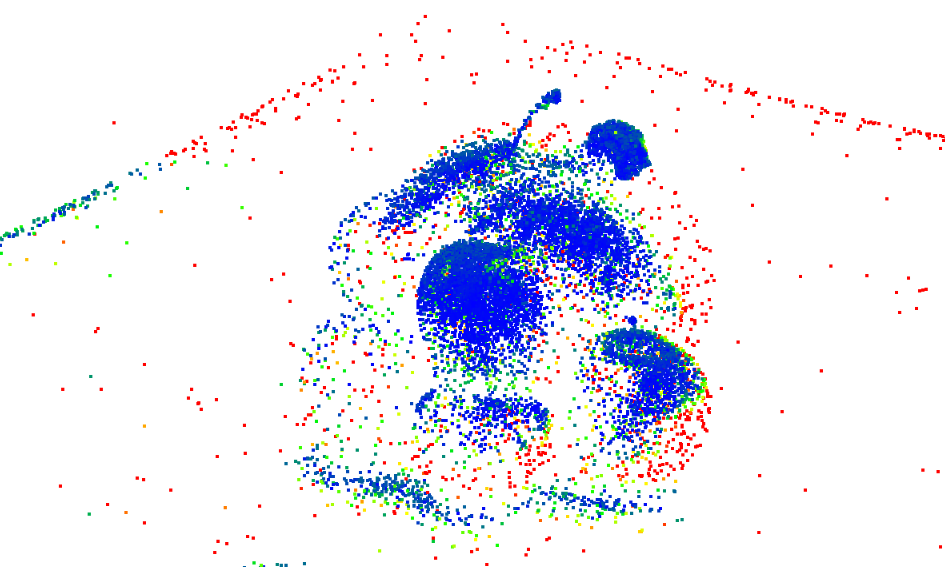} &
        \hspace{-4mm}
        \includegraphics[width=0.225\linewidth]{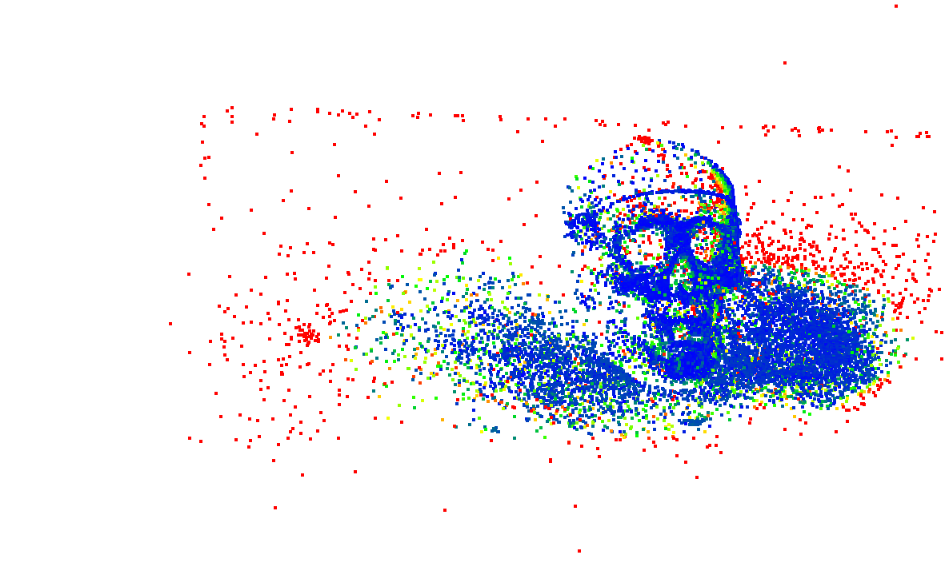} &
        \hspace{-4mm}
        \includegraphics[width=0.225\linewidth]{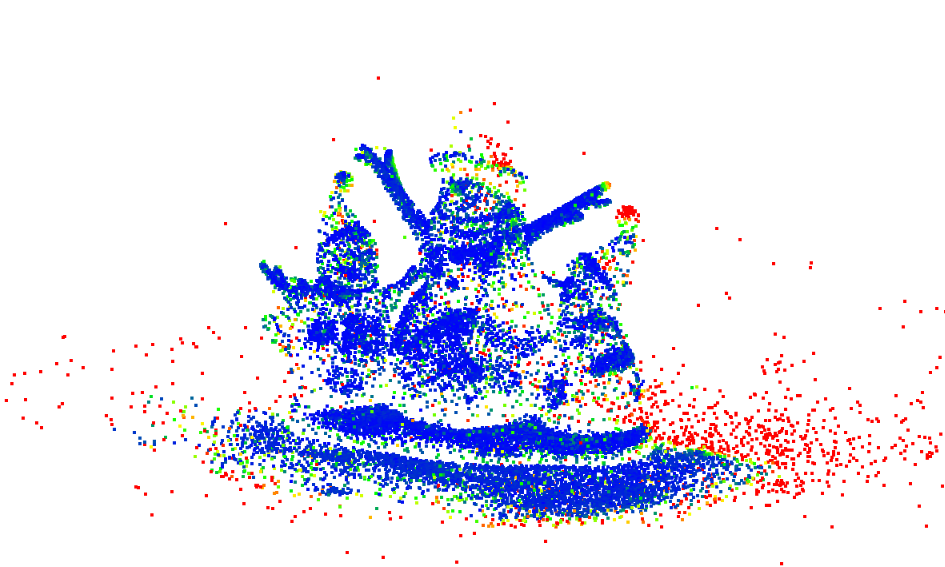} &
        \hspace{-4mm}
        \includegraphics[width=0.225\linewidth]{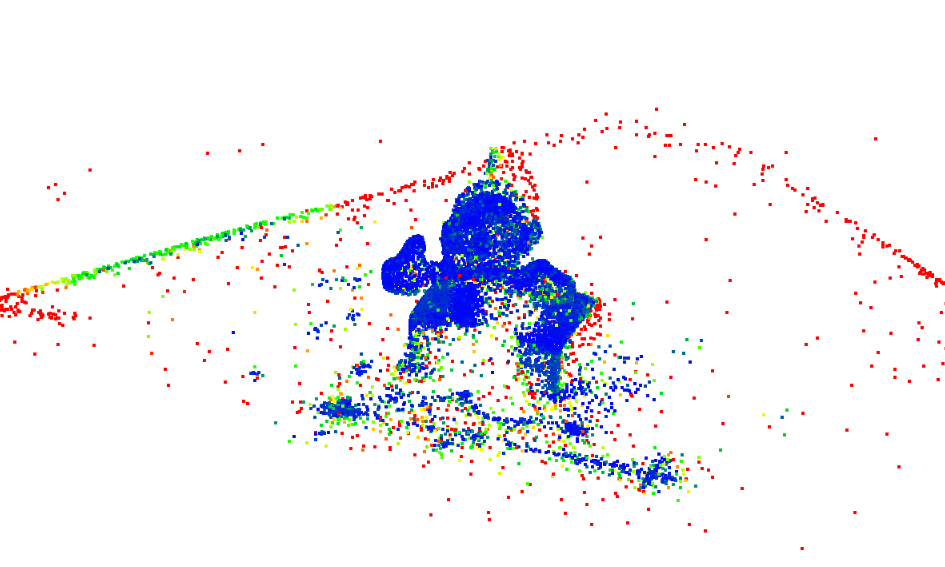} \\[2ex]

                \textbf{} & \textbf{scan97} & \textbf{scan105} & \textbf{scan106} & \textbf{scan110}\\[1ex]
        
        \rotatebox{90}{\textbf{3DGS}} &
         \vspace{-4mm}
        \includegraphics[width=0.225\linewidth]{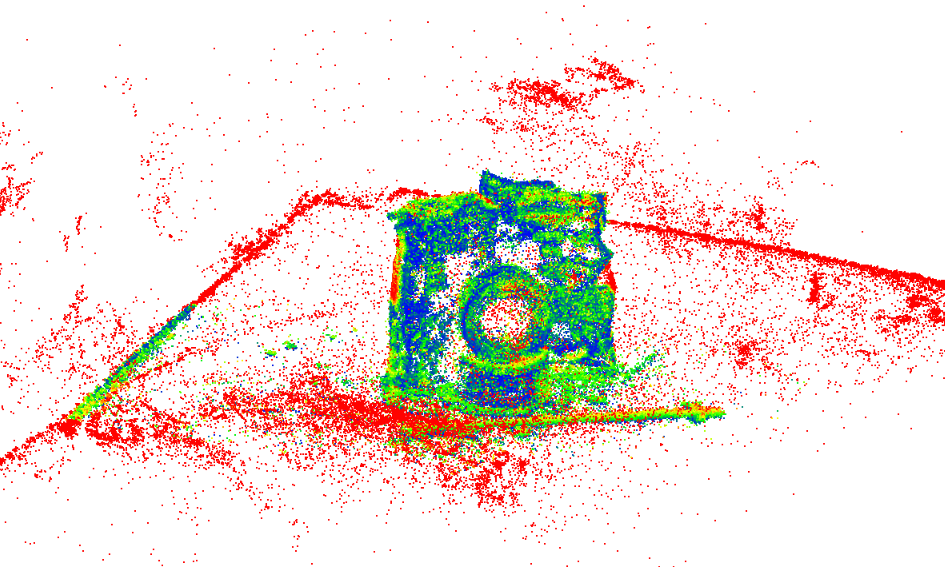} &
        \hspace{-4mm}
        \includegraphics[width=0.225\linewidth]{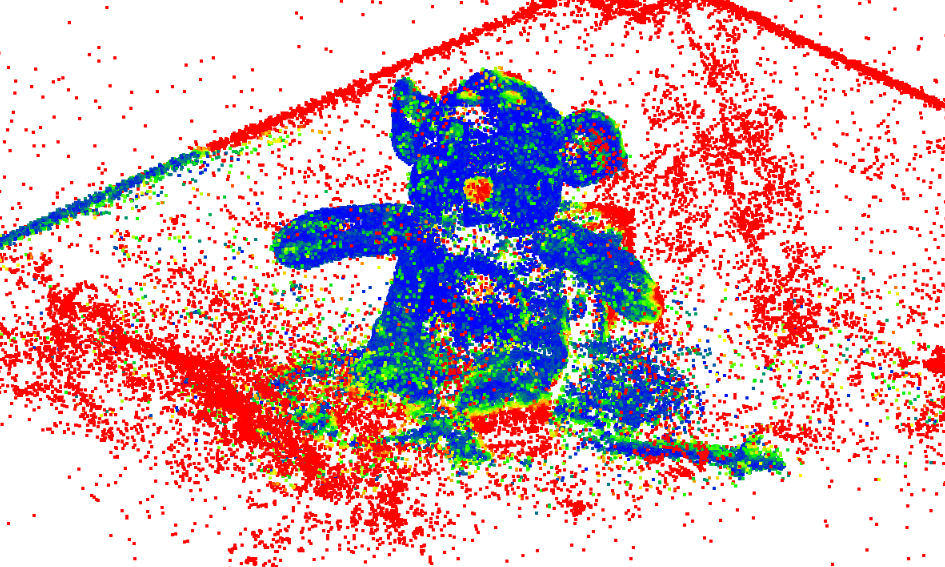} &
       \hspace{-4mm}
        \includegraphics[width=0.225\linewidth]{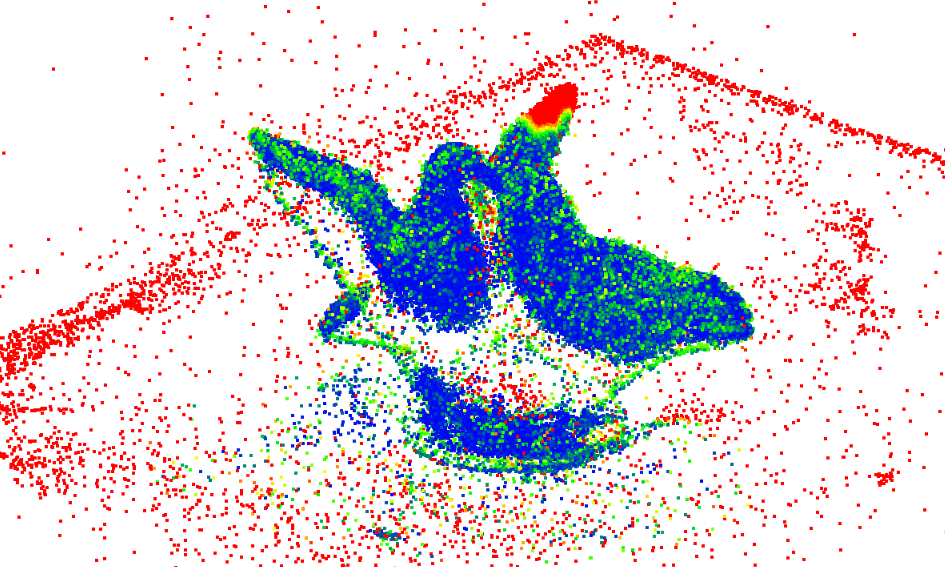} &
       \hspace{-4mm}
        \includegraphics[width=0.225\linewidth]{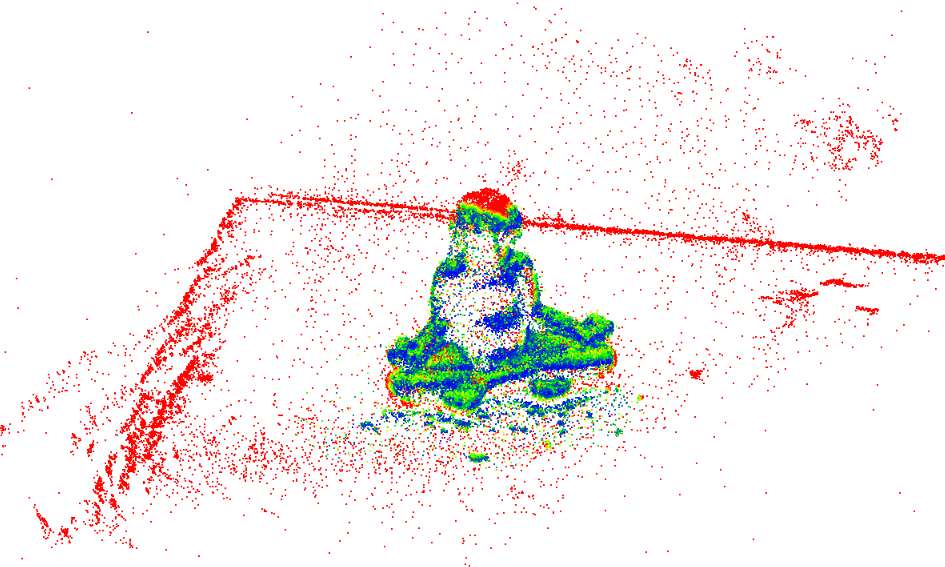}  \vspace{-2mm}\\[2ex]
    
        \rotatebox{90}{\textbf{FeatureGS}} &
         \vspace{-4mm}
        \includegraphics[width=0.225\linewidth]{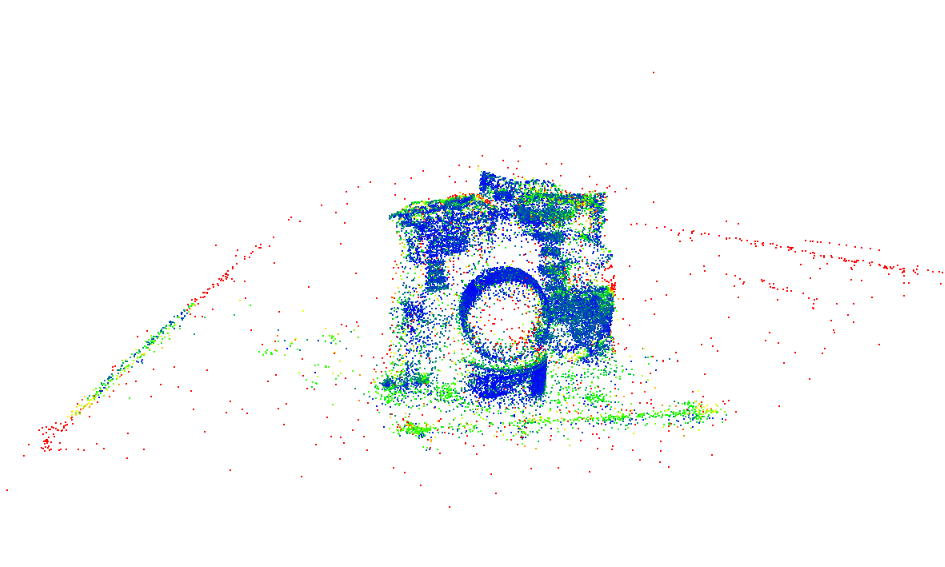} &
        \hspace{-4mm}
        \includegraphics[width=0.225\linewidth]{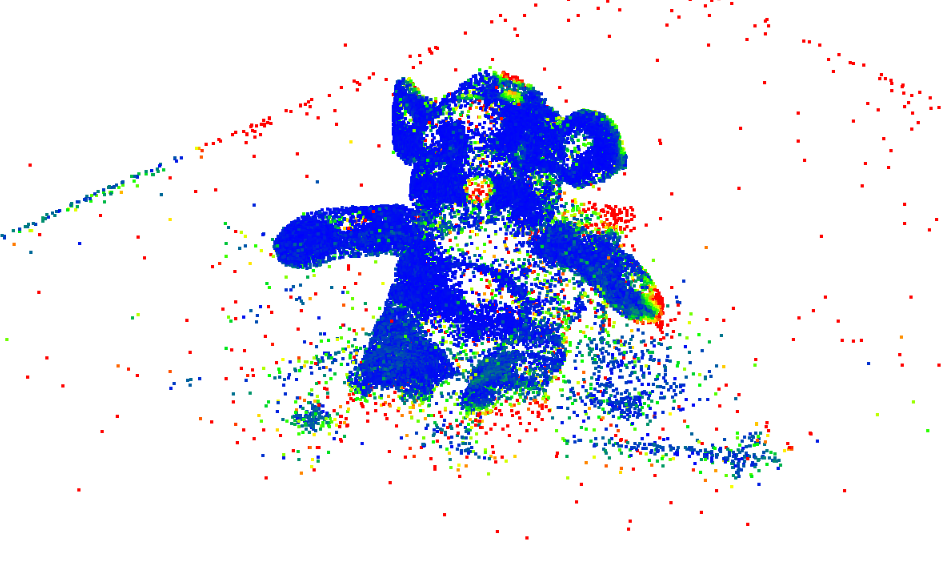} &
        \hspace{-4mm}
        \includegraphics[width=0.225\linewidth]{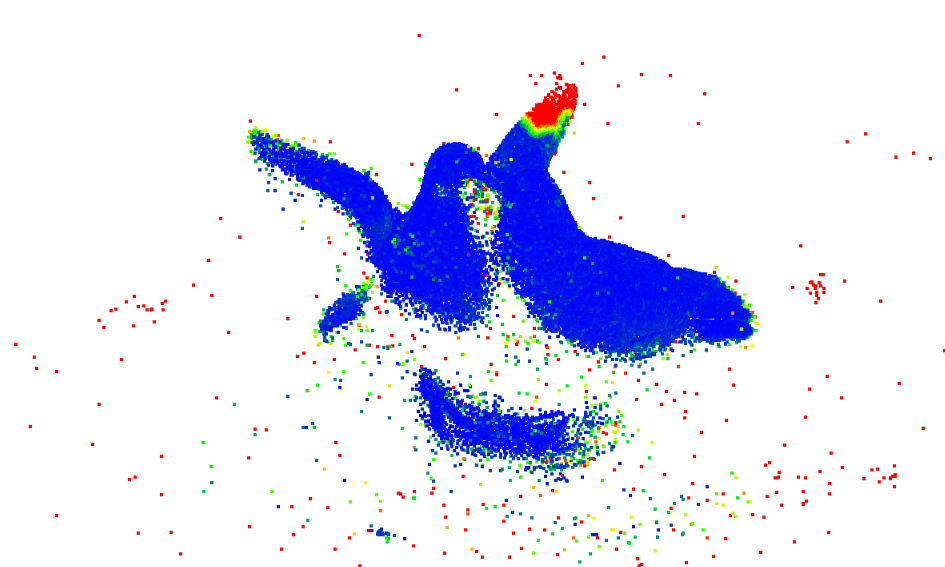} &
        \hspace{-4mm}
        \includegraphics[width=0.225\linewidth]{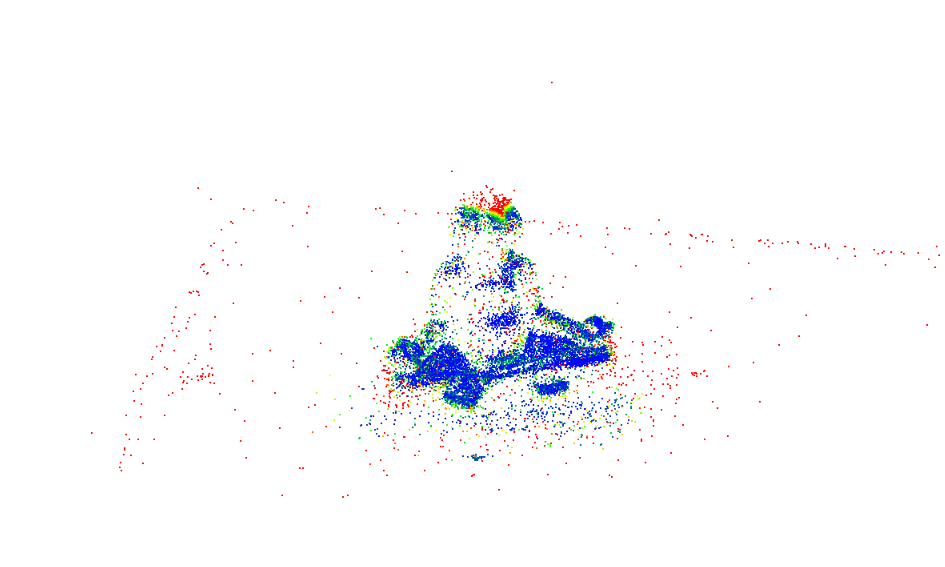} \\[2ex]

        \textbf{} & \textbf{scan114} & \textbf{scan118} & \textbf{scan122} & \\[1ex]
        
        \rotatebox{90}{\textbf{3DGS}} &
         \vspace{-4mm}
        \includegraphics[width=0.225\linewidth]{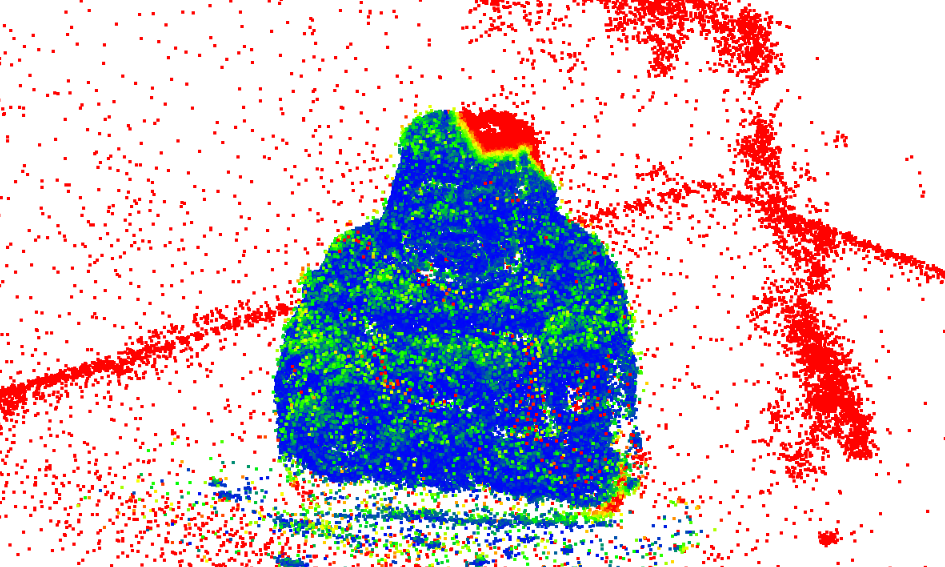} &
        \hspace{-4mm}
        \includegraphics[width=0.225\linewidth]{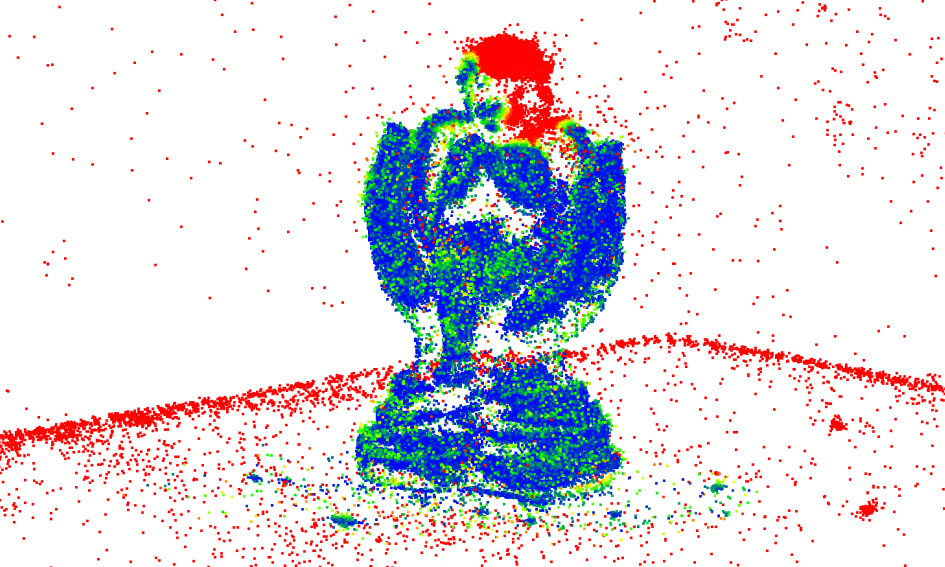} &
       \hspace{-4mm}
        \includegraphics[width=0.225\linewidth]{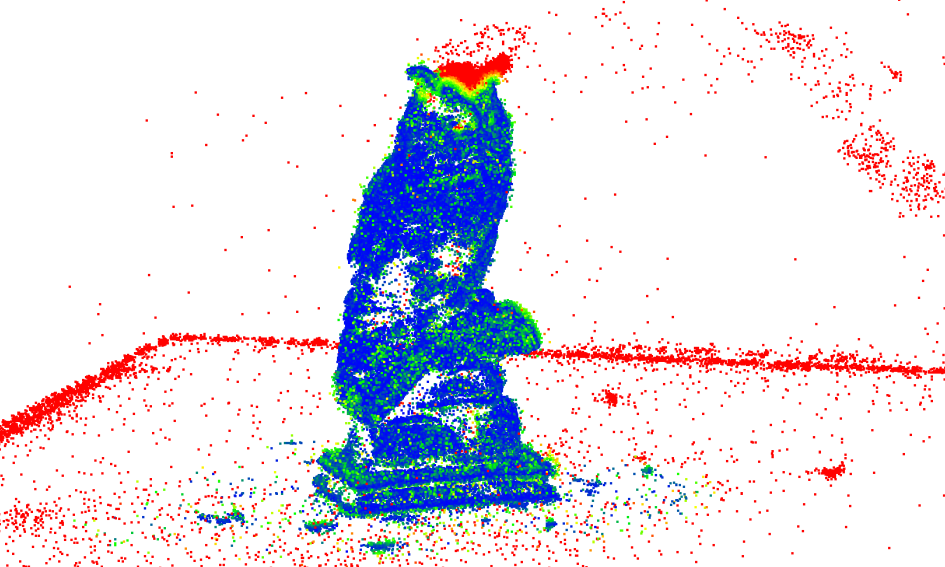} &
        \hspace{3cm}\\[2ex]
    
        \rotatebox{90}{\textbf{FeatureGS}} &
         \vspace{-4mm}
        \includegraphics[width=0.225\linewidth]{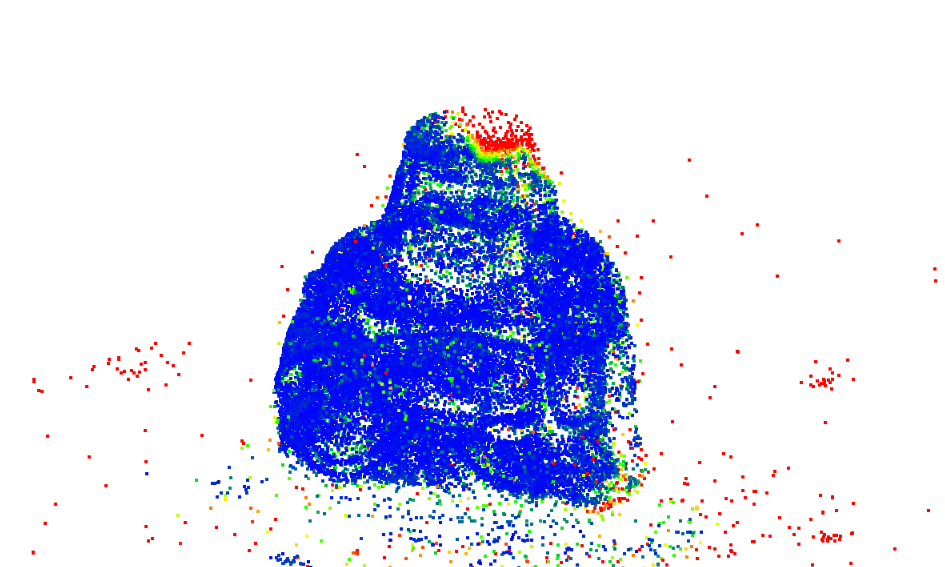} &
        \hspace{-4mm}
        \includegraphics[width=0.225\linewidth]{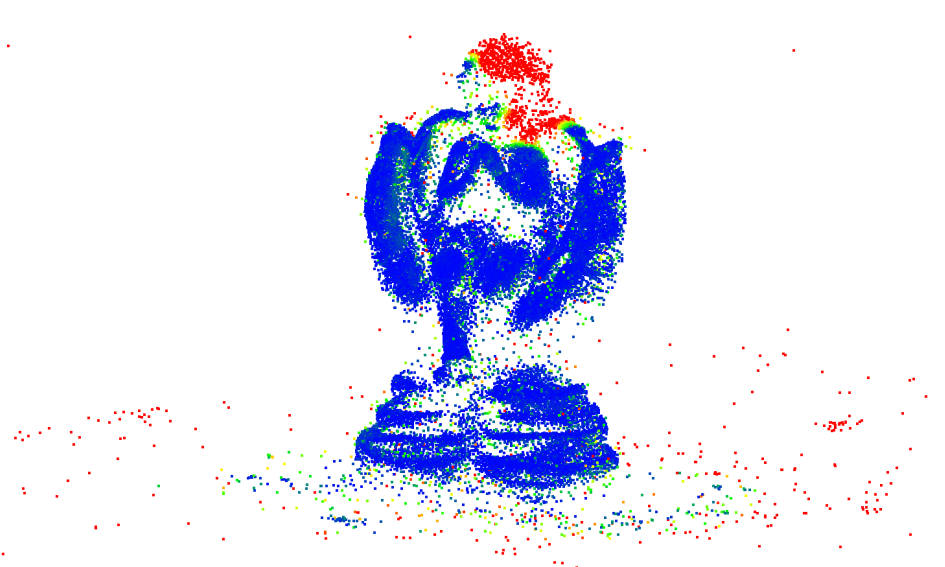} &
        \hspace{-4mm}
        \includegraphics[width=0.225\linewidth]{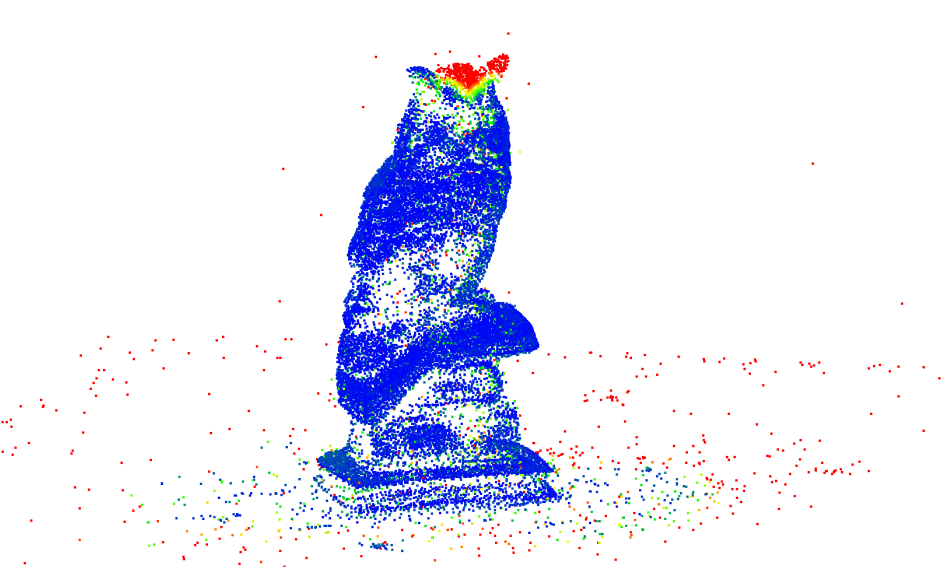} &
        \hspace{3cm} \\[2ex]

    \end{tabular}
    \includegraphics[width=0.2\textwidth]{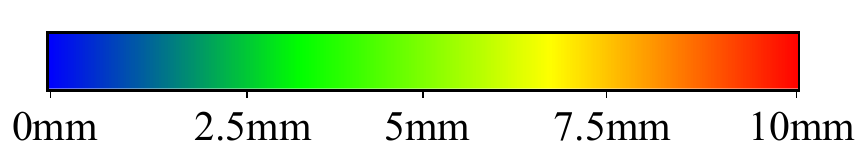}
    \caption{\textbf{Geometric accuracy} comparison on the DTU dataset with Chamfer cloud-to-cloud distances $\downarrow$ for the \textbf{same PSNR}. Color values are cropped at 10mm distance. }
     \label{fig:Qualitative_c2c}
\end{figure*}

The geometric accuracy of Gaussian centers, evaluated using the Chamfer cloud-to-cloud distance (Figure \ref{fig:Qualitative_c2c}), highlights the superior performance of FeatureGS compared to 3DGS. For the FeatureGS, the configurations that yielded the highest surface accuracy for the respective scenes are visualized. It is important to note that the reference point clouds are incomplete, which leads to high values on all object edges. On the one hand, it shows that the accuracy of the surface points is higher for FeatureGS, achieving submillimeter accuracy. Furthermore, the surface points appear less noisy. On the other hand, the drastic reduction in floater artifacts is striking, whereby floater artifacts prevent the reconstruction of the geometry via direct extraction of the Gaussian centers. While 3DGS leads to a lot of floater artifacts in all scenes, the scenes with FeatureGS are almost artifact-free.

\paragraph{Rendering Quality}

\begin{figure*}[htbp]
\vspace{-10mm}
    \centering
    \begin{tabular}{c c c c c}
    
        \textbf{} & \textbf{scan24} & \textbf{scan37} & \textbf{scan40} & \textbf{scan55} \\[1ex]
        
        \rotatebox{90}{\textbf{3DGS}} &
        \vspace{-5mm}
        \includegraphics[width=0.22\linewidth]{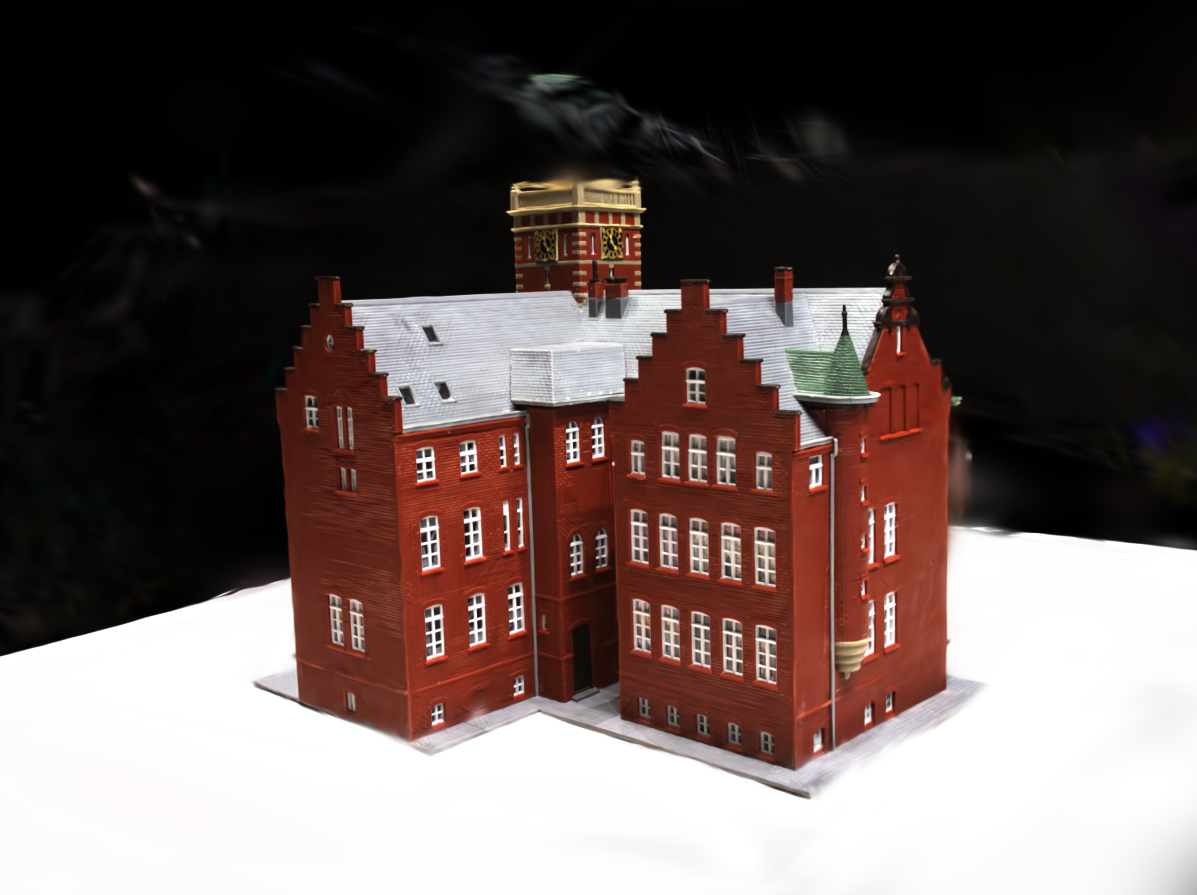} &
      \hspace{-4mm}
        \includegraphics[width=0.22\linewidth]{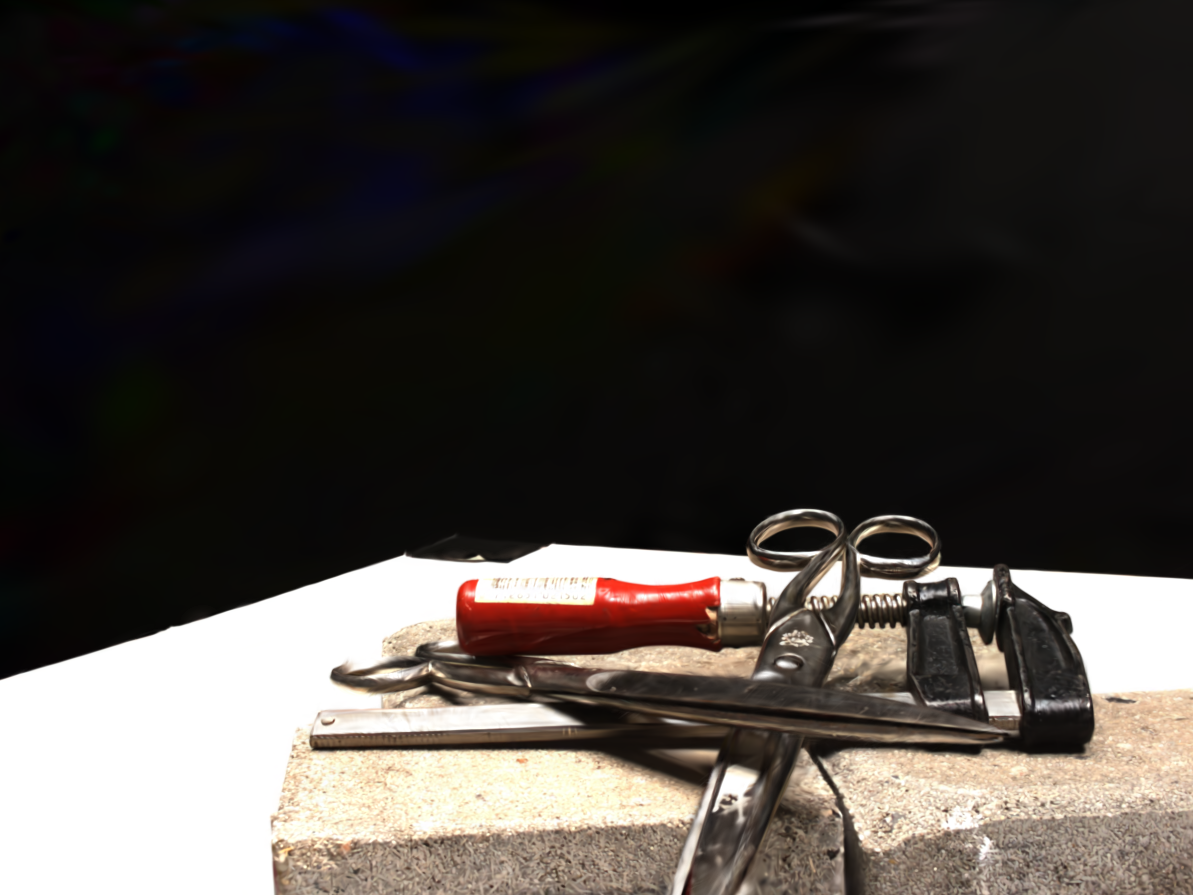} &
         \hspace{-4mm}
        \includegraphics[width=0.22\linewidth]{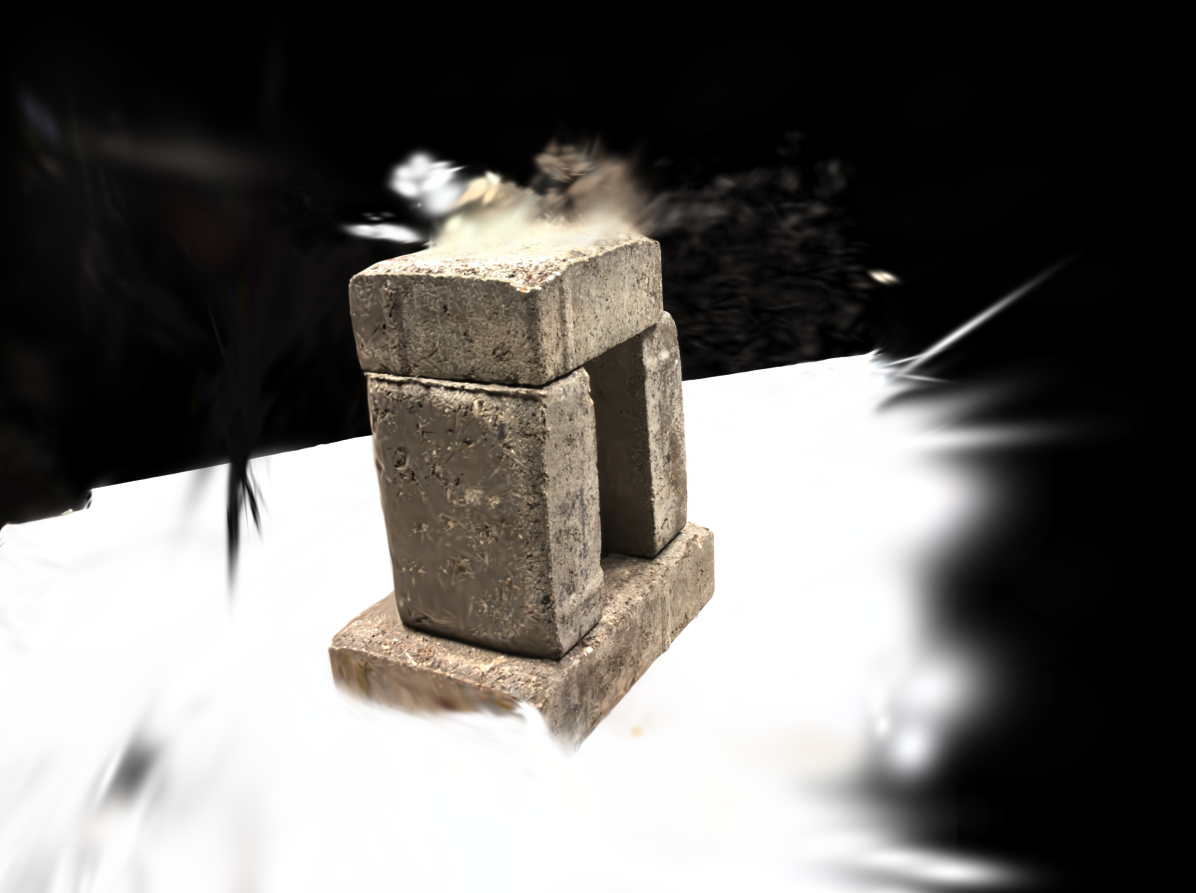} &
        \hspace{-4mm}
        \includegraphics[width=0.22\linewidth]{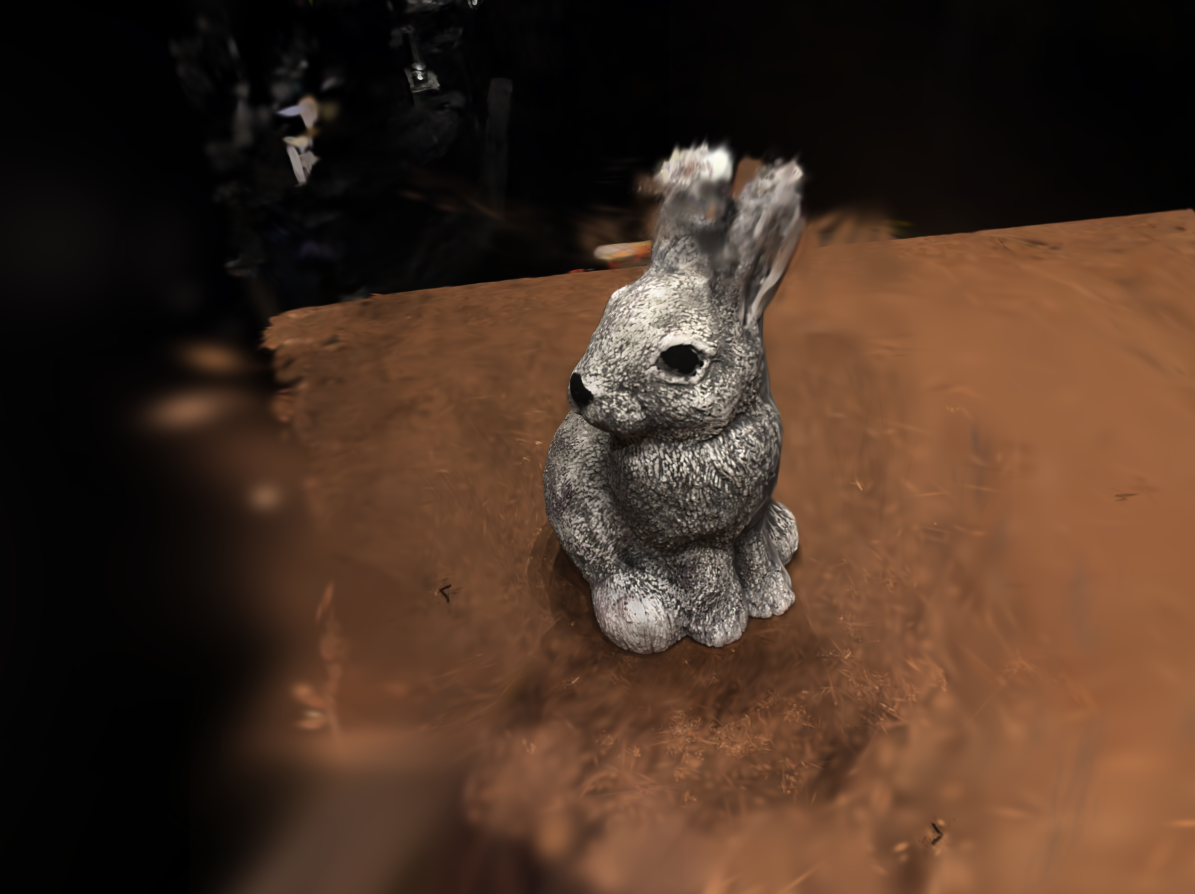} \\[2ex]
  
        \rotatebox{90}{\textbf{FeatureGS}} &
         \vspace{-5mm}
        \includegraphics[width=0.22\linewidth]{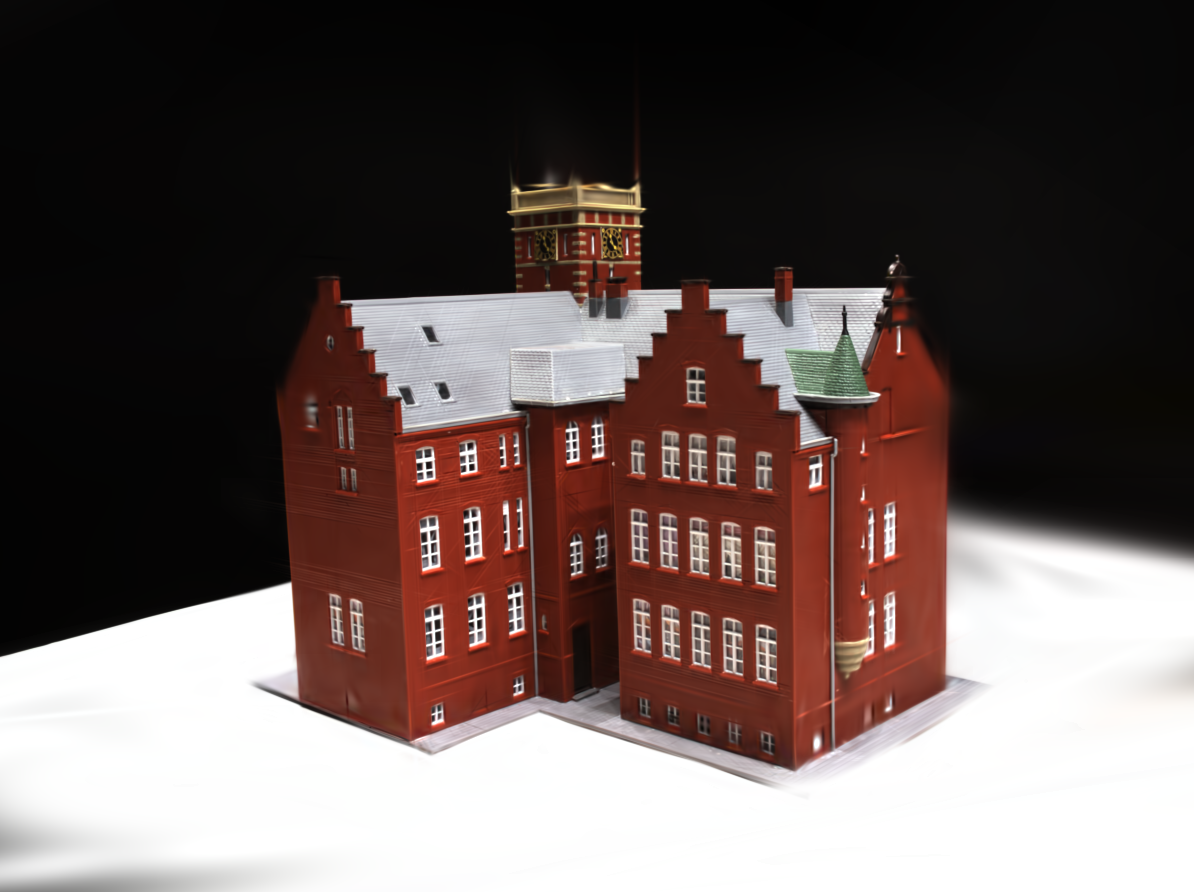} &
      \hspace{-4mm}
        \includegraphics[width=0.22\linewidth]{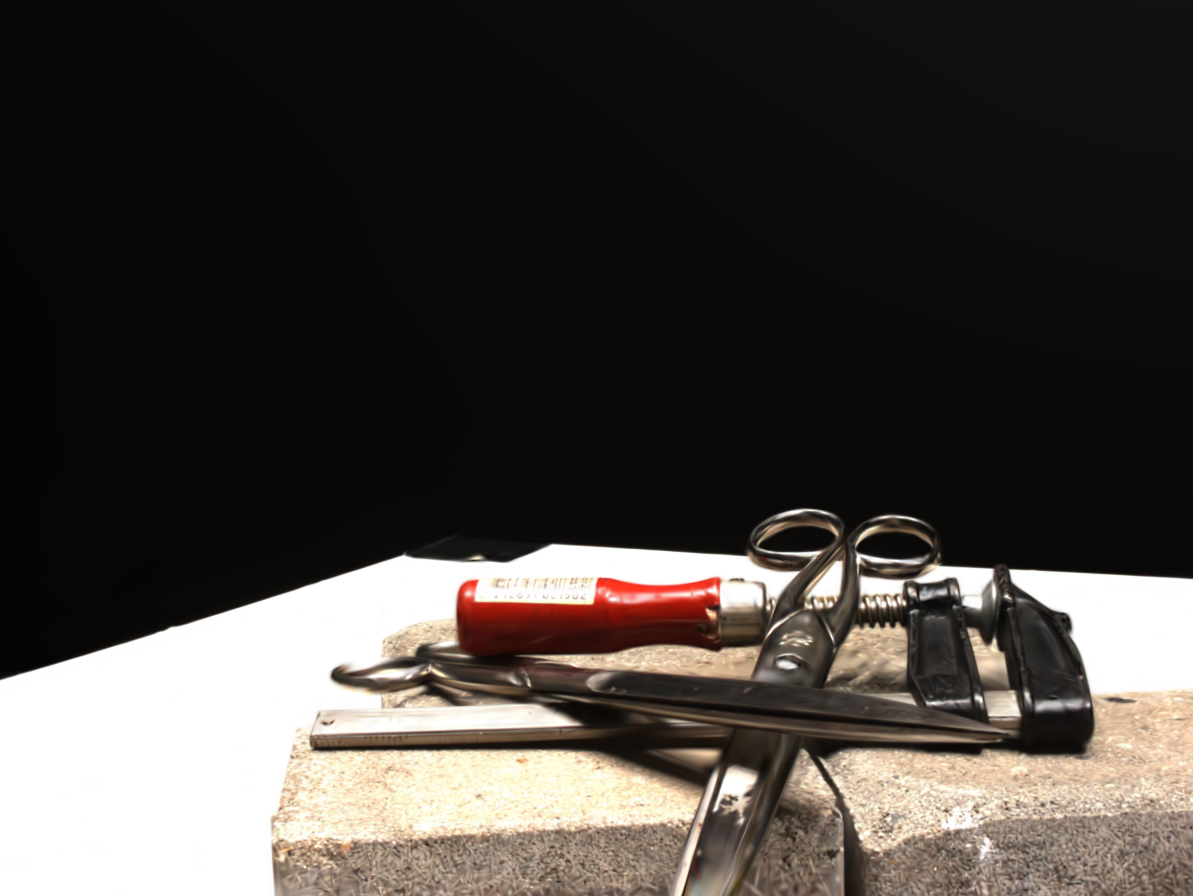} &
         \hspace{-4mm}
        \includegraphics[width=0.22\linewidth]{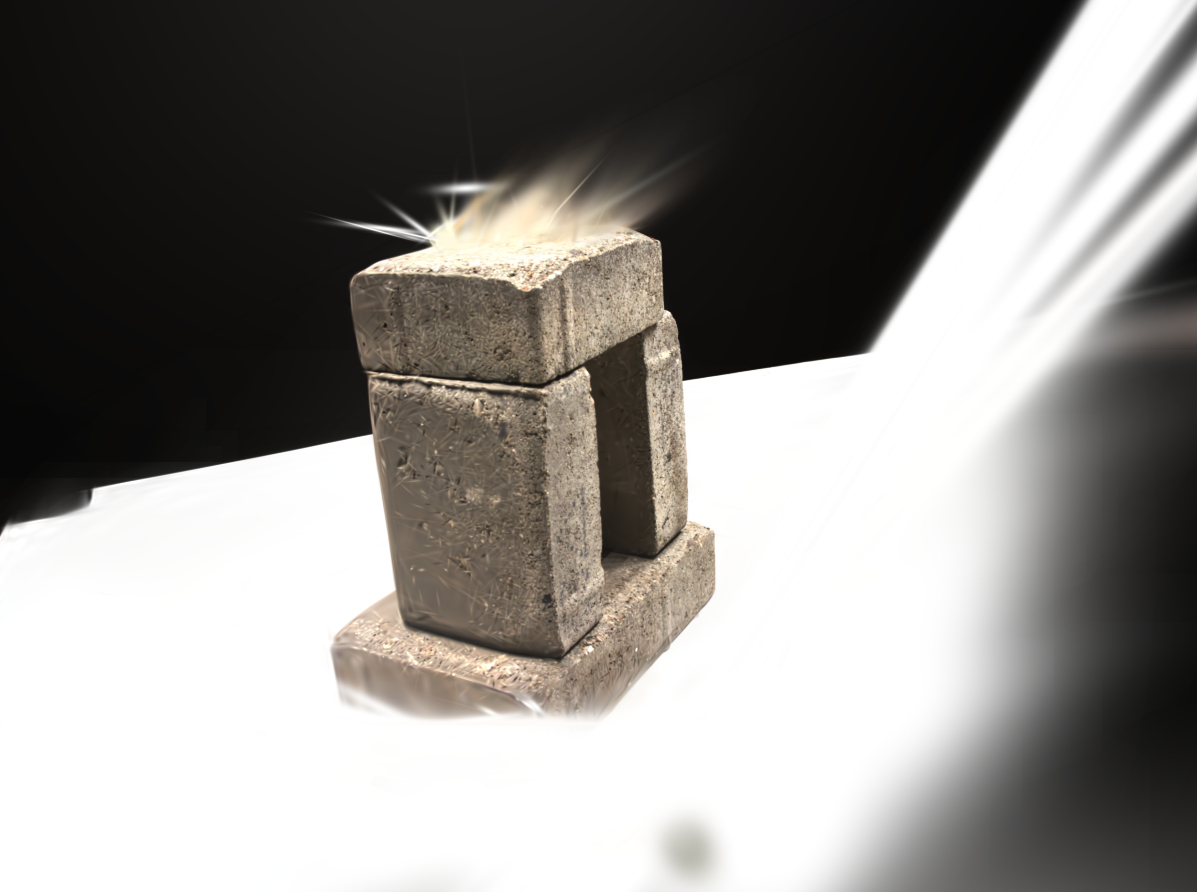} &
      \hspace{-4mm}
        \includegraphics[width=0.22\linewidth]{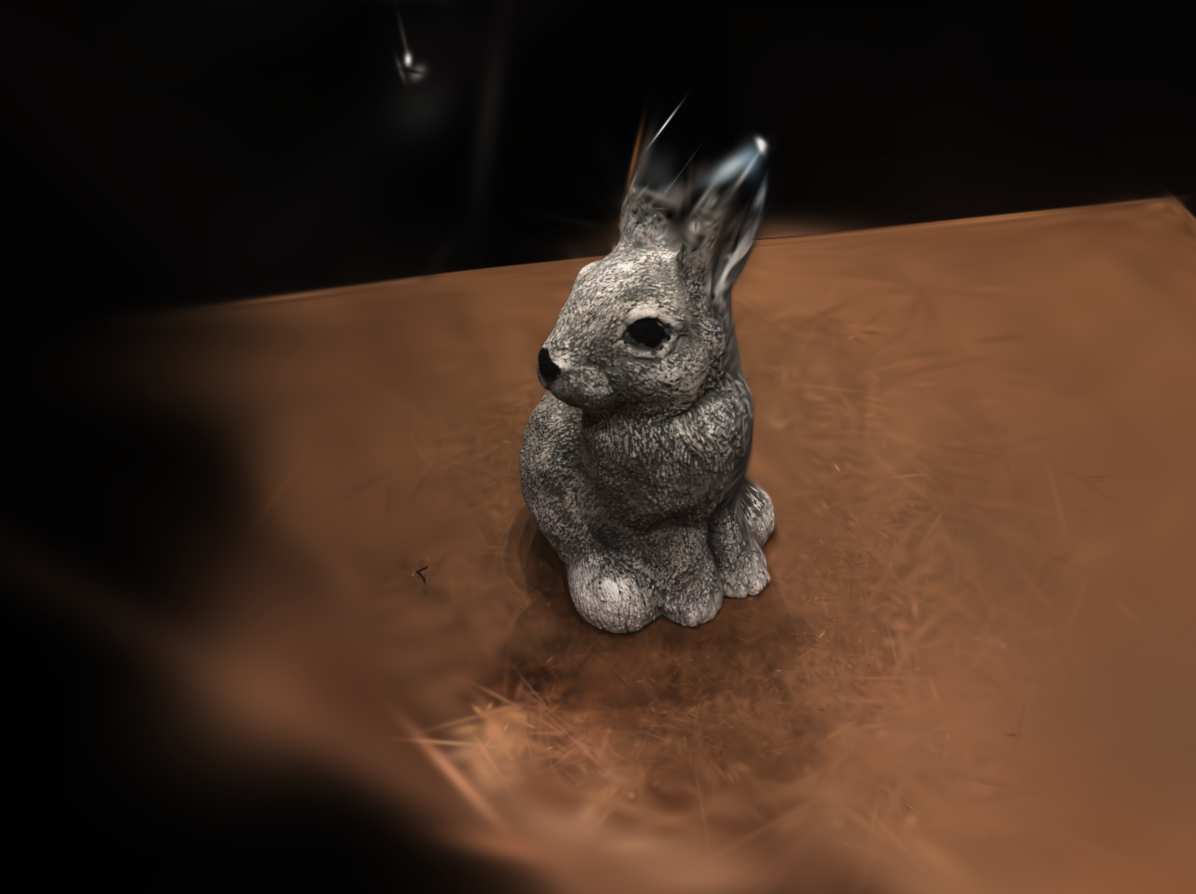} \\[2ex]

        \textbf{} & \textbf{scan63} & \textbf{scan65} & \textbf{scan69} & \textbf{scan83}\\[1ex]
        
        \rotatebox{90}{\textbf{3DGS}} &
         \vspace{-5mm}
        \includegraphics[width=0.22\linewidth]{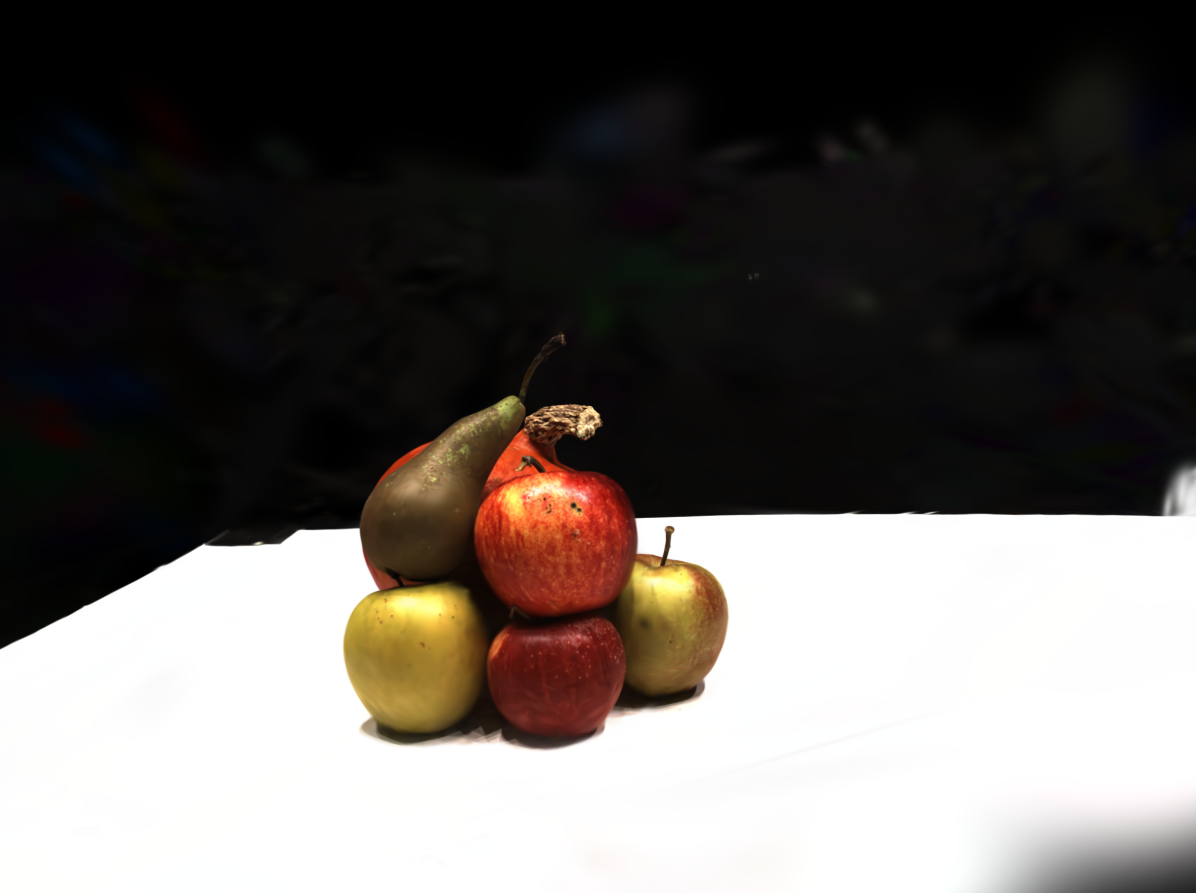} &
        \hspace{-4mm}
        \includegraphics[width=0.22\linewidth]{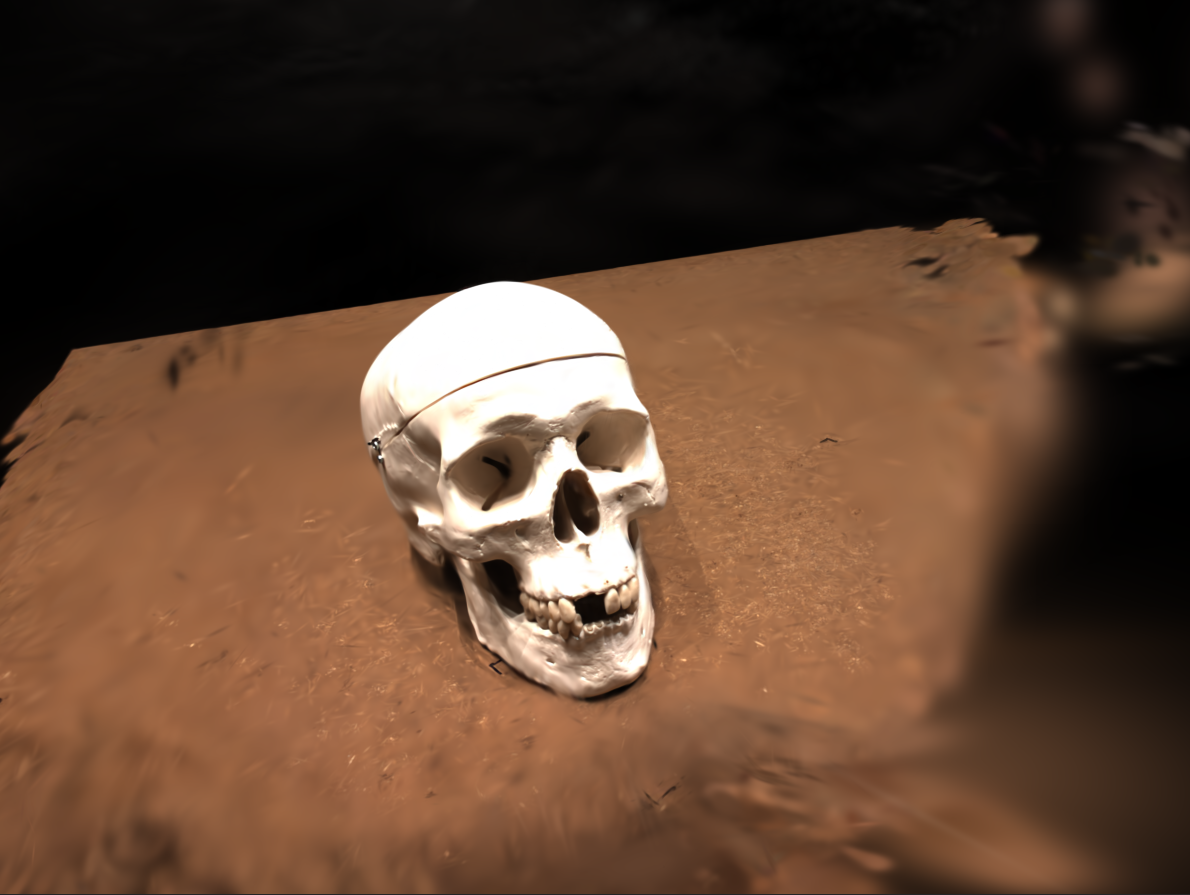} &
       \hspace{-4mm}
        \includegraphics[width=0.22\linewidth]{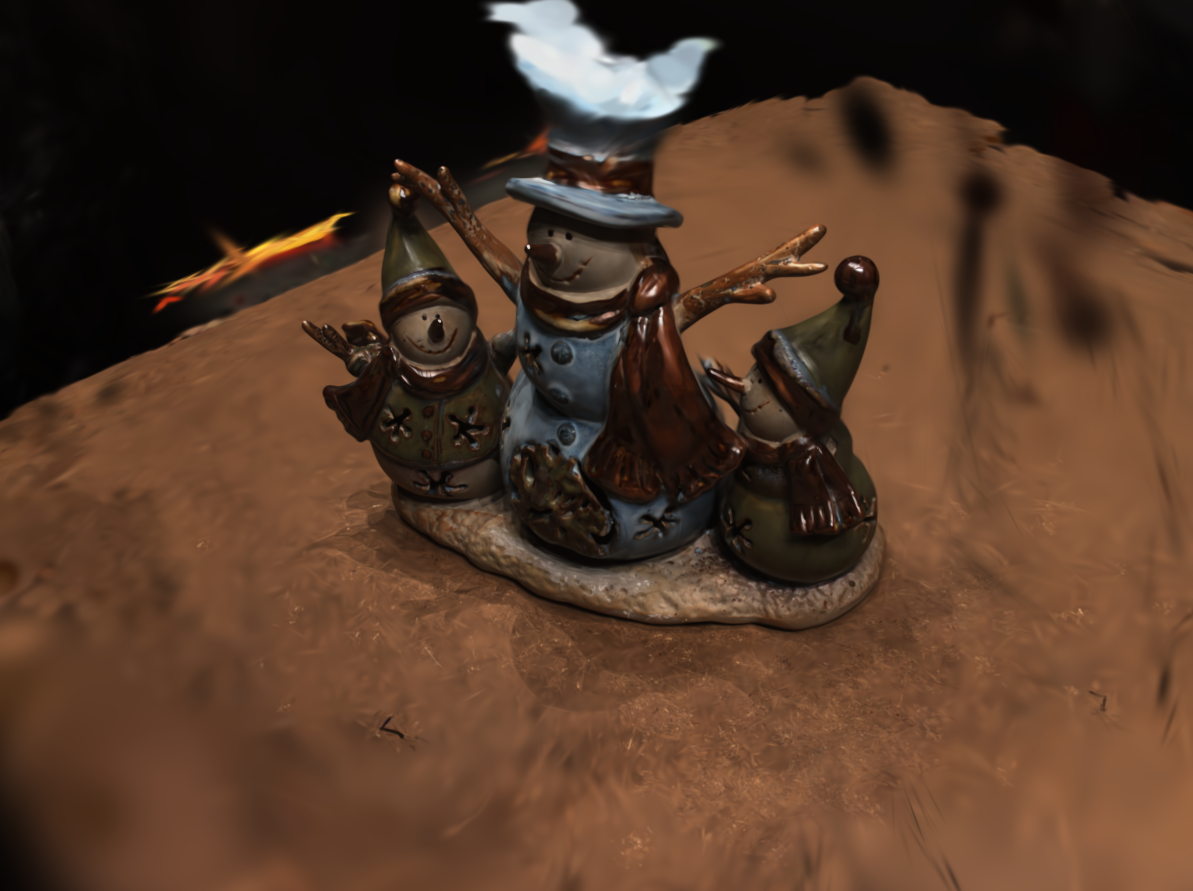} &
       \hspace{-4mm}
        \includegraphics[width=0.22\linewidth]{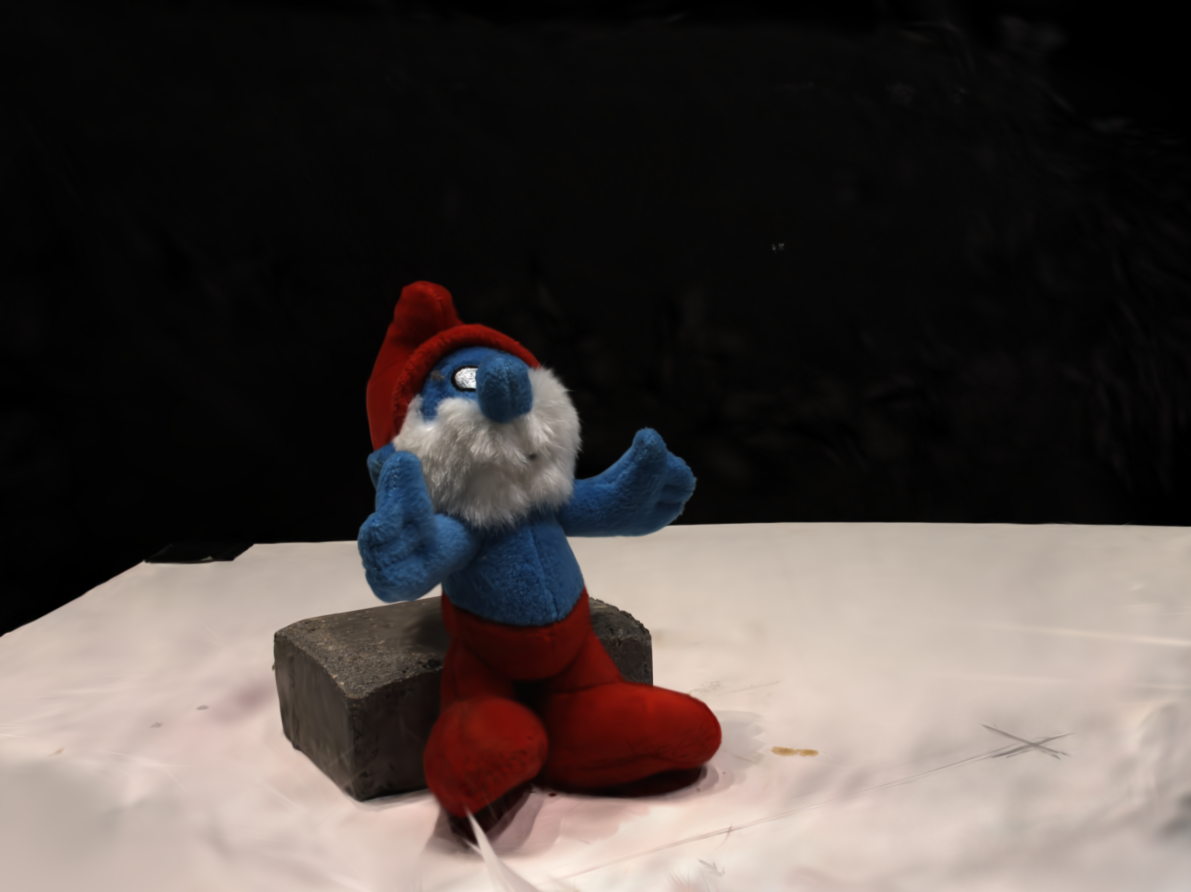} \\[2ex]
    
        \rotatebox{90}{\textbf{FeatureGS}} &
         \vspace{-5mm}
        \includegraphics[width=0.22\linewidth]{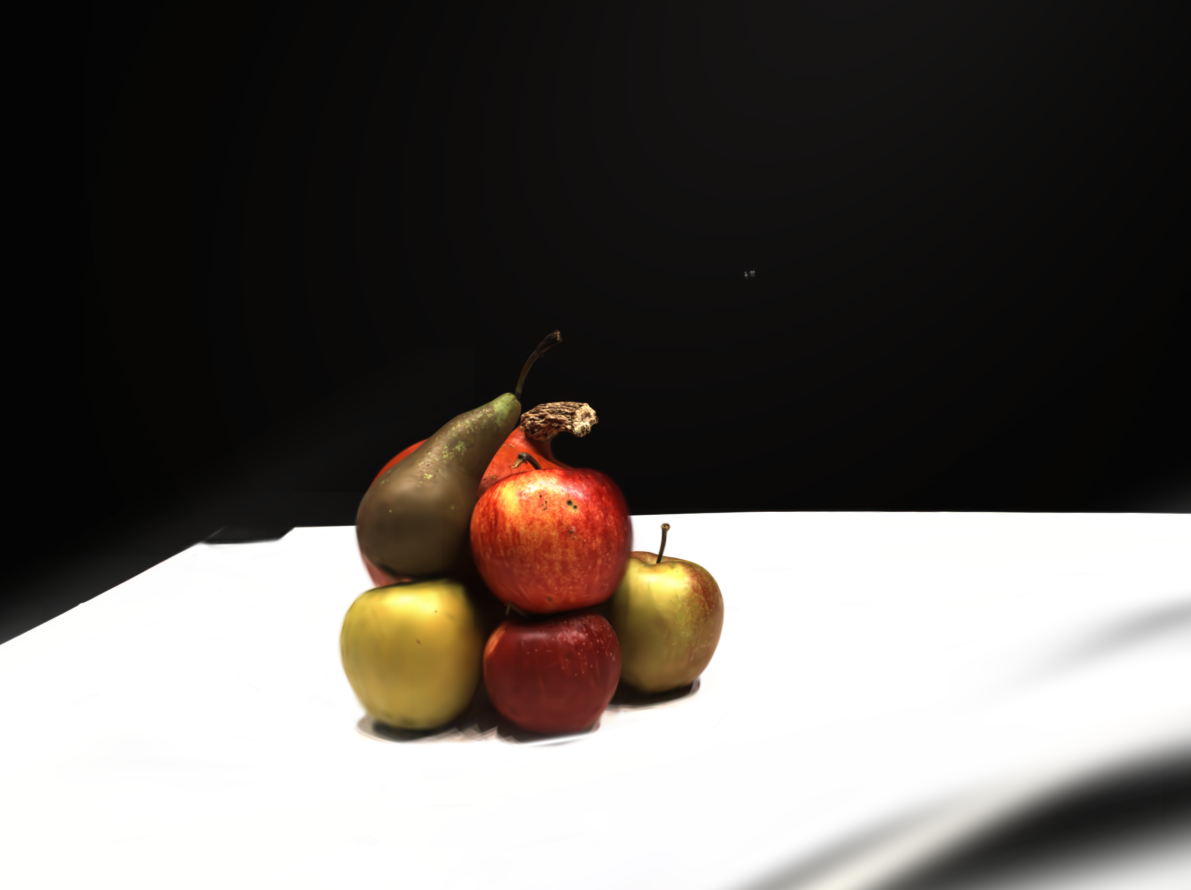} &
        \hspace{-4mm}
        \includegraphics[width=0.22\linewidth]{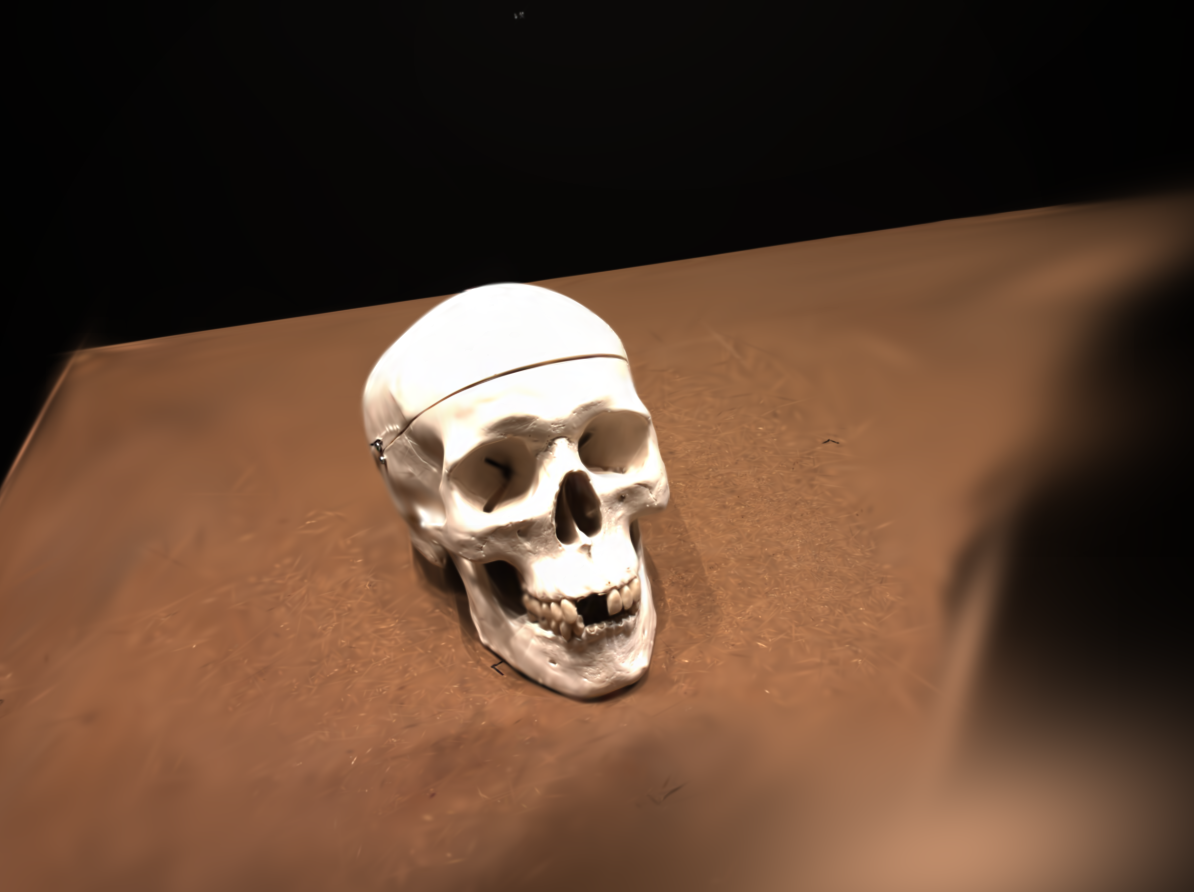} &
        \hspace{-4mm}
        \includegraphics[width=0.22\linewidth]{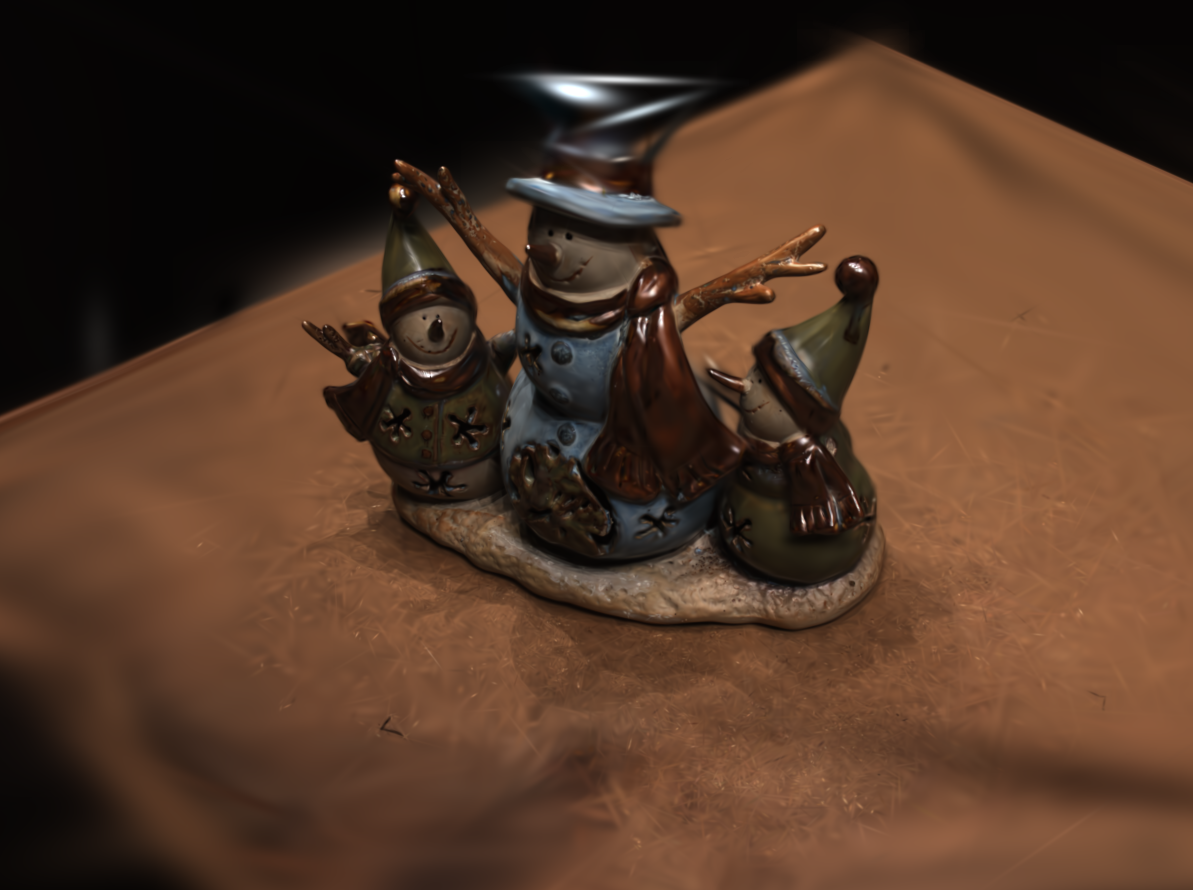} &
        \hspace{-4mm}
        \includegraphics[width=0.22\linewidth]{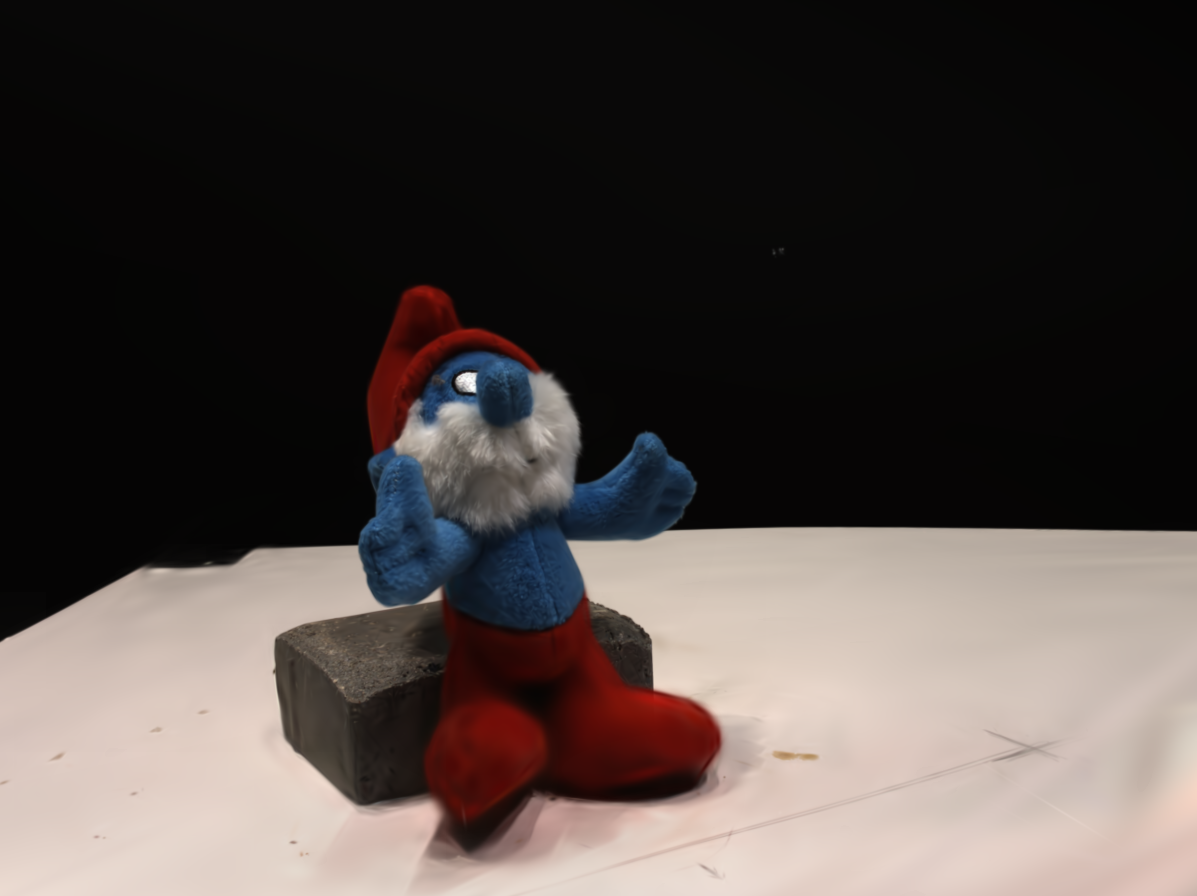} \\[2ex]

                \textbf{} & \textbf{scan97} & \textbf{scan105} & \textbf{scan106} & \textbf{scan110} \\[1ex]
        \rotatebox{90}{\textbf{3DGS}} &
         \vspace{-5mm}
        \includegraphics[width=0.22\linewidth]{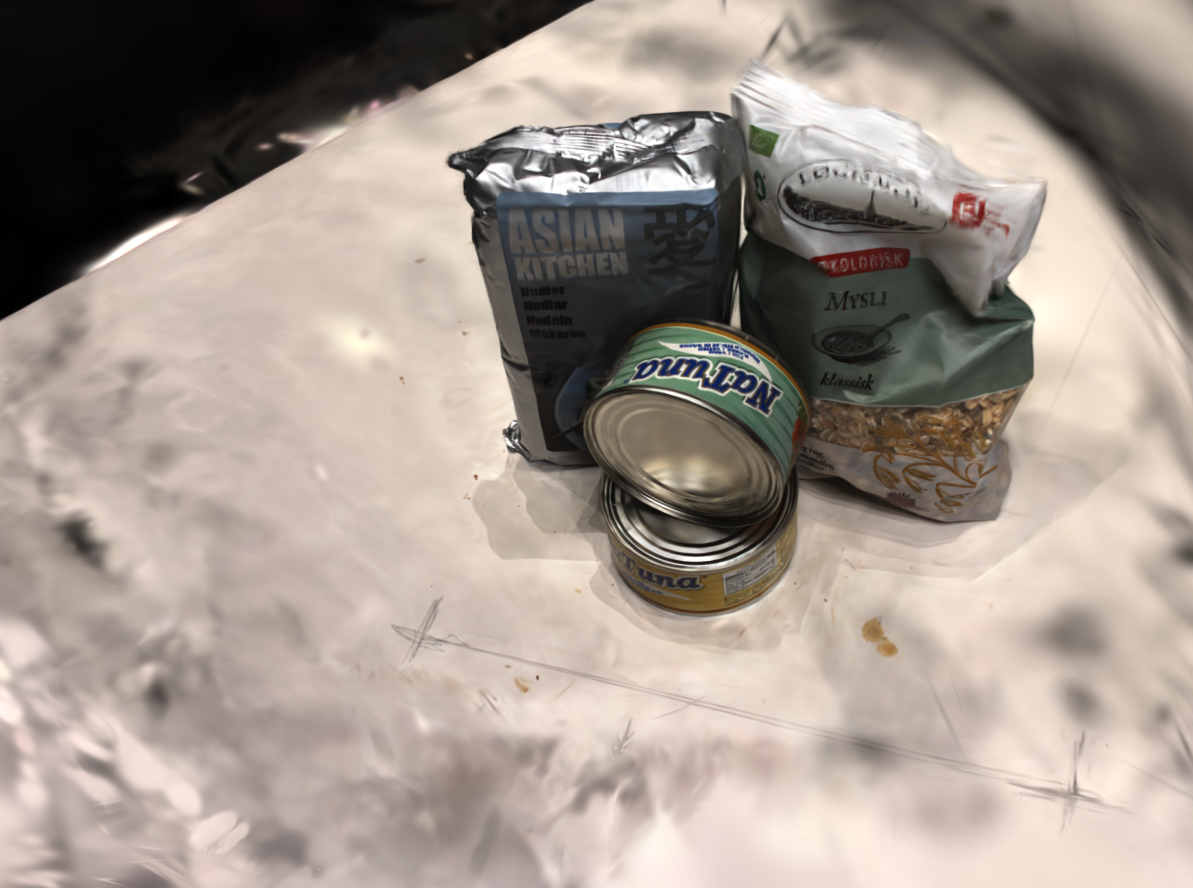} &
         \hspace{-4mm}
        \includegraphics[width=0.22\linewidth]{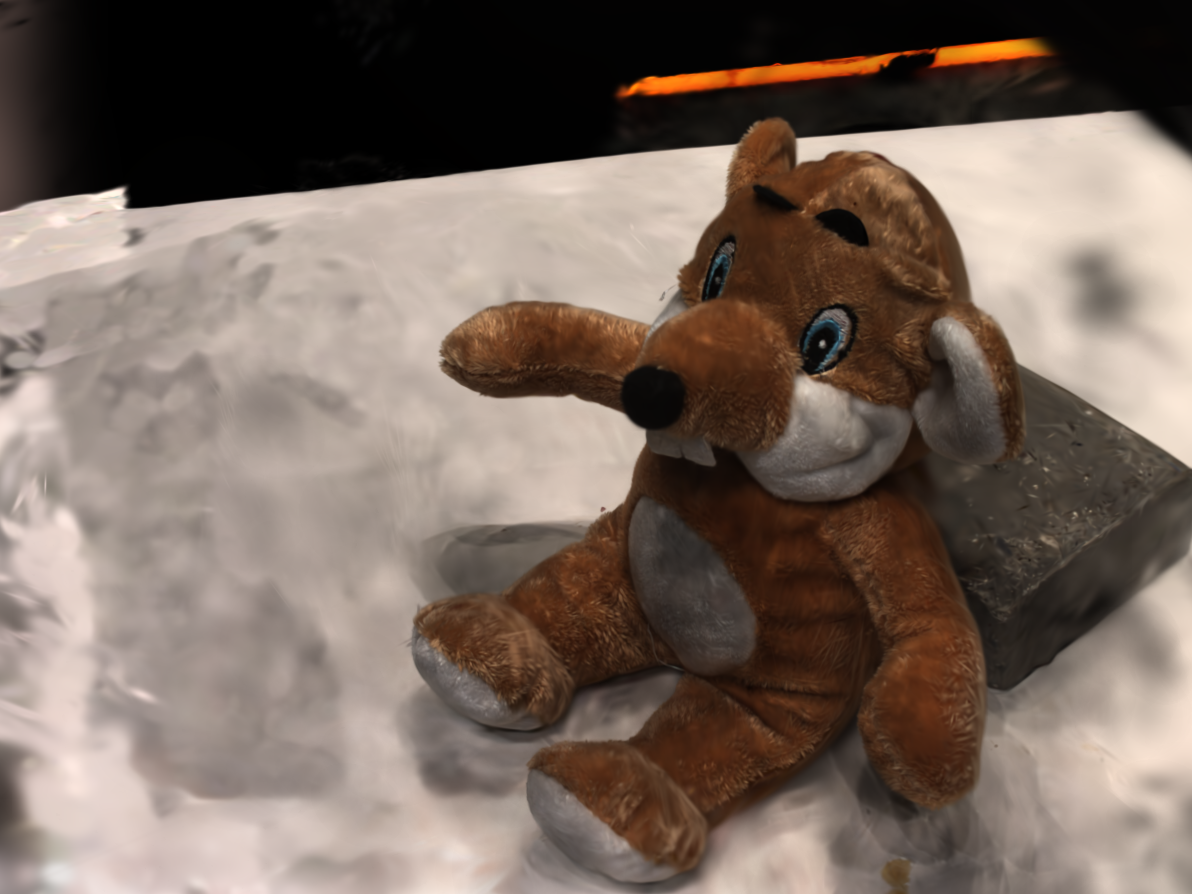} &
        \hspace{-4mm}
        \includegraphics[width=0.22\linewidth]{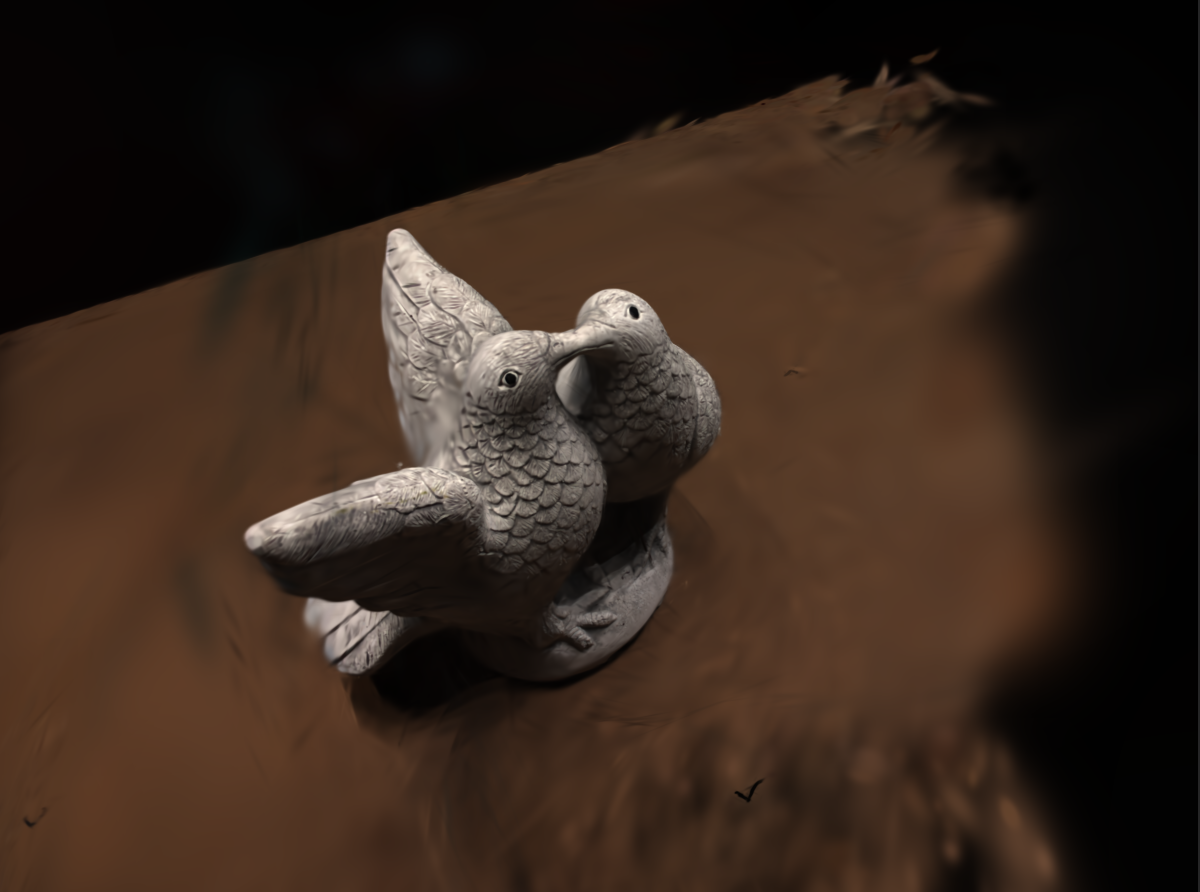} &
       \hspace{-4mm}
        \includegraphics[width=0.22\linewidth]{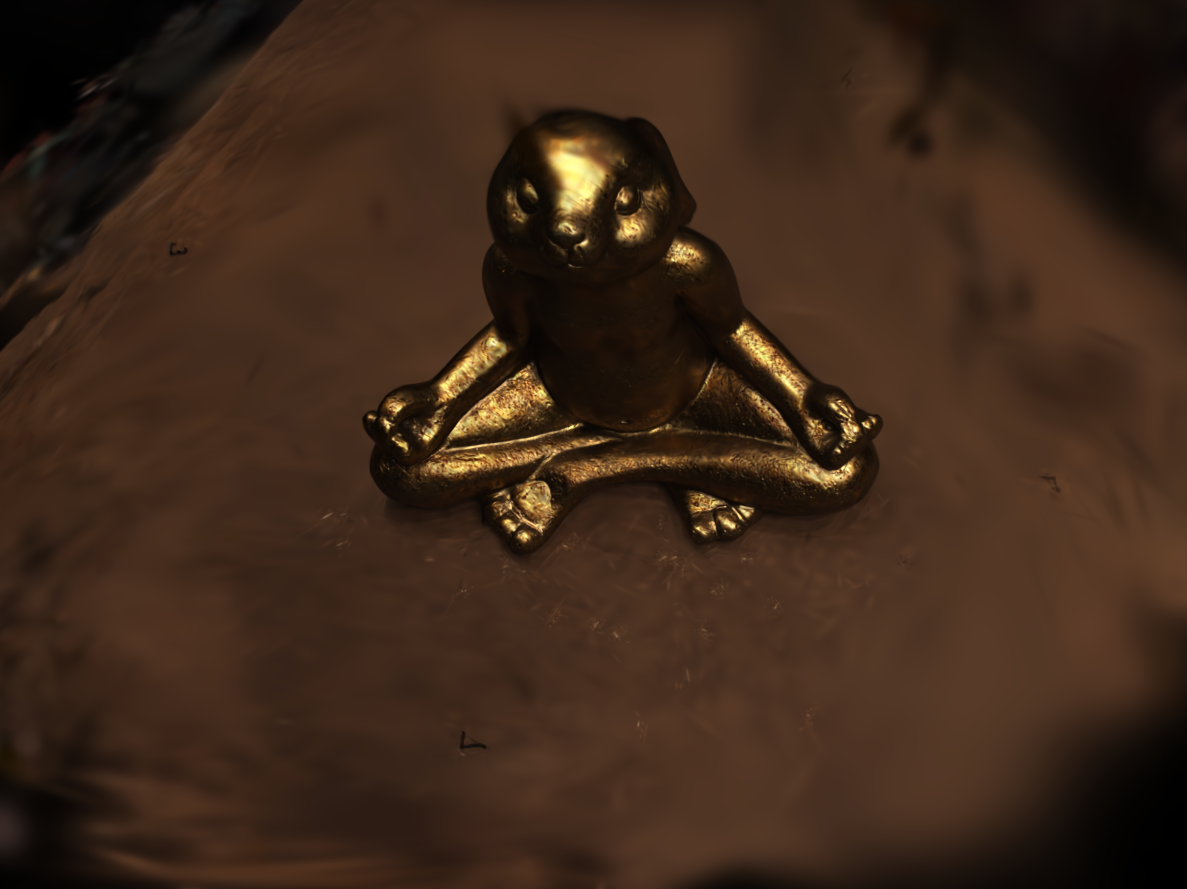} \\[2ex]
    
        \rotatebox{90}{\textbf{FeatureGS}} &
        \vspace{-5mm}
        \includegraphics[width=0.22\linewidth]{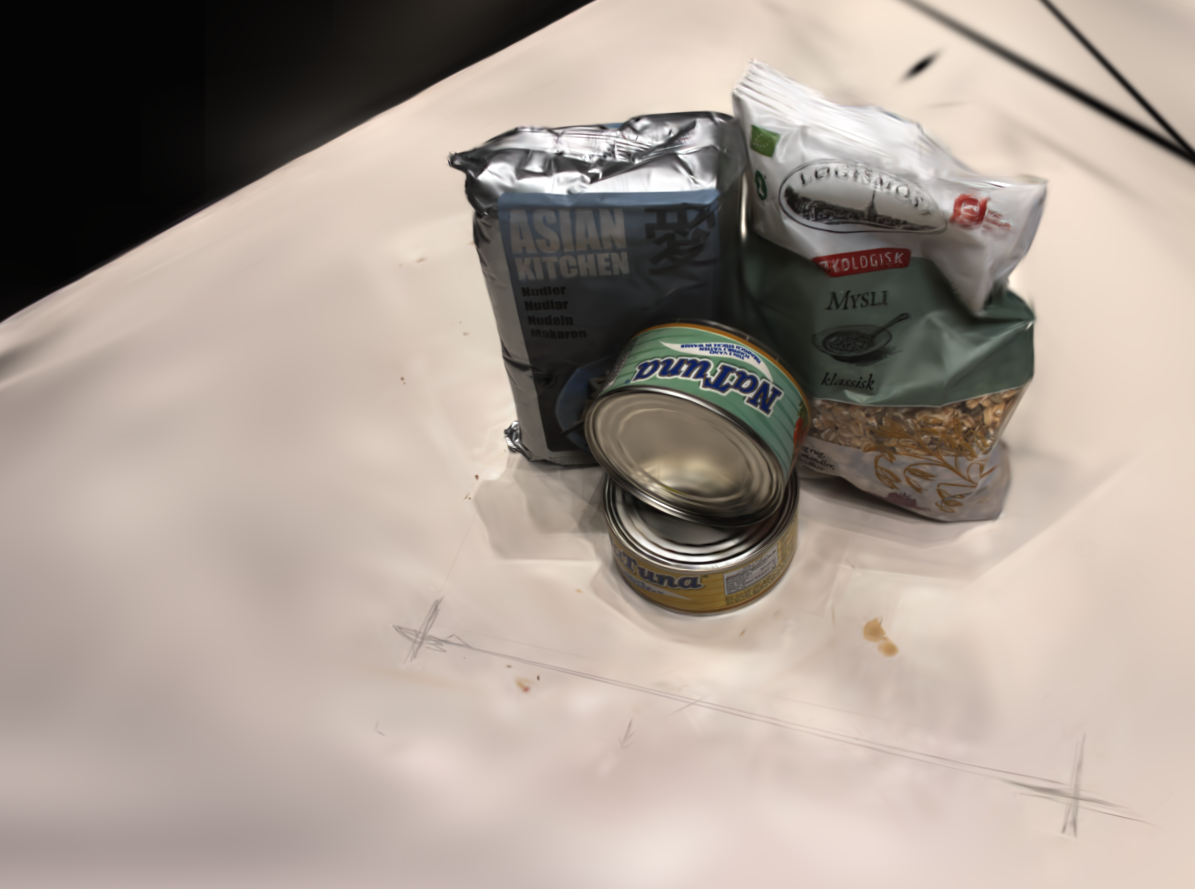}    &
         \hspace{-4mm}
        \includegraphics[width=0.22\linewidth]{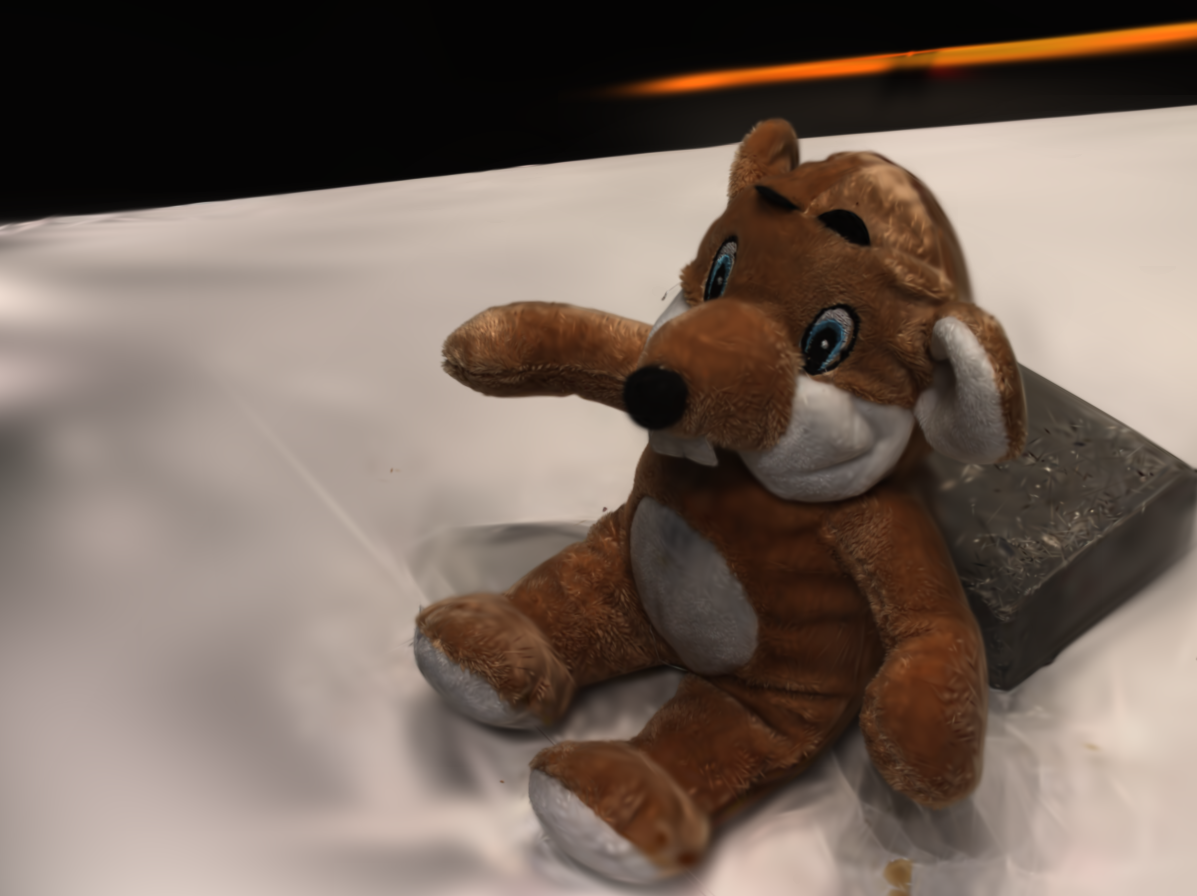} &
        \hspace{-4mm}
        \includegraphics[width=0.22\linewidth]{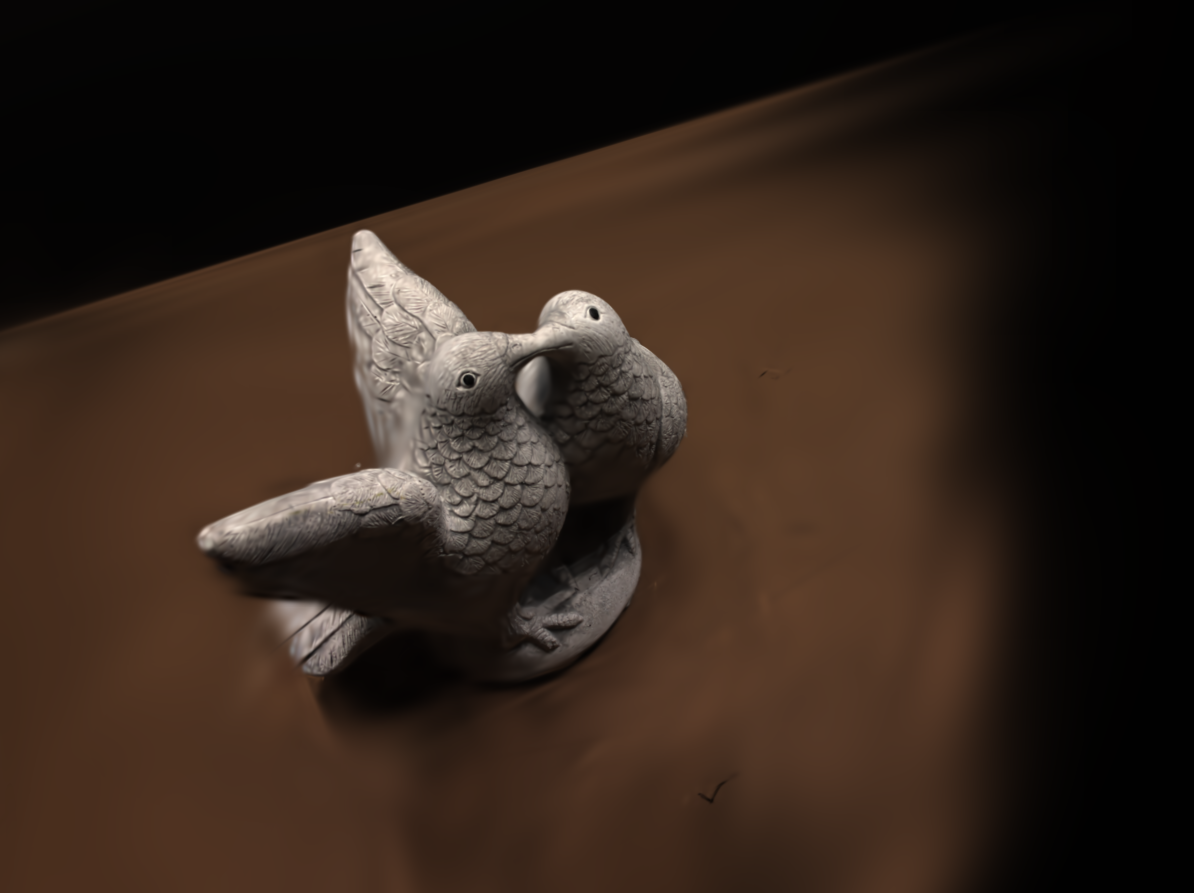} &
        \hspace{-4mm}
        \includegraphics[width=0.22\linewidth]{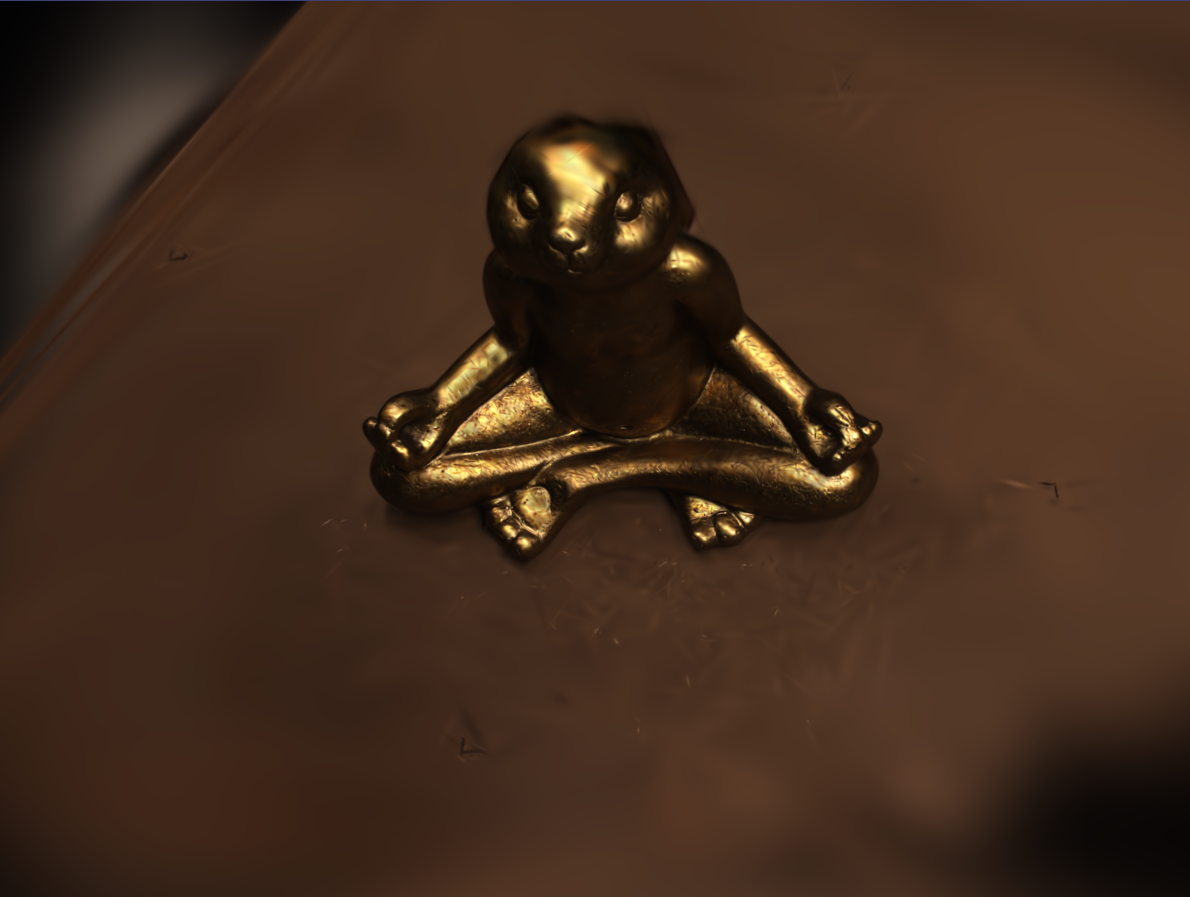} \\[2ex]

        \textbf{} & \textbf{scan114} & \textbf{scan118} & \textbf{scan122} &  \\[1ex]
        
        \rotatebox{90}{\textbf{3DGS}} &
         \vspace{-5mm}
        \includegraphics[width=0.22\linewidth]{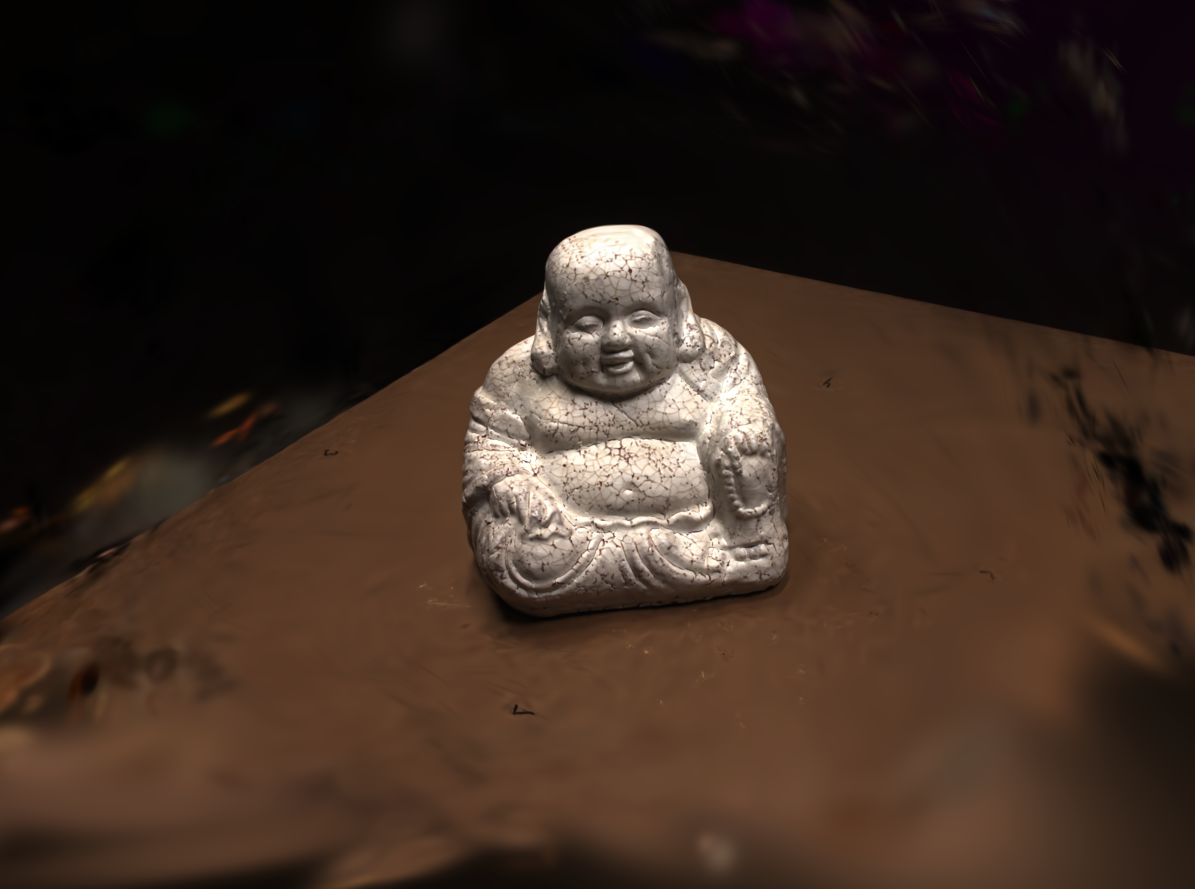} &
         \hspace{-4mm}
        \includegraphics[width=0.22\linewidth]{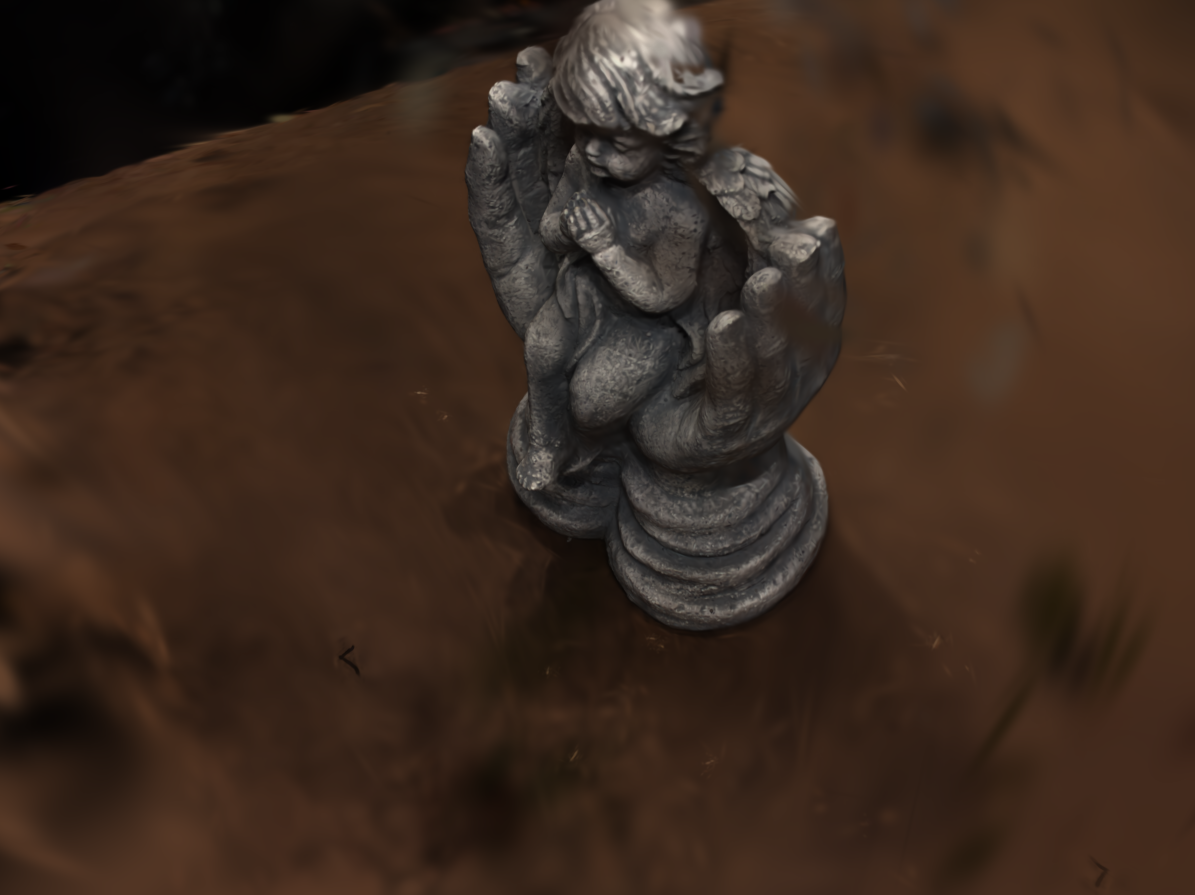} &
        \hspace{-4mm}
        \includegraphics[width=0.22\linewidth]{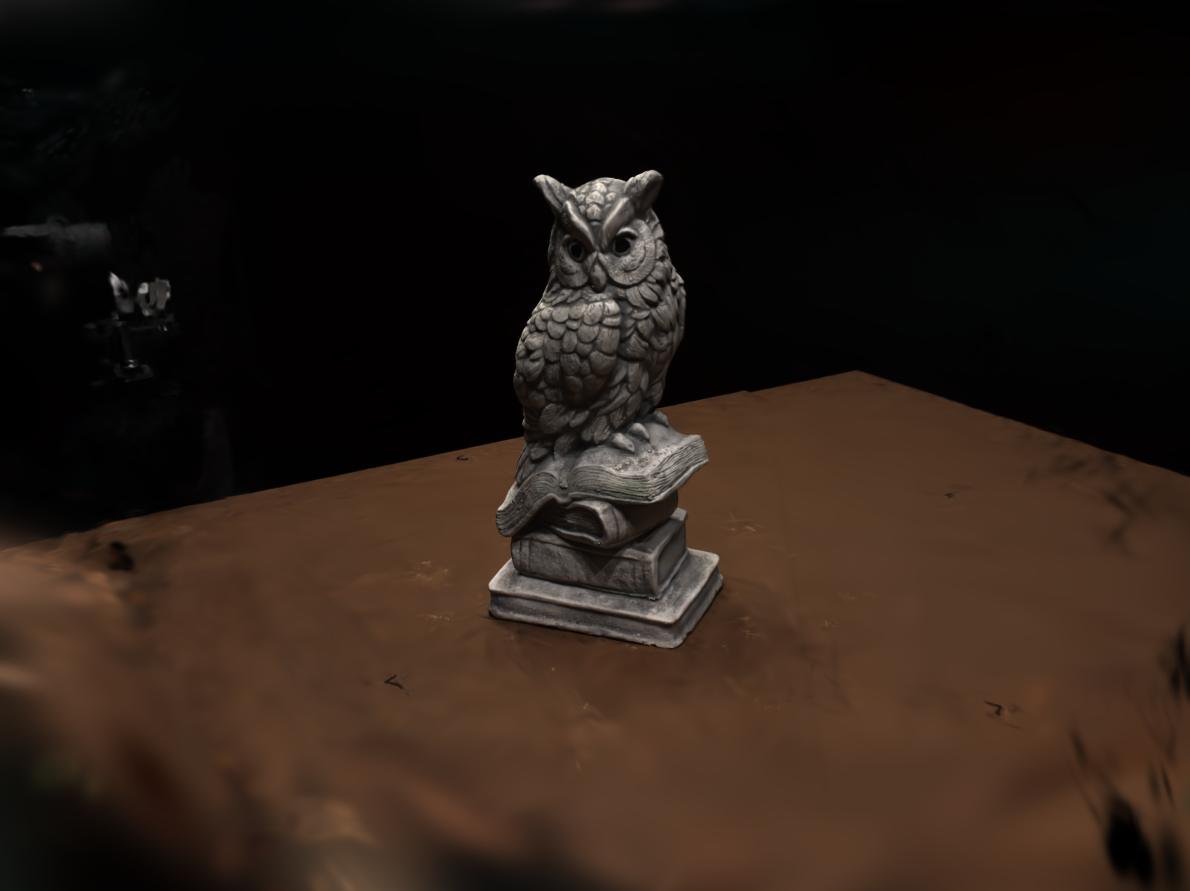} &
       \hspace{3cm}\\[2ex]
    
        \rotatebox{90}{\textbf{FeatureGS}} &
        \vspace{-5mm}
        \includegraphics[width=0.22\linewidth]{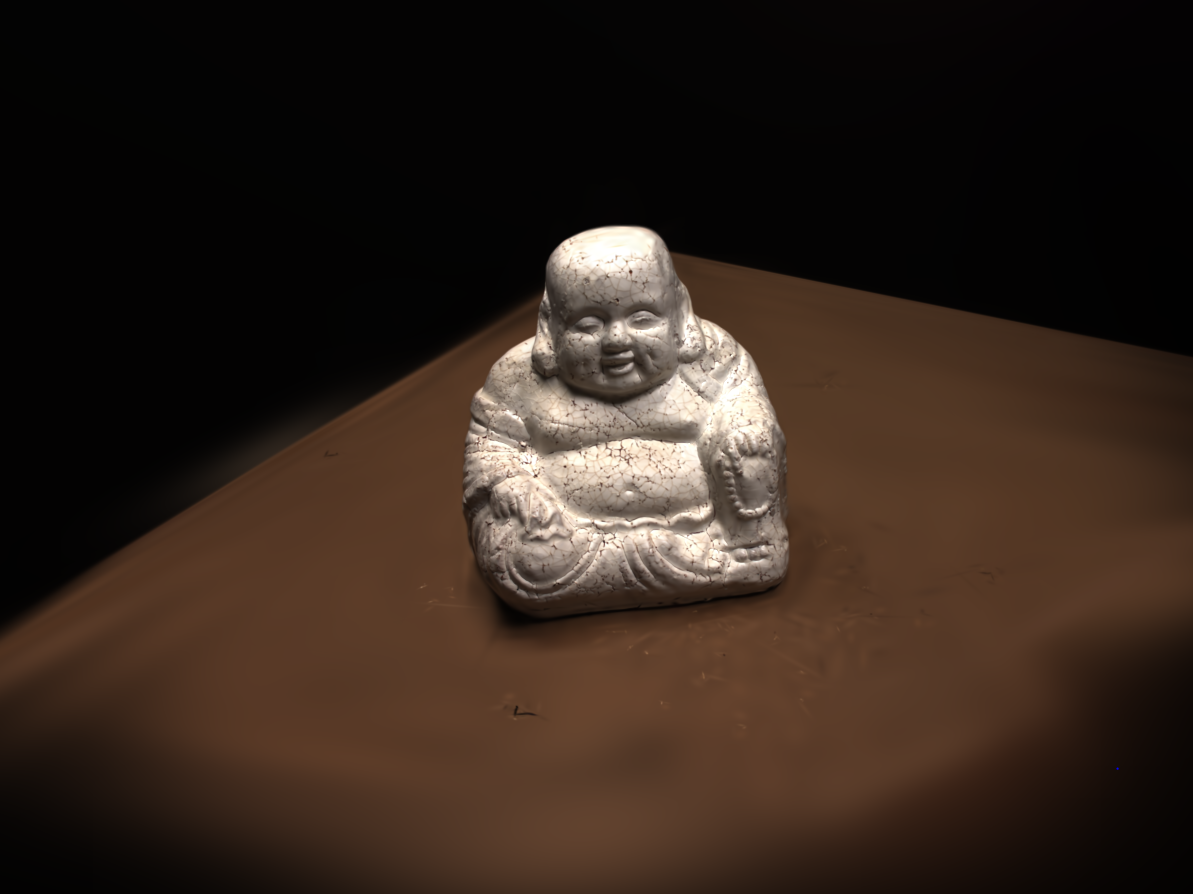}    &
         \hspace{-4mm}
        \includegraphics[width=0.22\linewidth]{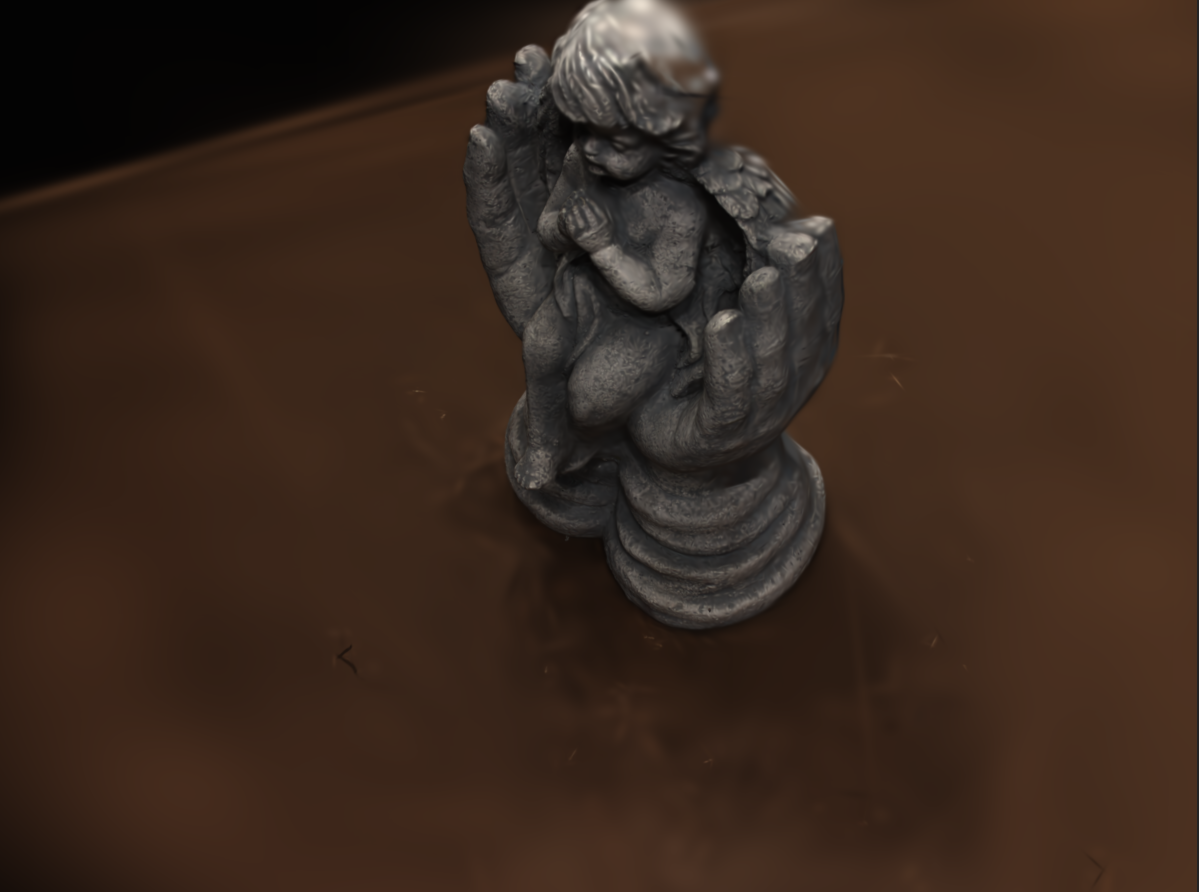} &
        \hspace{-4mm}
        \includegraphics[width=0.22\linewidth]{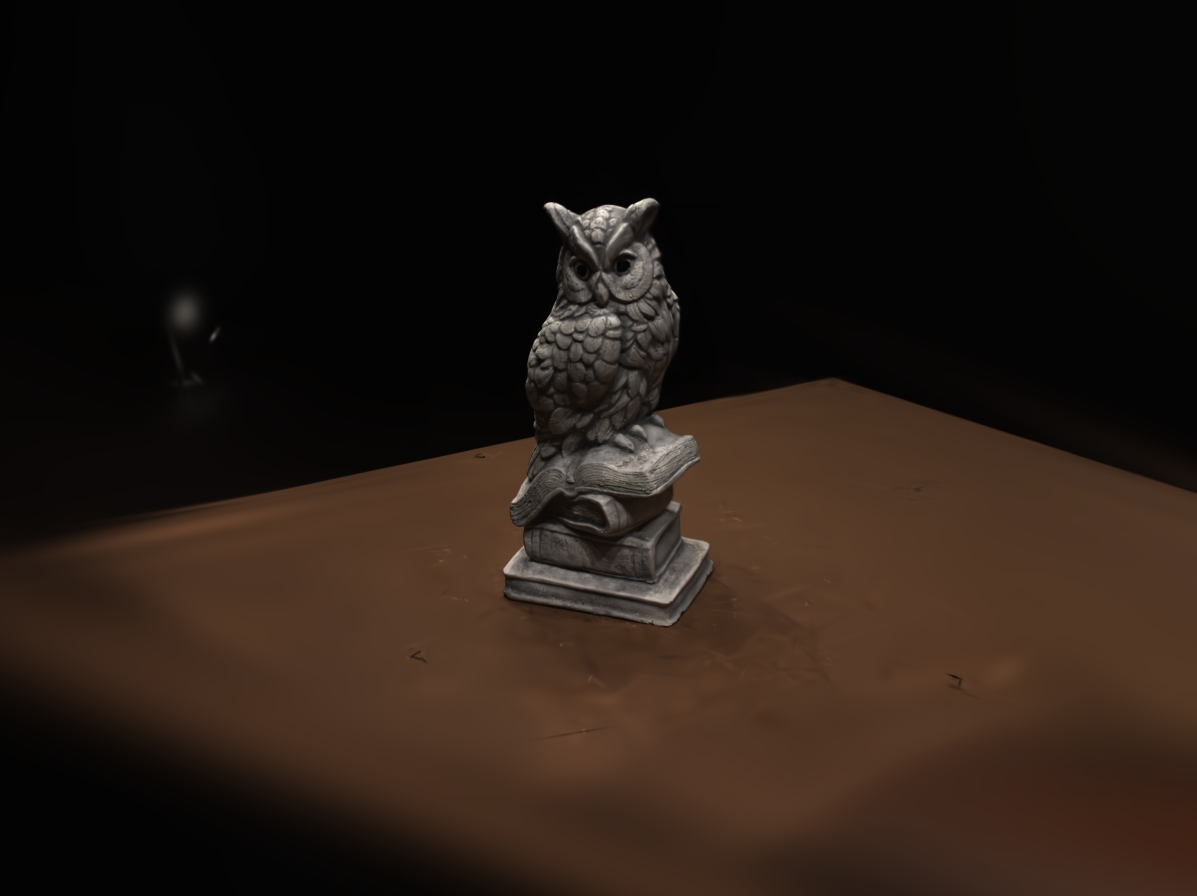} &
        \hspace{3cm}
        \\[2ex]
        
    \end{tabular}
    \caption{\textbf{Rendering quality} comparison on the DTU dataset for the \textbf{same PSNR}.}
     \label{fig:Qualitative_rendering}
\end{figure*}

The rendering quality (Figure \ref{fig:Qualitative_rendering}) shown by the rendered test images also underlines the overall strong performance of FeatureGS compared to 3DGS. 
The results on the FeatureGS configurations that yielded the highest floater reduction for the respective scene are shown. It is evident that the geometric loss terms of FeatureGS significantly reduce the floater artifacts while maintaining the same quantitative rendering quality. Large dark floater artifacts disappear in hardly all scenes. In addition, the scenes appear smoother, which can be seen, e.g., in the subsoil of objects. Since the PSNR values are the same, the high PSNR value is supposedly due to the focus being on rendering the object itself and not overfitting the entire scene, which causes the creation of floater artifacts. It can also be seen that the floaters that were visible in the figures of the geometric accuracies (Figure \ref{fig:Qualitative_c2c}) are actually also clearly present in the synthetically rendered results. Therefore, they cannot only be removed by filtering the Gaussians with e.g. very small opacity values. In addition, FeatureGS also removes artifacts which merge with the objects and leads to a kind of smoothing effect, such as in scan55 or scan69.

\section{Discussion}

The evaluation of FeatureGS under fixed training time and fixed rendering quality conditions highlights its clear superiority over 3DGS in terms of geometric accuracy, floater artifact reduction, and memory efficiency. Although these improvements are accompanied by a slight compromise in rendering quality, they demonstrate the robustness and scalability of FeatureGS suitable for various applications.

Under fixed number of 15\,000 training iterations, FeatureGS achieves clear enhancements in geometric accuracy, reducing the mean Chamfer cloud-to-cloud distance by approximately 20\%. Thereby, the loss formulations of FeatureGS with $L_\text{Planarity, Gaussian}$, and $L_\text{Eigenentropy, kNN}$, frequently deliver the best results. Nonetheless, the differences between the four FeatureGS losses are generally small and stable across different scenes. This stability suggests that FeatureGS performs consistently well, regardless of scene complexity. In addition, the reduction in floater artifacts is remarkable. FeatureGS suppresses floater artifacts by around 90\% compared to 3DGS. Among the loss with $L_\text{Planarity, Gaussian}$ demonstrates the strongest ability to minimize artifacts.
Another standout advantage of FeatureGS is its drastic reduction in the number of Gaussians, achieving an average reduction of 95\%. This leads to significant storage savings, making FeatureGS highly suitable for large-scale applications. However, these improvements come with a slight decrease in rendering quality, with an average drop in PSNR of approximately 3.3\,dB.

Under fixed rendering quality conditions using early-stopping to ensure comparable PSNR values, FeatureGS continues to demonstrate a superior performance. It improves the geometric accuracy by approximately 30\% over 3DGS, achieving a 0.5\,mm better Chamfer cloud-to-cloud distance compared to 3DGS. Additionally, FeatureGS maintains its advantage in floater artifact suppression by achieving a reduction of approximately 90\%. This is consistent across all four FeatureGS loss formulations. Furthermore, even under the identical rendering quality, FeatureGS reduces the number of Gaussians by about 90\%. This ensures a significant memory efficiency without compromising the geometric accuracy. A similar behavior is evident in the qualitative results of the point clouds, colorized by Chamfer distance, and the rendered images. With FeatureGS, the accuracy of surface points increases clearly across all scenes. Additionally, floater artifacts associated with high geometric inaccuracies are removed by FeatureGS. These artifacts are also absent in the rendered images from FeatureGS, and a kind of smoothing effect in external areas is observed. Since the PSNR for the rendered images is the same in this scenario, we suggest that the geometric loss terms shift the focus of rendering quality to the object's surface rather than overfitting the background, which likely leads to floaters.

For a denser point cloud, we suggest an additional splitting of the Gaussians depending on their size in 3D space or screen size. This would result in more Gaussians than necessary for the 3D scene representation, but a more uniform and higher point density. For the necessity of a trade-off between geometric accuracy and rendering quality, we recommend adjusting the weighting of the loss hyperparameter depending on the purpose of the application of FeatureGS.

\section{Conclusion}

In conclusion, FeatureGS strikes a remarkable balance between geometric accuracy, and floater artifact suppression and memory efficiency, through integrating 3D shape feature properties into the optimization process of 3D Gaussian Splatting with additional geometric loss terms.
By maintaining the same photometric rendering quality, it reduces the number of Gaussians needed by 90\%, while improving geometric accuracy by 30\%. This results in a more geometric accurate 3D scene representation with clearly fewer floater artifacts. 
Altogether, all proposed 3D features offer a clear benefit and only differ slightly. However, with the same rendering quality, the 3D feature 'planarity' of Gaussians itself provides the highest geometric accuracy, while the 3D feature 'omnivariance' in the Gaussian neighborhood reduces the floater artifacts and thus the number of Gaussians the most.
As a result, FeatureGS allows the direct use of Gaussian centers as a geometric representation. Although a trade-off between geometric accuracy and rendering quality may be necessary depending on the application, FeatureGS offers an effective solution for geometrically high-accurate, memory-efficient and almost artifact-free 3D scene reconstruction.

{\small
\bibliographystyle{ieee_fullname}
\bibliography{egbib}
}

\end{document}